\documentclass[sigconf]{acmart}

\usepackage{etoolbox}

\usepackage[utf8]{inputenc}
\usepackage[export]{adjustbox}
\usepackage{breqn}
\usepackage{amsmath}
\usepackage{amsfonts}
\usepackage{mathtools}

\usepackage{amsfonts}
\usepackage{mathtools}
\usepackage{nicefrac}

\def\P{\calP}

\def\X{\calX}

\def\EE{\bbE}

\def\PP{\bbP}

\def\RR{\bbR}

\def\calX{\mathcal{X}}

\def\bbE{\mathbb{E}}

\def\bbP{\mathbb{P}}

\def\bbR{\mathbb{R}}

\makeatletter
\let\IfStar\@ifstar 

\newcommand*{\eqdef}{\@ifstar{\eqdef@B}{\eqdef@A}}
\newcommand*{\eqdef@A}{\stackrel{{\scriptscriptstyle \mathrm{def}}}{\coloneqq}}
\newcommand*{\eqdef@B}{\coloneqq}

\newcommand*{\defeq}{\@ifstar{\defeq@B}{\defeq@A}}
\newcommand*{\defeq@A}{\stackrel{{\scriptscriptstyle \mathrm{def}}}{\eqqcolon}}
\newcommand*{\defeq@B}{\eqqcolon}

\newcommand*{\set}{\@ifstar{\set@B}{\set@A}}
\newcommand*{\set@A}[1]{\{#1\}}
\newcommand*{\set@B}[1]{\left\{#1\right\}}

\newcommand*{\abs}{\@ifstar{\abs@B}{\abs@A}}
\newcommand*{\abs@A}[1]{|#1|}
\newcommand*{\abs@B}[1]{\left|#1\right|}

\newcommand*{\norm}{\@ifstar{\norm@B}{\norm@A}}
\newcommand*{\norm@A}[1]{\Vert#1\Vert}
\newcommand*{\norm@B}[1]{\left\lVert#1\right\rVert}

\newcommand*{\lr}{\@ifstar{\lr@B}{\lr@A}}
\newcommand*{\lr@A}[1]{(#1)}
\newcommand*{\lr@B}[1]{\left(#1\right)}

\newcommand*{\LR}{\@ifstar{\LR@B}{\LR@A}}
\newcommand*{\LR@A}[1]{[#1]}
\newcommand*{\LR@B}[1]{\left[#1\right]}

\newcommand*{\noRelSpace}[1]{#1} 
\newcommand*{\LMR}{\@ifstar{\LMR@B}{\LMR@A}}
\newcommand*{\LMR@A}[2]{[#1 \mid \noRelSpace{#2}]}
\newcommand*{\LMR@B}[2]{\left[#1\middle|\noRelSpace{#2}\right]}

\newcommand*{\lmr}{\@ifstar{\lmr@B}{\lmr@A}}
\newcommand*{\lmr@A}[2]{(#1 \mid \noRelSpace{#2})}
\newcommand*{\lmr@B}[2]{\left(#1\middle|\noRelSpace{#2}\right)}

\newcommand*{\vecList}{\@ifstar{\vecList@B}{\vecList@A}}
\newcommand*{\vecList@A}[1]{{\LR{#1}}^\top}
\newcommand*{\vecList@B}[1]{{\LR*{#1}}^\top}

\makeatother

\newcommand*{\IfSubscript}[2]{\ifx&#2&{#1}\else{#1_{#2}}\fi}

\newcommand*{\prob}[1]{\IfSubscript{\operatorname*{\PP}}{#1}\LR}
\newcommand*{\expe}[1]{\IfSubscript{\operatorname*{\EE}}{#1}\LR}

\newcommand*{\condprob}[1]{\IfSubscript{\operatorname*{\PP}}{#1}\LMR}
\newcommand*{\condexpe}[1]{\IfSubscript{\operatorname*{\EE}}{#1}\LMR}

\newcommand*{\err}[1]{\IfSubscript{\operatorname*{err}}{#1}\lr}

\usepackage[ruled, lined, longend,noend,linesnumbered]{algorithm2e}
\SetKwRepeat{Do}{do}{while}%
\usepackage{booktabs,multirow}
\usepackage{pifont}
\usepackage{hyperref}
\usepackage{lscape}
\usepackage{makecell}
\usepackage{tabularx}
\usepackage{array}

\DeclareMathOperator*{\argmax}{arg\,max}

\usepackage{enumerate}
\usepackage[shortlabels]{enumitem}
\usepackage[parfill]{parskip} 
\usepackage{hyperref}
\usepackage{graphicx}

\usepackage{tikz}
\usetikzlibrary{arrows,shapes}
\usetikzlibrary{arrows.meta}

\makeatother\usepackage[justification=raggedright,singlelinecheck=false]{subcaption}
\usepackage{float}

\newcommand{\pr}{\mathbb{P}}

\newcommand{\nperp}{\mathrlap{\perp}{\;/}}

\usepackage{xcolor}
\usepackage[colorinlistoftodos]{todonotes}

\definecolor{caribbeangreen}{rgb}{0.0, 0.8, 0.6}

\definecolor{commentColorCarlos}{rgb}{0.6, 0.8, 0.0}

\usepackage{amsthm}
\usepackage{thmtools}

\pgfarrowsdeclarecombine*{ring and diamond}{ring and diamond}{open diamond}{open diamond}{o}{o}
\def\CustomArrow#1{\mathrel{\raisebox{0.6ex}{\tikz[baseline]{\draw[#1](0,0)--(1.5em,0);}}}}
\def\CustomArrowShortest#1{\mathrel{\raisebox{0.6ex}{\tikz[baseline]{\draw[#1](0,0)--(0.5em,0);}}}}

\def\angles#1{\ensuremath{\langle#1\rangle}}
\def\anglesSBCN#1#2{\angles{\mathsf{#1}=\mathsf{#2}}}

\declaretheoremstyle[
  spaceabove=1em plus 0.75em minus 0.25em,
  spacebelow=1em plus 0.75em minus 0.25em,
  qed={\itshape \scriptsize (End)},
]{exmpstyle2}

\numberwithin{equation}{section}
\theoremstyle{plain}
\newtheorem{theorem}{\protect\theoremname}
\theoremstyle{definition}

\theoremstyle{plain}

\theoremstyle{definition}
\newtheorem{definition}[theorem]{\protect\definitionname}
\theoremstyle{definition}

\theoremstyle{definition}

\providecommand{\definitionname}{Definition}
\providecommand{\examplename}{Example}
\providecommand{\lemmaname}{Lemma}
\providecommand{\theoremname}{Theorem}
\providecommand{\factname}{Fact}
\providecommand{\propositionname}{Proposition}

\renewcommand\footnotetextcopyrightpermission[1]{} 
\AtBeginDocument{%
  \providecommand\BibTeX{{%
    \normalfont B\kern-0.5em{\scshape i\kern-0.25em b}\kern-0.8em\TeX}}}

\begin{document}

\title{Causal Discovery for Fairness}

\author{R\=uta Binkyt\.e-Sadauskien\.e}
\email{ruta.binkyte-sadauskiene@inria.fr}
\affiliation{%
  \institution{INRIA, École Polytechnique, IPP}
  \city{Paris}
  \country{France}
}

\author{Karima Makhlouf}
\email{karima.makhlouf@lix.polytechnique.fr}
\orcid{0000-0001-6318-0713}
\affiliation{%
  \institution{INRIA, École Polytechnique, IPP}
  \city{Paris}
  \country{France}
}

\author{Carlos~Pinzón}
\email{carlos.pinzon@inria.fr}
\affiliation{%
  \institution{Inria, École Polytechnique, IPP}
  \city{Paris}
  \country{France}
}

\author{Sami Zhioua}
\email{sami.zhioua@lix.polytechnique.fr}
\orcid{0000-0001-7491-6271}
\affiliation{%
  \institution{INRIA, École Polytechnique, IPP}
  \city{Paris}
  \country{France}
}

\author{Catuscia Palamidessi}
\email{catuscia@lix.polytechnique.fr}
\orcid{0000-0003-4597-7002}
\affiliation{%
  \institution{Inria, École Polytechnique, IPP}
  \city{Paris}
  \country{France}
}

\renewcommand{\shortauthors}{Makhlouf, et al.}

\begin{abstract}
    It is crucial to consider the social and ethical consequences of AI and ML based decisions for the safe and acceptable use of these emerging technologies. Fairness, in particular, guarantees that the ML decisions do not result in discrimination against individuals or minorities. Identifying and measuring reliably fairness/discrimination is better achieved using causality which considers the causal relation, beyond mere association, between the sensitive attribute (e.g. gender, race, religion, etc.) and the decision (e.g. job hiring, loan granting, etc.). The big impediment to the use of causality to address fairness, however, is the unavailability of the causal model (typically represented as a causal graph). Existing causal approaches to fairness in the literature do not address this problem and assume that the causal model is available. In this paper, we do not make such assumption and we review the major algorithms to discover causal relations from observable data.  This study focuses on causal discovery and its impact on fairness. In particular, we show how different causal discovery approaches may result in different causal models and, most importantly, how even slight differences between causal models can have significant impact on fairness/discrimination conclusions. These results are consolidated by empirical analysis using synthetic and standard fairness benchmark datasets. The main goal of this study is to highlight the importance of the causal discovery step to appropriately address fairness using causality.  
\end{abstract}

\maketitle
\pagestyle{plain}

\section{Introduction}
\label{intro}

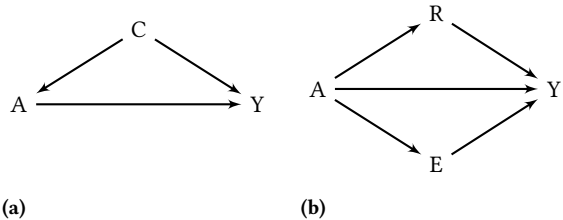
\begin{figure}[htp!]  
\centering 
  \begin{subfigure}[b]{0.4\linewidth}
\begin{tikzpicture}[%
        >=latex',
        scale=0.9
      ]
        \draw[-,thick] (0,0) node[left] () {} ;
        \draw[-,thick] (4.5,0) node[left] () {} ;
        \draw[-,thick] (0,1) node[left] (A) {A} ;
        \draw[-,thick] (3.5,1) node[left] (Y) {Y} ;
        \path (A) -- coordinate (middle) (Y);
        \node[above of=middle] (C) {C};
        \draw[->,thick] (A) -- (Y);
        \draw[->,thick] (C) -- (A);
        \draw[->,thick] (C) -- (Y);
    \end{tikzpicture}
    \caption{} \label{subfig:example1a}  
  \end{subfigure}
  \hspace{0.05\linewidth}
\begin{subfigure}[b]{0.4\linewidth}
  \begin{tikzpicture}[%
        >=latex',
        scale=0.9
      ]
        \draw[->,thick] (0,0) node[left] (A) {A} -- (3,0) node[right] (Y) {Y}; 
        \path (A) -- coordinate (middle) (Y);
        \node[above of=middle] (R) {R};
        \path (A) -- coordinate (middle2) (Y);
        \node[below of=middle2] (E) {E};
        \draw[->,thick] (A) -- (R);
        \draw[->,thick] (R) -- (Y);
        \draw[->,thick] (A) -- (E);
        \draw[->,thick] (E) -- (Y);
    \end{tikzpicture}
\caption{} \label{subfig:example1b}  
\end{subfigure}
  \hspace{0.05\linewidth}
\caption{Causal graphs illustrating confounding (a) and mediation analysis (b).}
\label{fig:example1}
\end{figure}

With  the proliferation of machine learning based automated decision systems, there are growing concerns about discrimination on the basis of several personal characteristics (e.g. race, color, religion, sex, age, or national origin) prohibited by human rights laws~\cite{InternationalHR,ECHR,act1964civil}.
As decisions of such systems may have critical impacts on people's lives (e.g. job hiring, loan granting, predicting recidivism during parole, etc.), 
addressing the discrimination problem is crucial to safely use these automated systems.
Several fairness criteria have been introduced in the literature to assess discrimination (statistical parity~\cite{darlington1971}, equal opportunity~\cite{hardt2016equality}, calibration~\cite{chouldechova2017fair}, etc.)~\cite{makhlouf2021ipm}.
The most recent fairness criteria, however, are causal-based~\cite{makhlouf2020causal} and reflect the now widely accepted idea that causality is necessary to appropriately address the problem of fairness.
There are at least three benefits of using causality to assess fairness.
First, in presence of a common cause (confounder) between the sensitive attribute $A$ (e.g. gender) and the decision $Y$ (e.g. job hiring), using conditional probability $\pr(Y|A)$ leads to wrong conclusions about the dependence of $Y$ on $A$ (Figure~\ref{subfig:example1a}).
Confounders are the reason why we say that ``correlation is different than causation''.
A more reliable measure of the dependence between $Y$ and $A$ is the causal effect of $A$ on $Y$ which is typically computed by adjusting on confounders.
Second, causality is well equipped to carry out mediation analysis, that is, distinguishing the different paths of causal effects.
As shown in Figure~\ref{subfig:example1b}, a causal effect between $A$ and $Y$ can be classified as direct ($A\rightarrow Y$), indirect ($A\rightarrow R \rightarrow Y$ and $A \rightarrow E \rightarrow Y$), or a path-specific effect (only through $A \rightarrow R \rightarrow Y$ or $A \rightarrow E \rightarrow Y$).

This is very relevant to fairness as a direct effect is always unfair, while an indirect or a path-specific effect may be unfair or fair depending on the mediator variable: an indirect effect through a redlining/proxy variable ($R$) is unfair, while an indirect effect through an explaining variable ($E$) is acceptable (fair).
In the simple example of job hiring, $R$ might be the hobby of the candidate which generally indicates the gender of the candidate (a mechanical hobby indicates typically a male candidate) and $E$ might be the education level of the candidate which can be used to justify an observed discrimination between male and female candidates.
Third, in some legal liability frameworks such as disparate treatment~\cite{barocas2016big}, discrimination claims require the plaintiff to demonstrate a causal connection between the challenged decision (e.g. hiring, firing, admission) and the sensitive attribute (e.g. gender, race, age).
It is then necessary to investigate the causal relationship between $A$ and $Y$, not the mere statistical correlation between them.

Reasoning about causal effects between variables is called causal inference.
The main impediment to causal inference is the unavailability of the true causal graph which indicates the causal relations between variables.
Causal graphs can be set manually by experts in the field, but very often generated using experiments (called also interventions).
The process of identifying the causal graph is called causal discovery or structure learning. 

The gold standard of causal discovery is to perform randomized controlled trials (RCT)~\cite{fisher1936design}.
RCT\footnote{A commonly used variant of RCT is A/B testing~\cite{abtesting20}.}  consists in randomly selecting subjects from the population, allocating each one of them randomly to one of two groups (treatment and control/placebo), and then comparing the two groups with respect to an outcome.
RCT is generally not feasible because of practical, ethical, and scalability reasons.
For instance, assessing the causal relation of the gender $A$ on the hiring decision $Y$ requires changing the gender of a randomly selected job candidate.
Even if such intervention/experiment was possible, there are practical concerns about the number of experiments to be performed.

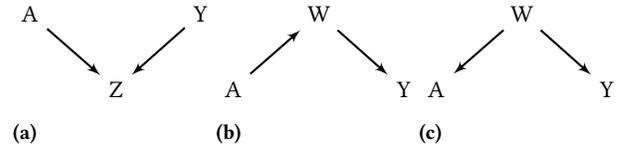
\begin{figure}[htp!]  
\centering 
  \begin{subfigure}[b]{0.25\linewidth}
\begin{tikzpicture}[%
        >=latex',
        scale=0.9
      ]
        \draw[-,thick] (0,0) node[left] (A) {A};
        \draw[-,thick] (2.5,0) node[left] (Y) {Y}; 
        \path (A) -- coordinate (middle) (Y);
        \node[below of=middle] (Z) {Z};
        \draw[->,thick] (A) -- (Z);
        \draw[->,thick] (Y) -- (Z);
    \end{tikzpicture}
    \caption{} \label{subfig:example2a}  
  \end{subfigure} 
  \hspace{0.05\linewidth}
\begin{subfigure}[b]{0.25\linewidth}
  \begin{tikzpicture}[%
         >=latex',
         scale=0.9
      ]
        \draw[-,thick] (0,0) node[left] (A) {A};
        \draw[-,thick] (2.5,0) node[left] (Y) {Y}; 
        \path (A) -- coordinate (middle) (Y);
        \node[above of=middle] (W) {W};
        \draw[->,thick] (A) -- (W);
        \draw[->,thick] (W) -- (Y);
    \end{tikzpicture}
\caption{} \label{subfig:example2b}  
\end{subfigure}
\hspace{0.05\linewidth}
\begin{subfigure}[b]{0.25\linewidth}
\begin{tikzpicture}[%
         >=latex',
        scale=0.9
      ]
        \draw[-,thick] (0,0) node[left] (A) {A};
        \draw[-,thick] (2.5,0) node[left] (Y) {Y}; 
        \path (A) -- coordinate (middle) (Y);
        \node[above of=middle] (W) {W};
        \draw[->,thick] (W) -- (A);
        \draw[->,thick] (W) -- (Y);
    \end{tikzpicture}
\caption{} \label{subfig:example2c}  
\end{subfigure}
\hspace{0.05\linewidth}
\caption{The three basic structures of causal graphs: collider (a), mediator (b), and confounder (c).}
\label{fig:example2}
\end{figure}

As an alternative to RCT, causal discovery is typically carried out using statistical tests on observable data.
For instance, in presence of three variables $A$, $Y$, and $Z$, if $A$ and $Y$ are independent ($A \perp Y$) but dependent when conditioned on $Z$ ($A \nperp Y | Z$), then the causal graph is the one in Figure~\ref{subfig:example2a}\footnote{$Z$ is called a collider.}.
However, even assuming the availability of an oracle that returns answers about conditional independencies in the data, a causal discovery procedure can still be undecided about the causal graph.
For instance, if we know all conditional independence relations between the three variables $A$, $Y$, and $W$, that is, $A \nperp Y$, $A \nperp W$, $W \nperp Y$, and $A \perp Y | W$, then it is not possible to tell if the causal graph is the one in Figure~\ref{subfig:example2b} or~\ref{subfig:example2c}, no matter how much data is available.
It is said that both graphs belong to the same Markov equivalence class because they imply the same independence constraints.
In such cases, a researcher can use some background knowledge about the problem at hand to rule out some causal relations and hence narrow down the set of valid causal graphs.
The most common sources of background knowledge are temporal order (e.g. treatment happens before a symptom) and experimental design~\cite{haslam2004experimental}.
For instance, using the example above, if the variable $A$ is always set before the variable $W$, the graph of Figure~\ref{subfig:example2c} will be ruled out leaving only the graph in Figure~\ref{subfig:example2b} as compatible with the input data. This is achieved by splitting variables into \textit{tiers} and placing variable $A$ in tier1 while $W$ in tier2.  

Another difficulty of the causal discovery task is that, unlike supervised learning, the true causal graph (ground truth) is typically not available.
As a consequence, evaluating the performance of causal discovery algorithms is not always possible.

A large number of causal discovery algorithms exist in the literature.
The majority of these algorithms fall into three categories: constraint-based, score-based, and procedures that exploit semi-parametric assumptions.
In the constraint-based category, algorithms rely mainly on the (conditional) independencies present in the data to discover causal relations between variables as explained in the previous paragraph.
Therefore their efficiency depends on the reliability of the conditional independency test procedure.
Score-based algorithms rely instead on goodness-of-fit tests.
They learn causal graphs by maximizing a scoring criteria such as the Bayesian Information Criterion (BIC)~\cite{schwarz1978estimating} which trades-off accuracy (fitness of graph to the data) with complexity (the number of parameters in the model).
Algorithms in the third category use additional assumptions to learn causal relations more efficiently and in more details.
The most common assumptions relevant to the third category are linearity of the model and non-gaussianity of the regression residuals.
As described, algorithms in the two first categories do not make strong assumptions on the parametric form or functions for the causal connections.

Therefore, they can be, theoretically, applied on much more scenarios than the third category. However, the majority of available implementations of constraint-based and score-based causal discovery algorithms model variables as multivariate Gaussian mixture, which implies linearity and Gaussianity of all continuous variables.
Causal graphs returned by algorithms in the third category are more accurate than those of the two first categories which are simply Markov equivalence classes.

This paper studies the problem of discovering causal graphs to be used to assess fairness of machine learning based decision systems.

  
To illustrate the different approaches of causal discovery, we describe in detail one notable algorithm in each of the three categories, namely, PC algorithm~\cite{spirtes1999algorithm} and its FCI extension~\cite{spirtes1999algorithm} for the constraint-based category, GES~\cite{hauser2012gies} for the score-based category, and directLiNGAM~\cite{shimizu2011directlingam} for the third category. All these approaches are assuming the standard interpretation of causality (based on interventions~\cite{pearl2009causality}).
In addition, we describe a causal discovery algorithm for SBCN (Suppes Bayes Causal Networks)~\cite{bonchi2017exposing} which is a specific type of causal graph to measure fairness but which is based on a different interpretation of causal relations (probabilistic causation~\cite{hitchcock2002probabilistic}).

Given a dataset, selecting which causal discovery algorithm to use depends on the type of data, how it is pre-processed, and on the assumptions made.
Some of the algorithms require the same type for the variables (i.e.
all variables must be continuous, binary, etc.) while others accept mixed data types.
On the other hand, to be applicable, most of algorithms make some assumptions on the causal model.
This includes, causal Markov, causal faithfulness, causal sufficiency, linearity, and gaussianity.

The main contributions of this paper are the following:
\begin{itemize}
    \item We provide an intuitive explanation of the major causal discovery algorithms.
    \item We discuss relevant details about their application in practice, including the assumptions, types of data, and (conditional) independence tests.
    \item We carry out an experimental analysis to illustrate the impact of causal discovery procedure on the structure of the causal graph and consequently on fairness conclusions.
\end{itemize}

\section{Related work}
\label{sec:related}

Several survey papers on causal discovery can be found in the literature~\cite{glymour2019review,malinsky2018causal,spirtes2016survey,guo2020survey,yu2016review,cheng2022evaluation, nogueira2021causal,nogueira2022methods}.
Most of these surveys use the same classification that is used in this paper, namely, constraint-based, score-based, and those relying on additional semi-parametric assumptions.
Glymour et al.~\cite{glymour2019review} review causal discovery algorithms and their application in the biology and neurosciences fields.
In particular they provide general guidelines for their applicability in practice.
Malinsky and Danks~\cite{malinsky2018causal} focus more on their application for problems in Philosophy. 
The recent survey of Cheng et al.~\cite{cheng2022evaluation} address mainly the evaluation procedures of causal discovery approaches.
Unlike existing surveys, this study provides a detailed but concise description of the major causal discovery algorithms (i.e.
PC, FCI, GES, and LiNGAM) and features a comparative empirical analysis.
But more importantly it tackles the causal discovery problem in the context of fairness.

Existing causal-based fairness approaches in the literature clear up the causal graph problem in two ways.
Either they assume that the causal graph is known~\cite{nabi2018fair,chiappa2019path,kusner2017counterfactual} or they use the tetrad~\cite{tetrad} implementation of PC algorithm~\cite{zhang2016causal,wu2019counterfactual,wu18Ranking,wu2020effort,zhang2017achieving,silva2020}.
Both ways are akin to skipping the important step of causal discovery from observable data and its impact on the fairness conclusions.
In this paper, we consider major causal discovery algorithms and illustrate the importance of the (different) graph structures on causal-based fairness notions.

\section{Preliminaries and notation}
\label{sec:preliminaries}
Variables are denoted by capital letters.
In particular, $A$ is used for the sensitive variable (e.g. gender, race, etc.) and $Y$ is used for the outcome of the automated decision system (e.g. hiring, admission, etc.).
Small letters denote specific values of variables (e.g., $A=a$, $W=w$).
Bold capital and small letters denote a set of variables and a set of values, respectively.
In particular, $\mathbf{V}$ denotes the set of all variables in the data.


A directed acyclic graph (DAG) $\mathcal{G} = (\mathbf{V},\mathcal{E})$ is composed of a set of variables/vertices $\mathbf{V}$ and a set of (directed) edges $\mathcal{E}$ between them such that no cycle is formed.
Let $\mathcal{P}$ be the probability distribution over the same set of variables $\mathbf{V}$.
$\mathcal{G}$ and $\mathcal{P}$ are related through the Markov condition if every variable is conditionally independent of its non-descendants given its parents.
Assuming the Markov condition, the joint distribution of variables $V_1,V_2,\ldots \in \mathbf{V}$ can be factorized as:
\begin{equation}
    \pr(V_1,V_2,\ldots) = \prod_{i}\pr(V_i | Pa(V_i)) \label{eq:markov}
\end{equation}
where $Pa(V_i)$ denotes the set of parents of $V_i$.

Pairing a DAG $\mathcal{G}$ and a probability distribution $\mathcal{P}$ such that they are related with the Markov condition forms a Bayesian network (BN) $\mathcal{B} = \langle \mathcal{G},\mathcal{P}\rangle$~\cite{pearl1988probabilistic}.
A causal bayesian network (CBN)~\cite{pearl2009causality} is a BN where edges have causal interpretations.
That is, an edge between two variables $V_i$ and $V_j$ ($V_i \rightarrow V_j$) means that if all other variables are fixed to some values and we change the value of $X_i$, then $X_j$ will possibly change, but never the other way around.
In the rest of the paper we call a causal graph: a DAG that describes the causal relations between variables.
That is, a directed edge $V_i \rightarrow V_j$ indicates a causal relation from the cause variable $V_i$ to the effect variable $V_j$\footnote{A causal graph corresponds to the graph component of a CBN (the $\mathcal{G}$ in $\langle \mathcal{G},\mathcal{P}\rangle$).}. A partially directed acyclic graph (PDAG) is a special type of DAG that contains directed and undirected edges.


Conditional independence between variables can be graphically identified using the d-separation criterion~\cite{pearl2009causality}.
A path\footnote{A sequence of directed edges between two variables not necessarily pointing to the same direction.
For instance, $A \leftarrow C \rightarrow Y$ in Figure~\ref{subfig:example1a} is path although the edges are not pointing to the same direction.} $p$ is $d-$separated (or blocked) by a set of vertices $\mathbf{W}$ if and only if (1) if $p$ contains a chain ($X \rightarrow M \rightarrow Y$) or a fork ($X \leftarrow M \rightarrow Y$), then $M$ is in $\mathbf{W}$ and (2) if $p$ contains an inverted fork\footnote{Called also v-structure.} ($X \rightarrow C \leftarrow Y$), then the collider $C$ and all its descendants are not in $\mathbf{W}$.
If a set $\mathbf{W}$ d-separates (blocks) every path from $X$ to $Y$, then $X$ and $Y$ happen to be conditionally independent given $\mathbf{W}$.
For instance, in Figure~\ref{subfig:example3a}, the set $\mathbf{W}=\{S,M\}$ $d-$separates $A$ and $Y$, hence $A$ and $Y$ are conditionally independent given $\{S,M\}$ ($A \perp Y | S, M)$.
DAGs that have the same d-separation properties are called Markov equivalent and imply the same conditional independence relations.
Any maximal collection of DAGs, all of which are Markov equivalent, is called a Markov Equivalence Class (MEC). A completed partially directed acyclic graph (CPDAG) is a special type of PDAG that serves as representative for Markov equivalence classes of DAGs.

Given a causal graph $\mathcal{G}$, the causal relations between variables (vertices) can be represented using structural equations of the form $X = f_X(Pa(X),\epsilon_X)$ where $\epsilon_X$ denotes an error term that represents random sources of noise. 
For instance, in Figure~\ref{subfig:example3a}, $A = f_A(\epsilon_A)$, $S = f_S(A,\epsilon_A)$, $Y = f_Y(S,M,\epsilon_Y)$, etc.

\begin{figure}[htp!]  
\begin{subfigure}[b]{0.25\linewidth}
  \begin{tikzpicture}[%
        >=latex',
        scale=0.8,
      ]
        \draw[-,thick] (0,0) node[left] (A) {A};
        \draw[-,thick] (2.5,0) node[left] (Y) {Y}; 
        \path (A) -- coordinate (middle) (Y);
        \node[above of=middle] (S) {S};
        \path (A) -- coordinate (middle2) (Y);
          \node[below of=S] (M) {M};
        \node[below of=middle2] (Z) {Z};
        \draw[->,thick] (A) -- (S);
        \draw[->,thick] (A) -- (M);
        \draw[->,thick] (M) -- (Y);
        \draw[->,thick] (S) -- (Y);
        \draw[->,thick] (A) -- (Z);
        \draw[->,thick] (Y) -- (Z);
    \end{tikzpicture}
\caption{} \label{subfig:example3a}  
\end{subfigure}
\hspace{0.03\linewidth}
\begin{subfigure}[b]{0.25\linewidth}
  \begin{tikzpicture}[%
        >=latex',
        scale=0.6
      ]
        \draw[-,thick] (0,0) node[left] (A) {A};
        \draw[-,thick] (2.5,0) node[left] (Y) {Y}; 
        \path (A) -- coordinate (middle) (Y);
        \node[below of=middle2] (M) {M};
        \draw[->,thick] (A) -- (M);
        \draw[->,thick] (M) -- (Y);
        \draw[->,thick] (A) -- (Y);
    \end{tikzpicture}
\caption{} \label{subfig:example3b}  
\end{subfigure}
\hspace{0.02\linewidth}
\begin{subfigure}[b]{0.25\linewidth}
  \begin{tikzpicture}[%
        >=latex',
        scale=0.8,
      ]
        \draw[-,thick] (0,0) node[left] (A) {A};
        \draw[-,thick] (2.5,0) node[left] (Y) {Y}; 
        \draw[-,thick] (3.5,0) node[left] (Z) {Z}; 
        \draw[-,thick] (1.1,0) node[left] (W) {W};
        \path (A) -- coordinate (middle) (Z);
        \node[above of=middle] (S) {S};
        \path (A) -- coordinate (middle2) (Y);
        \node[below of=middle2] () {};
        \draw[->,thick] (A) -- (W);
        \draw[dashed,->] (S) -- (W);
        \draw[dashed,->] (S) -- (Y);
        \draw[->,thick] (Z) -- (Y);
    \end{tikzpicture}
\caption{} \label{subfig:example3c}  
\end{subfigure}
\caption{}
\label{fig:example3}
\end{figure} 

Discovering causal relations from observable data requires typically a set of assumptions~\cite{eberhardt2009introduction}.
The most commonly used assumptions are causal Markov condition, causal faithfulness, causal sufficiency, linearity, and gaussianity (or non-gaussianity). 

\textit{\textbf{Causal Markov condition}} (Eq.~\ref{eq:markov}): there is a general consensus that it is fundamental to causal inference and hence typically required. 

\textit{\textbf{Causal faithfulness}}: a causal graph $\mathcal{G}$ and a probability distribution $\mathcal{P}$ over the same variables $\mathbf{V}$ are faithful to each other if all and only the conditional independence relations that hold in $\mathcal{P}$ are entailed by the Markov condition and d-separation in $\mathcal{G}$.
An example of faithfulness violation is when two variables are dependent in the causal graph but independent in the data.
Consider the graph in Figure~\ref{subfig:example3b}.
If, in the data, the direct causal effect of $A$ on $Y$ is exactly balanced out by the indirect causal effect $A \rightarrow M \rightarrow Y$, $A$ and $Y$ will appear independent in the data while they are dependent in the graph.
In that case, a causal discovery procedure assuming faithfulness will return a collider structure on $M$ ($A \rightarrow M \leftarrow Y$) as causal graph.

\textit{\textbf{Causal sufficiency}}: there are no latent (hidden) confounders between variables in $\mathbf{V}$.
It is a very strong assumption as its absence or presence may lead to very different causal graphs.
Violation of causal sufficiency may sometimes be detected from data.
For instance, in presence of four variables if the only independencies observed in the data are: $A \perp Y$, $A \perp Z$, $W \perp Z$, $A \perp Y | Z$, $A \perp Z | W$, $A \perp Z | Y$, and $W \perp Z | A$, 
then the only causal graph satisfying these constraints is the one in Figure~\ref{subfig:example3c} involving a hidden confounder variable $U$ (a dotted edge means a hidden confounder).
Hence, causal sufficiency fails in this example.

\textit{\textbf{Linearity}}: the value of every variable is determined by a linear combination of the values of its parents (causes) in addition to the error term. 
For instance, in a clinical context, a linear equation of a variable may capture a dose-response relationship with its parents. 
The linearity assumption can be considered when the data is continuous or mixed and  makes causal relations simpler and the statistical computations (regression analysis) more tractable. 
Linearity allows also to consider every causal path between two variables independently from any other causal path between the same two variables.
This is very relevant for fairness problems.


\textit{\textbf{Gaussianity of residuals}}: the error terms of structural equations have a Gaussian (normal) distribution.
This assumption is typically combined with the linearity assumption.
The main motivation is that if several individual minor sources of errors (not captured in the model) are combined, then for a large enough sample, the error distribution can reasonably be expected to be Gaussian.
The opposite assumption (non-gaussianity of error distributions), combined with the linearity assumption, is used by a several causal discovery algorithms to identify the direction of the causal relation between a given two variables~\cite{shimizu2006linear}.

\textit{\textbf{Multivariate Gaussianity}}: the joint distribution of the continuous variables follows a multivariate Gaussian distribution.
In the case of mixed data, the assumption takes the form of a mixture of multivariate distributions~\cite{andrews2018scoring}, in which for each combination of values for the discrete variables, there is a weight and a multivariate Gaussian distribution for the continuous variables.
Multivariate gaussianity implies both linearity and gaussianity of the residuals.

The following are common terms used in causal discovery algorithms.
$adj(X,\mathcal{G})$ is the set of variables adjacent to $X$ in $\mathcal{G}$.
$I(X,Y)$ denotes a statistical independence test between  variables $X$ and $Y$ that can be thresholded to generate a boolean value. 
Similarly, $I(X,Y|\mathbf{Z})$ is the conditional independence test.
A triple $(X,C,Y)$ in a graph $\mathcal{G}$ is unshielded if $X$ and $C$ as well as $C$ and $Y$ are adjacent, but $X$ and $Y$ are not adjacent in $\mathcal{G}$.
A v-structure $(X,C,Y)$ is an unshielded triple in $\mathcal{G}$ where the edges are oriented as $X \rightarrow C \leftarrow Y$.



\section{Causal Discovery Algorithms}
\label{sec:search}

The three main categories of causal discovery algorithms are constraint-based, score-based, and procedures that exploit semi-parametric assumptions.
This section describes one representative algorithm of each category.
In addition, we describe the only algorithm in the literature for discovering causal graphs for fairness purposes, namely, SBCN.

\subsection{PC Algorithm}
\label{subsec:PC}
PC algorithm~\cite{spirtes1991algorithm} is a constraint-based algorithm.
That is, it learns a set of causal graphs that satisfy the conditional independencies embedded in the data at hand.
There are two main steps.
The first (Algorithm~\ref{alg:PC_one}) takes as input the data at hand along with a significance level $\alpha$ and outputs a skeleton graph witch contains only undirected edges.
The second (Algorithm~\ref{alg:PC_two}) consists of orienting the undirected edges of the skeleton graph to form an equivalence class of DAGs.
Note that the first step contributes to most of the computational costs.

\SetKwComment{Comment}{/* }{ */}
\SetKwRepeat{Do}{do}{while}

\begin{algorithm}[ht]
  \caption{Step 1 of the PC algorithm: learning the skeleton.}\label{alg:PC_one}
  \small
  \SetAlgoLined
  \KwIn{Dataset $\mathcal{D}$ with a set of variables $\textbf{V}$, and significance level $\alpha$.}
  \KwOut{The undirected graph $G$ with a set of edges $\textbf{E}$.}
   Assume all nodes are connected initially\label{alg1-line1}\\
  Let depth $d = 0$\label{alg1-line2}\\
  \While{$|adj(X, G)\setminus{Y }| \geq d$ for every pair of adjacent vertices in $G$\label{alg1-line3}}{
    \For{each pair of adjacent vertices X and Y in G \label{alg1-line4}}{
    \If{$(|adj(X, G)\setminus{Y}| \geq d)$ \label{alg1-line5}}{
    \For{each $Z \subseteq adj(X, G)\setminus{Y }$ with $|Z| = d$ \label{alg1-line6}}
 
    \If{$I(X, Y |\textbf{Z}) \geq \alpha$ \label{alg1-line8}}{
    Remove edge between $X$ and $Y$\label{alg1-line9}\\
    Save $\textbf{Z}$ as the separating set of $(X,Y)$\label{alg1-line10}\\
    Update $G$ and $\textbf{E}$\label{alg1-line11}\\
    \textbf{break} \label{alg1-line12}\\
        } \label{alg1-line13}
      } \label{alg1-line14}
    } \label{alg1-line15}
    $d \gets d + 1$\label{alg1-line17}\\ 
  }  \label{alg1-line16}
\end{algorithm}

\begin{algorithm}[ht]
\caption{\small Step 2 of the PC algorithm: orienting the edges}\label{alg:PC_two}
\small
\SetAlgoLined
\KwIn{The undirected graph $G$ with a set of edges $\textbf{E}$.}
\KwOut{ PDAG $\P$.}
\For{each triple of vertices $(X,C,Y)$ such that $C \in adj(X, G)$ and $Y \notin adj(X, G)$ \label{alg2-line1}}{
    \If{$C \notin \textbf{Z}$ (separating set of $(X,Y)$) \label{alg2-line2}}{
        orient the edges $X - C - Y$ as $X\rightarrow C \leftarrow Y$ \label{alg2-line3}\\
    } \label{alg2-line4}
} \label{alg2-line5}
\While{unoriented edges exist \label{alg2-line6}}{
    \For{each triple of vertices $(X,C,Y)$ such that $X\rightarrow C - Y$ and $Y \notin adj(X, G)$ \label{alg2-line7}}{
    orient the edge $C - Y$ as $C \rightarrow Y$ \label{alg2-line8} \tcp*[r]{Rule 1}
    } \label{alg2-line9}
    \For{each chain $X\rightarrow C \rightarrow Y$ \label{alg2-line10}}{
    orient the edge $X - Y$ as $X \rightarrow Y$ \label{alg2-line11} \tcp*[r]{Rule 2}
    }
    \For{each two chains $X\rightarrow C_1 \rightarrow Y$ and $X\rightarrow C_2 \rightarrow Y$ such that $C_2 \notin adj(C_1, G)$ \label{alg2-line13}}{
    orient the edge $X - Y$ as $X \rightarrow Y$ \label{alg2-line14} \tcp*[r]{Rule 3}
    } \label{alg2-line15}
} \label{alg2-line16}
\end{algorithm}

As shown in Algorithm~\ref{alg:PC_one}, the PC algorithm starts with the fully connected graph (line~\ref{alg1-line1}) and relies on the conditional independence tests in order to either remove or keep edges.
For each edge $X\rightarrow Y$ and each subset $\textbf{Z}$ of all neighbors of $X$ and $Y$, the PC algorithm checks if $X$ and $Y$ are independent conditioned on $Z$(line~\ref{alg1-line8}).
The depth $d$ represents the size of the conditioning sets.
During the first iteration, all pairs of vertices are tested conditioning on the empty set $\emptyset$, i.e. $d=0$ (line~\ref{alg1-line2}).
Thus, some of the edges will be removed (line~\ref{alg1-line9}) and the algorithm will proceed only with the remaining edges in the next iteration $(d = 1)$.
The size of the conditioning set, $d$, is incremented at each new test (line~\ref{alg1-line17}) until $d$ is greater than the size of the adjacent sets of the testing vertices (condition of the while loop in line~\ref{alg1-line3}).
Note that the graph at hand is updated at each test after edge(s) deletion (line~\ref{alg1-line11}).
Moreover and most importantly, the set of conditioning variables $\textbf{Z}$ is stored (line~\ref{alg1-line10}) as it will be used later to detect potential presence of v-structures in the causal graph (Algorithm~\ref{alg:PC_two} - line~\ref{alg2-line2}).
If the conditional independence tests are reliable, then PC
returns the true graph~\cite{le2016fast}.
An important asset of the PC-algorithm is that for  high-dimensional sparse graphs, the conditional independence tests are organized in a way that makes the algorithm computationally efficient, since it only needs to test conditional independencies up to order $k-1$, where $k$ is the maximum size of the adjacency sets of the nodes in the DAG at hand.

Algorithm~\ref{alg:PC_two} takes as input the output generated by Algorithm~\ref{alg:PC_one}.
In other words, Algorithm~\ref{alg:PC_two} starts with a ``thinned'' version of the initial undirected graph $G$ and aims to orient all of its edges $\textbf{E}$.
Line~\ref{alg2-line3} detect the v-structures in $G$.
That is, it considers all unshielded triples in $G$ , and orients an unshielded triple $(X , C , Y)$ as a v-structure if and only if $C \notin$ $\textbf{Z}$ (separating set of $(X,Y)$).
Finally, the algorithm tries to orient as many of the remaining undirected edges as possible by applying repeatedly the rules shown in lines~\ref{alg2-line7} - \ref{alg2-line16} until no more edges can be oriented.

The PC-algorithm is proved to be efficient for sparse graphs.
The main reason for that is that the neighbors of a particular node are dynamically updated (line~\ref{alg1-line11} in Algorithm~\ref{alg:PC_one}) once and edge is deleted~\cite{le2016fast}.

\subsection{FCI Algorithm}
\label{subsec:FCI}

The FCI algorithm~\cite{spirtes1999algorithm} is also a constraint-based algorithm and is considered as a generalization of the PC algorithm.
The main difference between PC and FCI is that the latter takes into account the presence of common hidden confounders between observed variables.
Consequently, instead of producing a DAG, the output of FCI is a partial ancestral graph (PAG) with possibly four types of edges: $\CustomArrow{->}, \CustomArrow{<->}, \CustomArrow{o-}, \CustomArrow{o-o}, \CustomArrow{o->}$.
The ``$\CustomArrowShortest{-o}$'' mark represents undetermined edge mark.
In other words, ``$\CustomArrowShortest{-o}$'' can be either a tail ``$\CustomArrowShortest{-}$'' or a head ``$\CustomArrowShortest{->}$''.
$\CustomArrow{<->}$ shows that there are hidden confounders between the two variables on either side of the arrow.
$X \CustomArrow{o->} Y$ implies that either $X$ causes $Y$ or there are hidden confounders between both variables.
$X\CustomArrow{o-o}Y$ might be: $X$ causes $Y$, $Y$ causes $X$, there are common hidden confounders between both variables,
$X$ causes $Y$ and there are hidden confounders between both variables, or $Y$ causes $X$ and there are hidden confounders between both variables.
As in the first step of the PC algorithm (Algorithm~\ref{alg:PC_one}), FCI relies on statistical independence tests to infer the skeleton of the graph.
It is in the second step (Algorithm~\ref{alg:FCI-step2}) that FCI deviates from the PC algorithm.

After orienting all the edges in the graph as $\CustomArrow{o-o}$, the algorithm starts by an orientation rule in order to detect the v-structures in the graph (lines~\ref{alg3-line2}-\ref{alg3-line6}).

Another rule specific to FCI is the detection of Y-structures. Four variables define a Y-structure when: $C1 \rightarrow X \leftarrow C2$ and $X \rightarrow Y$. Within the Y-structure, both $C1$ and $C2$ are independent of $Y$ conditional on $X$. This conditional independence helps exclude the possibility of a latent confounder between $X$ and $Y$. That is, when FCI detects a Y-structure in the graph, no latent confounders exist between $X$ and $Y$; otherwise, FCI assumes that possibly latent confounders exist~\cite{mani2012theoretical}.

Afterwards, FCI applies four additional rules to direct the remaining edges.
Those rules are described in Algorithm~\ref{alg:FCI-step2} as follows: rule $1$ in lines~\ref{alg3-line8}-\ref{alg3-line10}, rule $2$ in lines~\ref{alg3-line11}-\ref{alg3-line13}, rule $3$ in lines~\ref{alg3-line14}-\ref{alg3-line17} and rule $4$ in lines~\ref{alg3-line17}-\ref{alg3-line23}.
Rules $1-3$ are similar to the inference rules used in the context of DAGs with a slight generalization.
However, rule $4$ is more specific to causal graphs with bidirectional edges witch considers the presence (absence) of discriminating paths for variables in the graph.
A path $\pi=\angles{X,...,W,V,Y}$ is said to be discriminating\footnote{The term discriminating path is not related to fairness.}
for $V$ if: $(1)$ $\pi$ has at least three edges, $(2)$ $V$ is a non-endpoint node on $\pi$ and is adjacent to $Y$ on $\pi$, and $(3)$ $Y \notin adj(X, G)$, and every node between $X$ and $V$ is a collider on $\pi$ and is a parent of $Y$.
Note that rules $1-3$ were initially proposed by Spirtes et al.
in ~\cite{spirtes1999algorithm}.
Zhang augmented FCI algorithm by providing additional orientation rules for completeness~\cite{zhang2008completeness}.
The star symbol ``$\CustomArrowShortest{-{Rays[n=8]}}$'' does not appear in the PAG generated by the algorithm but is used as a meta-symbol to represent any of the three kind of edge ends: ``$\CustomArrowShortest{-}$'', ``$\CustomArrowShortest{->}$'' or ``$\CustomArrowShortest{-o}$''.
Note that FCI is limited to several thousand variables.

\begin{algorithm}[ht]
\caption{\small Step 2 of the FCI algorithm: orienting the edges}\label{alg:FCI-step2}
\small
\SetAlgoLined
\KwIn{The undirected graph $G$.}
\KwOut{Partial ancestral graph $PAG$.}
orient each edge in $G$ as $\CustomArrow{o-o}$ \label{alg3-line1}\\
\For{each unshielded triple of vertices $(X,C,Y)$ in $G$ \label{alg3-line2}}{
    \If{$C \notin \textbf{Z}$ (separating set of $(X,Y)$) \label{alg3-line3}}{
    orient the edges $X\CustomArrow{{Rays[n=8]}-{Rays[n=8]}}C\CustomArrow{{Rays[n=8]}-{Rays[n=8]}}Y$ as $X\CustomArrow{{Rays[n=8]}->} C\CustomArrow{<-{Rays[n=8]}}Y$ \label{alg3-line4}\\
    }  \label{alg3-line5}
} \label{alg3-line6}
\Repeat{none of the above rules applies \label{alg3-line7}}
{
    \If{$X \CustomArrow{{Rays[n=8]}->} C \CustomArrow{o-{Rays[n=8]}} Y$, and $Y \notin adj(X, G)$ \label{alg3-line8}}
    {
    orient the triple as $X \CustomArrow{{Rays[n=8]}->} C \CustomArrow{->} Y$    \tcp*[r]{Rule 1} \label{alg3-line9}
    }  \label{alg3-line10}

    \If{$X \CustomArrow{->} C \CustomArrow{{Rays[n=8]}->} Y$ or $X \CustomArrow{{Rays[n=8]}->} C \CustomArrow{->} Y$, and $X \CustomArrow{{Rays[n=8]}-o} C$ \label{alg3-line11}}
    {
    orient $X \CustomArrow{{Rays[n=8]}-o} C$ as $X \CustomArrow{{Rays[n=8]}->} C$ \tcp*[r]{Rule 2}\label{alg3-line12}
    } \label{alg3-line13}

    \If{$X \CustomArrow{{Rays[n=8]}->} C *\CustomArrow{<-{Rays[n=8]}} Y$, $X \CustomArrow{{Rays[n=8]}-o} D \CustomArrow{o-{Rays[n=8]}} Y$, $Y \notin adj(X, G)$, and $D \CustomArrow{{Rays[n=8]}-o} C$ \label{alg3-line14}}
    {
    orient $D \CustomArrow{{Rays[n=8]}-o} C$ as $D \CustomArrow{{Rays[n=8]}->} C$ \tcp*[r]{Rule 3}\label{alg3-line15}
    } \label{alg3-line16}

    \If{ $\pi = \angles{D, ..., X,C,Y}$ is a discriminating path between $D$ and $Y$ for $C$, and $C \CustomArrow{{Rays[n=8]}-o} Y$\label{alg3-line17} }
    {
        \eIf {$C \notin Sepset(D,Y)$ \label{alg3-line20}}
        {
        orient $C \CustomArrow{o-{Rays[n=8]}} Y$ as $C \CustomArrow{->} Y$ \tcp*[r]{Rule 4} \label{alg3-line19}
        }{orient the triple $\angles{X,C,Y}$ as $X \CustomArrow{<->} C \CustomArrow{<->} Y$}\label{alg3-line21}
        
    } \label{alg3-line23}
} 
\end{algorithm}


\subsection{GES algorithm}
\label{subsec:GES}
\begingroup 
\def\P{\mathcal{P}}
\def\BIC{\mathsf{BIC}}
\def\data{\mathcal{D}}
\def\score{\mathsf{score}}
\def\phase{\mathsf{phase}}
\def\forward{\mathsf{forward}}
\def\backward{\mathsf{backward}}
\def\True{\mathsf{True}}
\def\neighboringStates{\mathsf{neighbors}}

Greedy Equivalence Search\cite{chickering2002optimal} (GES) is a score-based algorithm that, unlike PC and FCI, starts with a completely disconnected graph and then adds, deletes and modifies edges in a certain order until reaching the causal model that maximizes a regularized performance score, called \emph{BIC score}, that stands for Bayes Information Criterion and is described in detail in Section~\ref{subsec:BIC}.
The pseudo-code of GES is shown in Algorithm~\ref{alg:GES}.

\begin{algorithm}[ht]
  \caption{Pseudo-code of GES algorithm.}\label{alg:GES}
  \small
  \SetAlgoLined
  \KwIn{Dataset $\data$ of $K$ variables and $N$ samples.\\}
  \KwOut{Completed PDAG $\P$ that maximizes $\BIC$ score.}
  $\P \gets $ disconnected completed PDAG of $K$ nodes\\
  $\score \gets 0$\\
  \For{$\phase \in [\forward, \backward]$}{
    \While{$\True$}{
      $\neighboringStates\gets \set{\P' : \P\to\P' \text{ is a }\phase\text{-transition}}$\\
      \If{$|\neighboringStates|=0$}{
        $\mathbf{break}$\\
      }
      $\P' \gets \argmax_{\P'\in\neighboringStates}{\Delta\BIC(\P, \P', \data)}$\\
      $\Delta\score \gets \Delta\BIC(\P, \P', \data)$\\
      \If{$\Delta\score < 0$}{
        $\mathbf{break}$\\
      }
      $\P \gets \P'$\\
      Add $\Delta\score$ to $\score$\\
    }
  }
  \Return{$\P, \score$}
\end{algorithm}

A first remark about GES, evident in Algorithm~\ref{alg:GES}, is that its output is not necessarily a directed acyclic graph (DAG), but a  completed partially directed acyclic graph (CPDAG) which represents a Markov equivalence class of causal DAGs.

GES consists of searching over an abstract search space (graph) of states and transitions.
Each state is an equivalence class of DAGs, all of which happen to have the same BIC score, and is represented in the form of a CPDAG.
The search objective is the state that maximizes BIC score, hence, the abstract output of GES is an equivalence class of DAGs.

The states of the search space, i.e. the equivalence classes, are determined by the following equivalence relation: two DAGs are \emph{equivalent} when (i) they have the same set of v-structures and (ii) they can generate the same family of distributions by modifying the model parameters.
When the variables are all categorical or all continuous following a multivariate Gaussian distribution, these two conditions are equivalent and can be therefore reduced into the first one.
Each equivalence class (state) is represented uniquely as a CPDAG.
More precisely, the CPDAG $\P$ corresponds to the set of DAGs that agree with the directed edges of $\P$ and assume any direction for the undirected edges without forming a cycle.
In Algorithm~\ref{alg:GES}, the current state is represented with the variable $\P$.

The transitions of the search space are given by the following rule: a transition from a state to another exists if and only if there are two DAGs, one on the equivalence class of each state, that differ only in the addition or removal of exactly one edge.
Hence, the transitions can be partitioned into forward transitions that increase the number of edges of the completed PDAG by one, and backward transitions that decrease it by one.
Of all the forward and backward transitions of a given state, we will call \emph{best forward (or backward) transition} to the forward (backward) transition that reaches the neighboring state with maximal BIC score.
No special rule exists to resolve ties for this arg-max operation, although they are extremely unlikely to occur in practice.
In Algorithm~\ref{alg:GES}, the neighboring states for the state $\P$ are represented with the variable $\neighboringStates$.

The change in BIC score after following a transition can be computed using a simple rule instead of fitting the whole global model on both states.
The BIC score can be decomposed as the sum of the local BIC scores of each node and its directed and undirected parents.
Since there is a unique variable $Y$ whose parents change during a transition, then the global BIC score difference corresponds exactly with the local BIC score difference for the model of $Y$ and its parents. 
This optimization corresponds to $\Delta\BIC(\P, \P', \data)$ in Algorithm~\ref*{alg:GES}.

The greedy strategy of GES consists of repeatedly following the best forward transition at each state that it encounters until a local maximum is reached, i.e. until the next state reduces the BIC score, and then, analogously, repeatedly following the best backward transition until a local maximum is reached.
These two consecutive algorithms that form GES are called the forward and backward phases.
Notice from the definition of BIC score that during the forward phase, the regularization term plays no role when deciding which transition is the best, but it does determine when to stop.

The distinctive essential feature of GES is that its greedy technique, which prunes the search space dramatically, is nevertheless guaranteed to find the optimal state of the whole space, provided that the assumptions listed in Section~\ref{subsec:applicability-GES} are met.

The computation of the neighboring states of a given state (for both phases) is carried out by finding edges $X \to Y$ that can be added (or removed) in such a way that the resulting PDAG can be \emph{extended}, i.e. transformed into a DAG by smartly deciding the direction of the undirected edges.
The extension algorithm is described in~\cite{dor1992simple}.

Once the DAG is computed, it is \emph{completed} to obtain the CPDAG that represents the equivalence class containing it.
The completion algorithm is simple to implement from the definition of a completed PDAG and is explained in~\cite{chickering2002optimal}.

\endgroup

\subsection{LiNGAM algorithm}
\label{subsec:LiNGAM}

LiNGAM is an algorithm based on causal asymmetries that, unlike the previously discussed algorithms, yields a unique directed graph (DAG) and corresponding parameters.
However, the stronger causal discovery power comes at the expense of more assumptions that have to be satisfied.
LiNGAM requires linearity and non-gaussianity of the variables to recover causal directions and learn functional relationships~\cite{shimizu2006linear}.
The approach is closely related to Independent Component Analysis (ICA) algorithm as they both  base their premises in the Darmois-Skitovic theorem~\cite{oja2000independent}.

\begin{figure}[htp]
 
    \includegraphics[width=0.5\textwidth]{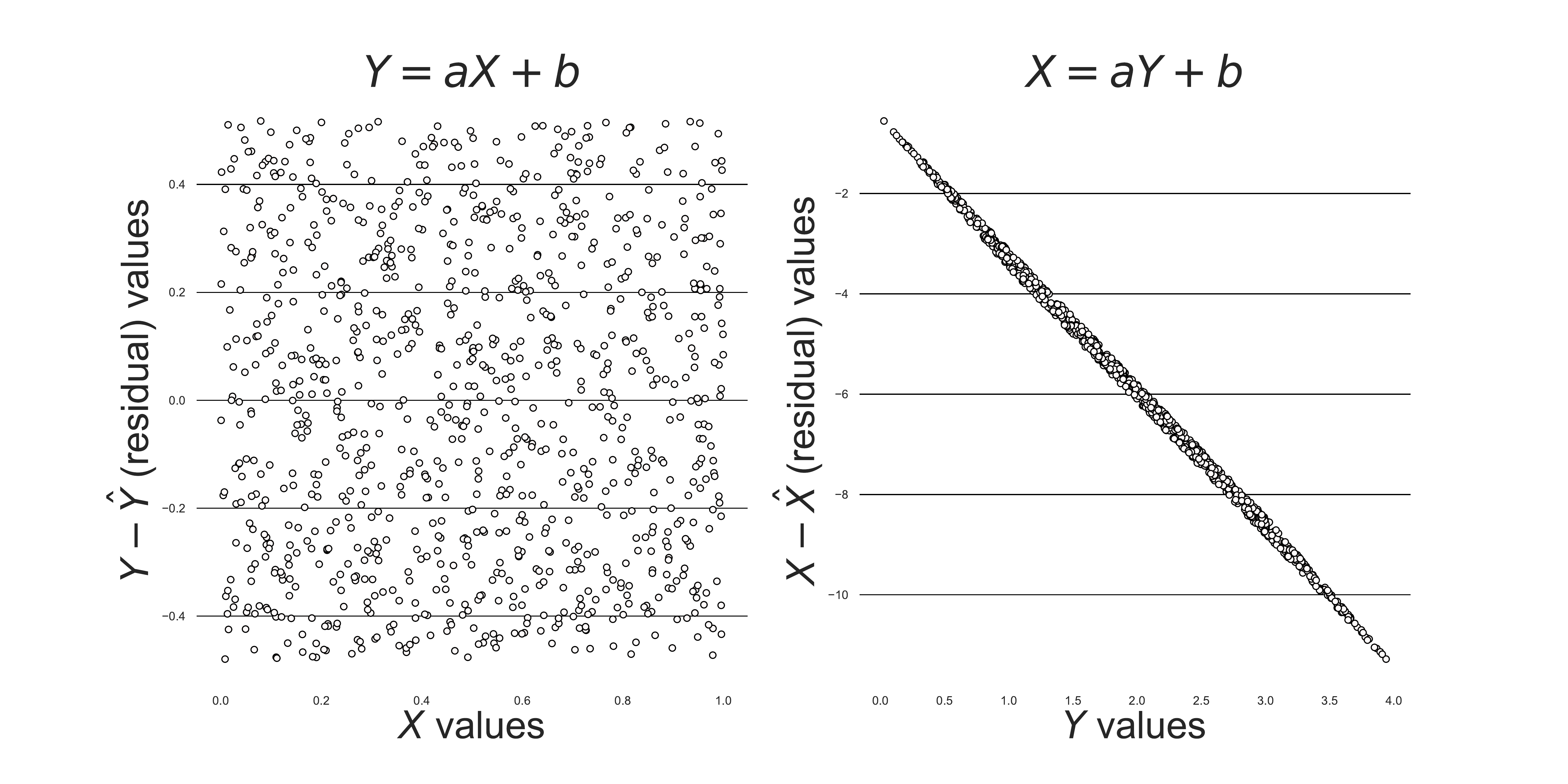}
    \caption{Plotting predicting variable against its residuals in case where causal direction is correct (left) and reversed (right).}
    \label{fig:lingam}
\end{figure}

The theorem implies that fitting backward model (trying to regress the cause on the effect) to the data would result in dependence between cause $X$ and the residuals of the effect $Y$, allowing to correct the causal direction. A residual of regressing variable $X_i$ on $X_j$ is defined as:
\begin{equation}
  r_i^{(j)}= X_i - \frac{cov(X_i,X_j)}{var(X_j)}X_j.\label{eq:residual}
\end{equation}
Figure~\ref{fig:lingam} shows the visual examination of the dependency between the residuals and the variable used for prediction in the case where causal direction is correct ($\hat{Y}=a_1X+b_1$) and where the model is fitted backwards ($\hat{X}=a_2Y+b_2$).

The variable $Y$ is equal to linear combination of $X$ and uniform noise.
It is easy to observe a strong dependence between the residuals of an effect variable and the predictor variable, when the causal direction is reversed.

The independence between the cause variable $X_j$ and the residuals $r_1^(j), r_2^(j),...$ of the effect variables is measured as follows:
$$T(X_j;\mathbf{V}) = \sum_{i \in \mathbf{V}, i\neq j} I_M(X_j, r_i^{(j)})$$
Where $X_j$ is a cause variable used in the regression, $\mathbf{V}$ is the set of all the variables (except $X_j$) regressed on $X_j$,  $r_i^{(j)}$ is the residual of $X_i$ when regressed on $X_j$ and $I_M(X_j, r_i^{(j)})$ is Mutual Information between $X_j$ and the residual $r_i^{(j)}$.
Mutual information is usually used as a metric $I$ for independence between $X_j$ and residuals $r_i^{(j)}$~\cite{hyvarinen2013pairwise}, although other metrics have been proposed~\cite{shimizu2014lingam}.

The first algorithmic implementation of LiNGAM, ICA-LiNGAM, was based on independent component analysis~\cite{shimizu2006linear}.
In this paper, we describe a subsequent improvement DirectLiNGAM~\cite{shimizu2011directlingam} which, in contrast to the ICA version, is not based on iterative search and therefore does not require initial guess or similar parameters and is guaranteed to converge to the right solution.
The DirectLiNGAM algorithm implementation learns the causal graph in two steps.
First, it finds the causal order of the variables: an ordered list, where the first is the exogenous variable (has no parents in the graph), second is the child of the exogenous variable, that has the most descendants etc. 
Next, the causal order is used to compute the adjacency matrix that specifies the strength of the connections. Specifically, starting from the end of the list, each variable is regressed on all the others that comes before it in the causal order (potential parents).

\SetKwRepeat{Do}{do}{while}

\begin{algorithm}[t]
  \caption{Pseudo-code of the main body of Direct LiNGAM algorithm.}\label{alg:Direct_LiNGAM}
  \small
  \SetAlgoLined
  \KwIn{$p$-dimensional vector of variables \textbf{x},$p$-dimensional set of variable indices $U$ and $p$ x $n$ observed data matrix $\textbf{X}$
  }
  \KwOut{Adjacency matrix $\mathcal{B}$ with parameters $\theta$ }
  Causal order list $K:=\emptyset$\\
  Residual matrix $\textbf{R}$
 
  \Do{ until $p$ variable indices are appended to $K$}{
   
    $U:=  \hat{U} \symbol{92} x_m$\\
    $\textbf{X}:=\textbf{R}^{(m)}$\\ \label{lingam_step3}
    
    \For{all $i,j\in U \symbol{92} K (i\neq j)$ }{
        Compute $\textbf{r}^{(j)}$ and the residual data matrix  $\textbf{R}^{(j)}$\\ \label{lingam_step1}
        $x_m=\operatorname*{argmin}_{j\in U \symbol{92} K}T(x_j;U)$, where $ T(x_j;U) = \sum_{i \in U, i\neq j} I(x_j;r_{i\in U \symbol{92} K}^j)$\\ \label{lingam_step2} 
         $x_m$ append to the end of $K$ \\
      
    }
  }

  Adjacency matrix $\mathbf{B}:= [0]_{pxp} $\\
  \For{$i \in [n-1, n-2, ..., 1]$}{
    $j = K_i$ \Comment{target}
    $I = [K_1,K_2,...,K_{i-1}]$ \Comment{potential parents}
    $\mathbf{B_{j, :}} = \text{Linear-Regression}(\mathbf{X_{j}}, \mathbf{X_{I}})$ \\ \label{lingam_step4} 
  
  }
  \Return{$\mathbf{B}$}
\end{algorithm}

Algorithm~\ref{alg:Direct_LiNGAM} shows the pseudo-code of Direct LiNGAM.
First, the algorithm computes residuals for all the pairs of variables in the dataset (line \ref{lingam_step1}). Next, the independence tests are applied to the variables and corresponding residuals to find the one that is most independent of all the resulting residuals (line \ref{lingam_step2}).
The independence metric, usually Mutual Information is used to compute the dependence score.
The most independent variable is considered to be an exogenous variable $x_m$ (the node that has no parents in the graph).
Once the first variable $x_m$ is appended to the causal order list.
Next, $x_m$, which is already ordered is removed from the data and all the other $x_i$ in the data are being substituted by residuals $r_i^m$ (resulting when regressing $x_i$ on $x_m$) that are shown to follow the same ordering and directions as the original variables.

This step is repeated for each iteration, so the new residuals are now computed from the residuals $r_i^m$ using the same equation \ref{eq:residual}.
The use of the residuals instead of the original variables allows to remove the effect of already identified parent and make the discovery process more efficient.
The process is repeated until all the variables are ordered.


After the causal order is defined, the connection coefficients are estimated by applying the covariance-based regression (least-squares or maximum-likelihood approach) on the original variables following the established ordering (line \ref{lingam_step4}).
Namely, starting from the end of the causal order list $K$ every variable is regressed on all the previous ones.
The resulting adjacency matrix specifies if there exist and how strong are the relations between the variable and its potential parents in the causal order list.
The connection weight with absolute value equal to zero or smaller than a selected threshold indicates no link between the effect variable and a potential parent. In this way a pattern of connections including confounding and colliding structures is identified resulting in the unique causal graph.



DirectLingam algorithm allows incorporating prior knowledge in the form of a matrix specifying forbidden and enforced edges between the variables that are known to be true.
The use of the prior knowledge improves speed and accuracy of the algorithm even though in theory external information is not required.
The pruning of the redundant edges of the graph favoring the simpler model can be done by repeatedly applying a Adaptive Lasso method which excludes the weaker connections.
DirectLiNGAM has been tested on real world sociology data and showed a better performance than PC or GES algorithms, even though the assumptions of the model were not guaranteed to be satisfied~\cite{shimizu2011directlingam}.

The causal asymmetries framework to identify unique causal structures gave rise to several algorithms that extend the applicability of Direct LiNGAM. Those include algorithm for estimating models with hidden variables~\cite{hoyer2008estimation}, time series data~\cite{hyvarinen2010estimation}, mixture of Gaussian and non Gaussian variables~\cite{hoyer2012causal}, approach that allows to drop the acyclicity assumption~\cite{lacerda2012discovering} and model for nonlinear functional relationships~\cite{zhang2012identifiability,hoyer2008nonlinear}.

\subsection{SBCN}
\label{subsec:SBCN}

A Suppes-Bayes Causal Network (SBCN)~\cite{bonchi2017exposing} is a different type of causal graph that is used specifically for fairness assessment purposes.
SBCN deviates from the causal graphs used above in three aspects.
First, vertices in an SBCN correspond to Bernoulli variables with binary values.
For example, $\langle$ Gender = female $\rangle$ and $\langle$ Gender = male $\rangle$ correspond to two different vertices.
Second, causal relations between vertices follow the Suppes's definition of causality~\cite{hitchcock2002probabilistic,suppes1973probabilistic} (different from the typical definition of causality~\cite{pearl2009causality})
which requires temporal priority and probability raising.
For example, a node $a$ is a cause of a node $y$ ($a \rightarrow y$) if and only if, $a$ occurs before $y$ (temporal priority) and the cause $a$ raises the probability of the effect $y$, that is, $\pr(y|a) > \pr(y|\neg a)$ (probability raising).
Third, every edge (causal relation) is assigned a weight corresponding to the confidence score.
The weight is simply the extent of the probability raising ($W(a,y) = \pr(y|a) - \pr(y|\neg a)$).

Discovering the SBCN structure from the data is a hybrid approach using constraint-based as well as score-based ideas.
Algorithm~\ref{alg:SBCN} illustrates how the SBCN is learned from the data.
The algorithm takes two inputs.
The first input is a dataset $D$ of $n$ samples
and a set $V$ of Bernoulli variables\footnote{SBCN require that all variables are either categorical or binary.
For continuous data, variables should be discretized.} of the type \anglesSBCN{variable}{value}.
The second input is a partial temporal order $r$ of the variables.
The algorithm outputs an SBCN $= (V,E^*,W)$ with $E^*$ maximizing the BIC score:
\begin{align}
        E^* = \underset{E \subseteq E', G=(V,E)}{\arg\max} (LL(D \mid G) - \frac{\log (n)}{2} dim(G)) \nonumber
\end{align}
where $E'$ is the set of edges satisfying the Suppes' definition of causality, $LL$ is the log likelihood function, $n$ is the number of samples in $D$ and $dim(G)$ is the number of edges in $G$.


\begin{algorithm}
\caption{\small SBCN}\label{alg:SBCN}
\small
\SetAlgoLined
\KwIn{Dataset $\mathcal{D}$ with a set of Bernoulli variables $\textbf{V}$, and a partial order $r$ of $V$.}
\KwOut{SBCN $= (V,E^*,W)$.}
\For{all pairs $(x,y) \in V$ \label{alg5-line1}}{
    \If{$r(y) \leq r(x)$ \textbf{and} $\mathbb{P} (x \mid y) >  \mathbb{P} (x \mid \neg y)$ \label{alg5-line2}}{
    add the edge $(x,y)$ to SBCN \label{alg5-line3}
    } \label{alg5-line4}
} \label{alg5-line5}
Consider $G(V,E^*,W)_{fit} = \emptyset $ \label{alg5-line6}\\
\While{$!StoppingCriterion()$ \label{alg5-line7}}{
let $G(V,E^*,W)_{neighbors}$ be the neighbor solutions of $G(V,E^*,W)_{fit}$ \label{alg5-line8}\\
remove from $G(V,E^*,W)_{neighbors}$ any solution whose edges are not included in SBCN\label{alg5-line9}\\
consider a random solution $G_{current}$ in $G(V,E^*,W)_{neighbors}$ \label{alg5-line10}\\
    \If{$score_{BIC}(D,G_{current}) > score_{BIC}(D,G_{fit})$ \label{alg5-line11}}{
    $G_{fit} = G_{current}$ \label{alg5-line12}\\
    $\forall edge (x,y)$ of $G_{fit}$, $W(x,y) = \mathbb{P} (x \mid y) -  \mathbb{P} (x \mid \neg y)$ \label{alg5-line13} \\
    } \label{alg5-line14}

} \label{alg5-line15}
SBCN = $G_{fit}$ \label{alg5-line16}\\
return SBCN \label{alg5-line17}
\end{algorithm}

The first loop (lines~\ref{alg5-line1}-\ref{alg5-line3}) adds only edges that satisfy the Suppes' definition of causality.
The second loop (lines~\ref{alg5-line7}-\ref{alg5-line16}) is a hill climbing search procedure for the SBCN that maximizes the BIC score.
It starts with an empty graph, then browse through randomly chosen neighbors ($G(V,E^*,W)_{neighbors}$) looking for an SBCN that improves the BIC score.
Neighbors of a particular SBCN are the SBCNs obtained by adding or removing at most one edge.
The search resumes from the SBCN that improves the BIC score.
The algorithm stops ($!StoppingCriterion$ in line~\ref{alg5-line7}) when either none of the neighbors of the current SBCN improves the score or enough number of iterations is performed.


Measuring fairness/discrimination using the generated SBCN is based on random-walks.
That is, based on the weighted edges between vertices, it is possible to measure several types of fairness notions (e.g., group and individual discrimination, direct and indirect discrimination, etc.).
For instance, group discrimination 
is measured using a number $n$ of random walks that begin from a node $v$ (e.g., \anglesSBCN{gender}{female}) and reach the node corresponding to the negative decision (e.g., \anglesSBCN{decision}{not\;hired}).
This corresponds to the discrimination score $ds^-$:
\begin{align}
\label{eq:ds-}
    ds^-(v) = \frac{rw_{v \rightarrow \delta^-}}{n} 
\end{align}
where $\delta^- \in V$ represents the node of the negative decision (e.g., $not\;hired$) and $rw_{v \rightarrow \delta^-}$ represents the number of random walks starting at vertex $v$ and reaching $\delta^-$ earlier than $\delta^+$ (node of the positive decision e.g. $hired$).
Note that the choice of a path in a random walk is based on the weights of the out-goings edges.
Being at node $x$, the probability of moving to node $y$ rather than another neighbor node is:
\begin{align}
    \label{eq:rw}
    p(x,y) = \frac{W(x,y)}{\sum_{z\in \mathsf{outgoing}(x)}W(x,z)}
\end{align}

$\mathsf{outgoing}(x)$ represents the set of outgoing edges from $x$.
In case a random walk reaches a node with no outgoing edges before attaining the decision node, it is restarted from the starting node. 

SBCN is used in a similar way to compute the favoritism (positive discrimination), indirect, genuine, individual, and sub-group discrimination.

\section{Independence tests and BIC score}
\label{sec:independence}

The metrics used for causal discovery by the above algorithms can be classified as information based independence measures (eg. Mutual Information), statistical conditional independence tests (eg. Fisher Z Test), likelihood based conditional independence tests (eg. G likelihood-ratio) and likelihood based scores (eg. BIC).
The applicability of most tests depends on the data distribution and the functional relationships between the variables even for the algorithms that do not make explicit parametric assumptions (e.g. PC).
Traditional conditional independence methods impose parametric assumptions on continuous data.
The most common assumption is linear relations with additive Gaussian errors~\cite{zhang2012kernel}.
Thus, the data distribution and functional relationship has to be considered when choosing appropriate independence tests for the dataset.
Conditional independence tests are usually used by constraint based algorithms (PC, FCI).
Likelihood based scores do not measure the independence between the variables directly, but rather the \textit{likelihood} of a data if we assume the links between the variables and their directions (GES). In a pairwise setting the unconditional independence is measured between the pairs of variables and compared to determined the most dependent/independent pair (LiNGAM).
In the following, we provide definitions and applicability recommendations for the most commonly used independence tests and scores. The list is not exhaustive and many case-specific or speed-optimized alternatives for the following metrics exist.




\textit{\textbf{Conditional Pearson Correlation Independence Test~\cite{tsagris2019bayesian}}}:
is a measure of a linear relationship between two variables.

Fisher Z transformation of Pearson correlation coefficient $r$ is applied to make it normally distributed if it cannot statistically be distinguished from zero. Using Fisher Z transformation of Pearson correlation for independence testing requires linearity and Gaussianity of the data. This test is often used by PC and FCI.

\begin{equation}
    T_{\text{Pearson}} = \frac{1}{2}\left|\log\frac{1+r_{X,Y|Z}}{1-r_{X,Y|Z}}\right|\sqrt{n-|\textbf{Z}|-3}
\end{equation}
Where, $n$ is the sample size, $|\textbf{Z}|$ is the number of conditioning variables in the set $\textbf{Z}$ and $r_{X,Y|Z}$ is the partial Pearson Correlation of $X$ and $Y$ conditioning on $\textbf{Z}$, which has a value between $-1$ and $1$, with a value of $-1$ meaning a total negative linear correlation, $0$ being no correlation, and $1$ implying a total positive correlation.

\textit{\textbf{K-CI Test~\cite{zhang2012kernel}}}: is a kernel-based non-parametric conditional independence test for continuous data that does not require discretization of the continuous variables values.
The test does not make assumptions on distribution and functional relationship.
This test is typically recommended for (1) faster computation, (2) when the conditioning set is large or (3) when the dataset is small.

\textit{\textbf{The G Likelihood-Ratio Test~\cite{tsagris2019bayesian}}}: is a test for categorical data based on the ratio of observed and expected frequencies if the variables were independent.
It is related to chi-squared test statistic.
It is defined as follows: 

\begin{equation}
    G^2 = 2\sum_{k}\sum_{i,j} O_{i,j|k}\log\frac{O_{i,j|k}}{E_{i,j|k}}
\end{equation}
Where $O_{i,j|k}$ are the observed frequencies of the $x_i$, $x_j$ in the $k$'th set of conditioning values, and $E_{i,j|k}$ are the corresponding expected frequencies.

\textit{\textbf{Mutual Information Test}}: is an independence test
applied pairwise between the variable and the residual in most implimentations of the LiNGAM algorithm.
The difference in mutual information for two pairs determines which variable is more independent from its residuals.
The mutual information between the variable and its residuals in both directions is computed by one-dimensional entropy ($H$) \cite{hyvarinen2013pairwise}:

\def\std{\mathsf{std}}
\begin{multline}
    I(x_1, r_2^{(1)}) -  I(x_2, r_1^{(2)}) = H(x_1)+H\left(\frac{r_2^{(1)}}{\std(r_2^{(1)})}\right)\\-\left[H(x_2)+H \left(\frac{r_1^{(2)}}{\std(r_1^{(2)})}\right) \right]
\end{multline}
Where $r_2^{(1)}$ is the residual of $x_2$ when regressed on $x_1$, 
$r_1^{(2)}$ is the residual of $x_1$ when regressed on $x_2$,  and $\std(r_2^{(1)})$, $\std(r_1^{(2)})$ are standard deviations of the residual.
A kernel based independence measure is also used in \cite{shimizu2011directlingam,bach2002kernel} to speed up the computation.

\textit{\textbf{Conditional Gaussian BIC Test\cite{andrews2018scoring}}}:
is a likelihood ratio test based on the conditional Gaussian likelihood function and a BIC score (see \ref{def:BIC}).
This test is convenient to use with datasets that include a mixture of continuous and discrete variables.
It is assumed that the continuous variables are Gaussian conditional on each combination of values for the discrete variables, though it will work fairly well even if that assumption does not hold strictly.
This test is suitable for any constraint-based algorithm (PC, FCI, etc.).

\textbf{\textit{BIC Score}}:\label{subsec:BIC}

The Bayes Information Criterion\cite{schwarz1978estimating} (BIC) is a likelihood based model selection criterion. 
It is a regularized performance score that summarizes the quality of the model into a single real number.
It is maximal when the statistical model is both simple and correct, that is, when it has few parameters and a high likelihood for the given data.

\begin{definition}(BIC score)\label{def:BIC}
Let $\X$ be an arbitrary domain endowed with a metric and a measure, e.g. euclidean space with the continuous or the discrete measure; let $\hat p_\Theta$ be a statistical model of dimension $k$ consisting of a collection $\set{\hat p_\theta : \theta\in\Theta\subseteq \RR^k}$ of dominated densities $\hat p_\theta$; and let $s=\lr{x_i}_{i=1}^N$ be a dataset of $N$ i.i.d. samples $x_i\in\X$, so that  $\hat p_\theta(s) = \Pi_{i=1}^N \hat p_{\theta}(x_i)$.
The \emph{BIC score} is defined as
\begin{equation}
      \mathrm{BIC}(\hat p_\Theta, s) \eqdef
  \ln \hat p_{\hat \theta}(s) - \frac{k}{2} \ln\lr{N},
\end{equation}
where $\hat \theta \eqdef \argmax_{\theta\in\Theta} \hat p_{\theta}(s)$ is the maximum likelihood estimator of the parameters.
\end{definition}

Therefore, the value of the BIC score (Definition~\ref{def:BIC}) consists of the sum between the log-likelihood of the model (performance) and a weighted regularization term that penalizes models with large number of tunable parameters.
For causal discovery, the BIC score is used by the GES algorithm.


\section{Applicability}
\label{sec:applicability}

\def\numOfVars{{|V|}}

Not all algorithms can be used to discover causal relations in a given observed data.
The type of the data variables (e.g. continuous vs categorical), the type of the structural functions between variables (e.g. linear vs non-linear), and the distribution of the noise (e.g. gaussian vs uniform) are used to tell which algorithms can/cannot be used.

\subsection{PC and FCI}
\label{subsec:applicability-pc-fci}
PC algorithm requires three assumptions to hold: causal markov condition (Equation~\ref{eq:markov}), causal faithfulness (conditional independence in the data reflects d-separation constraints in the graph), and causal sufficiency (no hidden confounders).
Initially, PC was designed to take as input either entirely continuous or entirely discrete data.
However current implementations allow the use of mixed data via Conditional Gaussian test. 
 
Because FCI is a variant of PC, the same assumptions hold for FCI, except causal sufficiency, which allows FCI to work in presence of hidden confounders.


The conditional independence tests used to discover the skeleton of the graph for both PC and FCI have an $\alpha$ value (input of Algorithm~\ref{alg:PC_one}) for rejecting the null hypothesis, which is always a hypothesis of independence or conditional independence.
For continuous variables, PC uses conditional Pearson correlation~\cite{tsagris2019bayesian} (if the functional relations are linear and the data distribution is normal) or K-CI~\cite{zhang2012kernel} (if no assumptions are made on the type of functions).
For categorical variables, PC uses either a chi square or G likelihood-ratio.
The default value of $\alpha$ is $0.01$. However, for categorical data, using a value of $0.05$ is recommended\footnote{https://cmu-phil.github.io/tetrad/manual/}.

In addition to Tetrad, PC and FCI are also implemented in \textsf{pcalg}\cite{pcalg,hauser2012gies} and \textsf{bnlearn}\cite{scutari2010learning} - two R packages.
The two algorithms are also implemented in a Python version of \textsf{pcalg}\footnote{https://github.com/chufangao/pcalg.}. 

Most of PC processing time is spent in the first phase of skeleton identification. The obvious strategy is to test all possible conditional independence relations for each pair of variables ($X$ and $Y$).
A naive implementation will go through all possible subset of variables, that is $2^\numOfVars$ subset where $\numOfVars$ is the number of variables.
However, in practice, only subsets composed of adjacent variables to $X$ and $Y$ need to be considered.
Overall, the complexity of PC is $\mathcal{O}(\numOfVars^{d_{max}})$ where $d_{max}$ denotes the maximal node degree in the graph~\cite{sondhi2019reduced}.

The degree of a node $X$ is the number of nodes adjacent to $X$. Hence the efficiency of PC depends heavily on the number of variables but most importantly on the sparsity of the causal relations.


\subsection{GES}
\label{subsec:applicability-GES}

GES makes the same assumptions as PC and FCI, namely, causal markov condition, causal faithfulness, and causal sufficiency. For the type of data, the formulation of GES is very general and hence it works for categorical, continuous, and mixed data.
However, most of the theory about the statistical guarantees of the algorithm assumes joint gaussianity of the continuous variables. For instance, Chickering, in the original GES paper~\cite{chickering2002optimal},
defines GES for datasets in which all the variables are categorical (multinomial\footnote{
  The distribution of $N$ i.i.d. categorical samples is multinomial.
}) or all the variables are continuous and follow a joint Gaussian distribution.

We consider the more general case of mixed data~\cite{andrews2018scoring} because it captures the assumptions used in both continuous and discrete cases.
The conditional gaussian score calculates conditional gaussian mixtures using the ratios of joint distributions and makes the following assumptions:
\begin{itemize}
  \item[A1.] The continuous data were generated from a single joint (multivariate) Gaussian mixture where each Gaussian component exists for a particular setting of the discrete variables.
  \item[A2.] The instances in the data are independent and identically distributed (iid).
  \item[A3.] All Gaussian mixtures are approximately Gaussian.
\end{itemize}
It is also crucial in practice to have enough samples when conditioning on several categorical variables simultaneously, especially when these variables are parents of the same variable.
This and A2 are the most relevant requirements in the all-discrete case.
Regarding continuous variables, A1 implies implicitly linearity and gaussianity of the residuals because of the nature of each Gaussian component; this is the most relevant assumption in the all-continuous case.
It is important to mention that in the existing standard implementations of GES, such as Tetrad's \textsf{fges}\cite{ramsey2018tetrad} (written in Java), the \textsf{pcalg}\cite{pcalg,hauser2012gies} library for R (written in C++ underneath), and other Python implementations~\cite{gamella2021python,kalainathan2019causal}, the predictive models are pre-configured to linear regression for continuous variables.
These assumptions may sound too strict, but the empirical evidence of several articles shows that GES performs well even when this assumptions do not hold exactly~\cite{andrews2018scoring, chickering2002optimal, hauser2012gies}.

In terms of complexity, the runtime of the worst-case scenario is upper bounded by $O(\numOfVars^4 k \cdot \max(\numOfVars\, k^2,\, n))$, where $n$ is the number of samples, $\numOfVars$ is the number of variables and $k$ is the maximum number of parents a node can have (bounded by $\numOfVars$ in general).
This complexity can be decomposed as follows:
(i) the algorithm visits at most $\numOfVars (\numOfVars-1)$ states because each transition adds or removes exactly one edge;
(ii) each state has at most $\numOfVars^2$ neighboring states because of the maximum number of edges that can be added or removed to a DAG;
(iii) computing each neighboring state takes time $O(|E|\cdot k^2)$~\cite{chickering2002optimal} where $E$ is the number of edges at the state (bounded by $\numOfVars\cdot k$);
and (iv) computing the BIC score difference between the state and one of its neighbors takes $O(n\, k)$ (assuming the continuous case).
Therefore, the worst-case complexity is upper bounded by $O(\numOfVars^4 (\numOfVars\, k^3 + n\, k)) = O(\numOfVars^4 k \cdot \max(\numOfVars\, k^2, n))$.

As a consequence, GES should be used preferably with datasets that consist of few columns (its runtime grows with $\numOfVars^5 k^3$) and many rows (linear w.r.t. $n$).

\subsection{DirectLiNGAM}
\label{subsec:applicability_lingam}

Unlike PC, FCI, and GES, causal faithfulness assumption is not required in LiNGAM.
Instead, directLiNGAM
assumes that the data is continuous, the functional relations between variables are linear, and most importantly, with non-Gaussian noise terms~\cite{shimizu2014lingam}. 




The linearity assumption is important, because the LiNGAM algorithm incapsulates fitting a linear regression model. It is possible to apply Direct LiNGAM despite the violations in linearity~\cite{shimizu2011directlingam}, however in such a case the results should be interpreted with caution, as the algorithm can fail to identify causal connections because of under-fitting. Linearity can be judged by simply eye-balling the pair-wise plots of the data.

Assumption of non-Gaussianity of the error terms is crucial requirement of the algorithm, allowing to determine causal directions. However, it cannot be tested \textit{before} fitting the linear regression model and plotting out the error terms.
An indication of the non-Gaussian distribution of the error terms in a linear model can be suspected if the distributions of the variables are strongly non-Gaussian.
The distribution of the variables can be checked by plotting the histograms or applying Q-Q\footnote{Q–Q (quantile-quantile) plot is a probability plot, which allows to  graphically compare two probability distributions by plotting their quantiles against each other.} tests.
The exogenous variables can be Gaussian and have non-Gaussian error terms, however, they are determined only \textit{after} applying the model, so post-modeling testing of the compliance with the assumptions is recommended.

Most of the processing time of directLiNGAM is spent in the computation of the residuals (Line~\ref{lingam_step2} in Algorithm~\ref{alg:Direct_LiNGAM}).
The other heavy processing step is the regression to estimate the model parameters (Line~\ref{lingam_step4} in the same algorithm).
According to Shimizu et al.~\cite{shimizu2011directlingam}, the total complexity of directLiNGAM algorithm is $\mathcal{O}(sn^3M^2 + n^4M^3)$, where $s$ is the number of samples, $n$ is the number of variables and $M (\ll s)$ is the maximal rank found by the low-rank decomposition used in the independence measure~\cite{shimizu2011directlingam}.
Alternatively, the use of prior knowledge can significantly reduce the complexity of residuals computation (Line~\ref{lingam_step2}). 

\section{Causal Structure and Fairness}
\label{sec:causFairness}


Several fairness notions rely on causality to assess fairness and hence require a causal graph. The most basic causal-based fairness notion is total effect ($TE$)~\cite{pearl2009causality}\footnote{Known also as average causal effect ($ACE$).} which considers the overall effect of a variable $A$ on a variable $Y$. Assume that the sensitive variable $A$ can take two possible values $a_0$ (e.g. female) and $a_1$ (e.g. male) and that the positive outcome is $y^{\scriptscriptstyle +}$ (e.g. hiring), $TE_{a_1,a_0}(y^{\scriptscriptstyle +})$ is defined as follows:  
\begin{equation}
\label{eq:TE}
\pr(Y=y^{\scriptscriptstyle +} | do(A=a_1)) - \pr(Y=y^{\scriptscriptstyle +} | do(A=a_0))
\end{equation}
which measures the effect of the change of $A$ from $a_1$ to $a_0$ on $Y=y^{+}$ along all the causal paths from $A$ to $Y$. $\pr(Y=y | do(A=a))$ denotes the probability of $Y=y$ after an intervention $do(A=a)$. This is equivalent to probability of $Y=y$ after forcing all individuals in the population to have value $A=a$. $\pr(Y=y | do(A=a))$ is denoted $\pr(y_a)$ for short.

Direct effect ($DE$) is another fairness notion which focuses exclusively on the direct path $A \rightarrow Y$ (ignoring all indirect paths between $A$ and $Y$). The most general formulation of $DE$ is natural direct effect ($NDE$)~\cite{pearl01direct} defined as: 
\begin{equation}
\label{eq:NDE}
NDE_{a_1,a_0}(y^{\scriptscriptstyle +}) = \pr(y^{\scriptscriptstyle +}_{{a_1},\mathbf{Z}_{a_0}}) - \pr(y^{\scriptscriptstyle +}_{a_0})
\end{equation}
where $\mathbf{Z}$ is the set of mediator variables and $\displaystyle \pr(y^{\scriptscriptstyle +}_{{a_1},\mathbf{Z}_{a_0}})$ is the probability of $Y=y^{\scriptscriptstyle +}$ had $A$ been $a_1$ and had $\mathbf{Z}$ been the value it would naturally take if $A=a_0$.

Indirect effect ($IE$), which focuses rather on the indirect paths from $A$ to $Y$, can be computed using the natural indirect effect ($NIE$) ~\cite{pearl01direct} formula:
\begin{equation}
\label{eq:NIE}
NIE_{a_1,a_0} (y^{\scriptscriptstyle +}) = \pr(y^{\scriptscriptstyle +}_{{a_0},\mathbf{Z}_{a_1}}) - \pr(y^{\scriptscriptstyle +}_{a_0})
\end{equation}

Using the identifiability theory of causal inference~\cite{shpitser08,makhlouf2022identifiability}, the above expressions of fairness notions, involving interventions and counterfactuals, can be expressed in terms of observable probabilities.

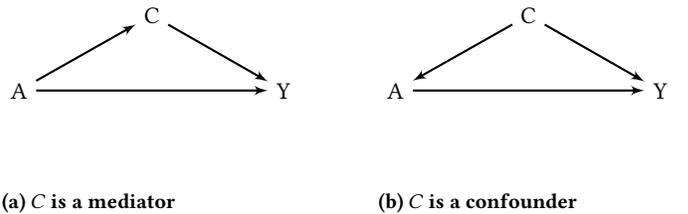
\begin{figure}[htp!]  
\centering 
\begin{subfigure}[b]{0.4\linewidth}
  \begin{tikzpicture}[%
        >=latex',
        cloud/.style={draw, thick, shape=ellipse, node distance=1cm, minimum width=0.8cm}]
        \draw[-,thick] (0,0) node[left] () {} ;
        \draw[-,thick] (4.5,0) node[left] () {} ;
        \draw[-,thick] (0,1) node[left] (A) {A} ;
        \draw[-,thick] (3.5,1) node[left] (Y) {Y} ;
        \path (A) -- coordinate (middle) (Y);
        \node[above of=middle] (C) {C};
        \draw[->,thick] (A) -- (Y);
        \draw[->,thick] (A) -- (C);
        \draw[->,thick] (C) -- (Y);
    \end{tikzpicture}
\caption{$C$ is a mediator} \label{subfig:fairness1a}  
\end{subfigure} \hfill
 \begin{subfigure}[b]{0.4\linewidth}
\begin{tikzpicture}[%
        >=latex',
        cloud/.style={draw, thick, shape=ellipse, node distance=1cm, minimum width=0.8cm}]
        \draw[-,thick] (0,0) node[left] () {} ;
        \draw[-,thick] (4.5,0) node[left] () {} ;
        \draw[-,thick] (0,1) node[left] (A) {A} ;
        \draw[-,thick] (3.5,1) node[left] (Y) {Y} ;
        \path (A) -- coordinate (middle) (Y);
        \node[above of=middle] (C) {C};
        \draw[->,thick] (A) -- (Y);
        \draw[->,thick] (C) -- (A);
        \draw[->,thick] (C) -- (Y);
    \end{tikzpicture}
    \caption{$C$ is a confounder} \label{subfig:fairness1b}  
  \end{subfigure} \hfill
\caption{Two simple causal graphs differing only on the direction of the edge between $A$ and $C$.}
\label{fig:simpleExampleFairness}
\end{figure}  

To see the impact of the causal graph structure on how these fairness notions can be identified and computed, consider the two simple graphs in Figure~\ref{fig:simpleExampleFairness}. Although both graphs differ in the orientation of a single edge between $A$ and $C$, they lead to significantly different expressions for the causal fairness notions. $TE$ is identifiable in both graphs. In the left graph, since there is no confounder, an intervention on $A$ ($do(A=a)$) coincides with conditioning on $A=a$. Hence, 
\begin{align}TE_{a_1,a_0}(y^{\scriptscriptstyle +}) &= \pr(y^{\scriptscriptstyle +}_{a_1}) - \pr(y^{\scriptscriptstyle +}_{a_0}) \label{eq:tv}\\
&=\pr(y^{\scriptscriptstyle +} | A=a_1) - \pr(y^{\scriptscriptstyle +} | A=a_0).\end{align}
However, in the slightly different graph on the right, $C$ is a confounder, and hence $TE$ is computed by adjusting on $C$ as follows\footnote{We are considering the discrete case.}:
\begin{align}TE_{a_1,a_0}(y^{\scriptscriptstyle +}) &= \pr(y^{\scriptscriptstyle +}_{a_1}) - \pr(y^{\scriptscriptstyle +}_{a_0})\\ 
&= \sum_{c\in dom(C)}\big(\pr(Y=y^{\scriptscriptstyle +} | A=a_1, C=c) \nonumber \\& \qquad - \pr(Y=y^{\scriptscriptstyle +} | A=a_0, C=c)\big)\;\pr(C=c)\end{align}

For the $NDE$, it is computed the same way in both graphs since it requires blocking all non-direct paths which is achieved by adjusting on variable $C$: 
\begin{align}NDE_{a_1,a_0}(y^{\scriptscriptstyle +}) 
&= \sum_{c\in dom(C)}\big(\pr(Y=y^{\scriptscriptstyle +} | A=a_1, C=c) \nonumber \\& \qquad - \pr(Y=y^{\scriptscriptstyle +} | A=a_0, C=c)\big)\;\pr(C=c)\end{align}

For the indirect effect, $NIE$ is equal to zero in the right graph since there is no causal indirect path between $A$ and $Y$, while for the left graph, it is equal to:
\begin{align}NIE_{a_1,a_0}(y^{\scriptscriptstyle +}) 
&= \sum_{c\in dom(C)}\pr(Y=y^{\scriptscriptstyle +} | A=a_0, C=c) \nonumber \\& \qquad \big(\pr(C=c | A=a_1) - \pr(C=c | A=a_0)\big).\end{align}

The disparity of identifying causal fairness notions due to slight diversion in the causal graphs holds also for other fairness notions as well~\cite{makhlouf2020causal}. This is further illustrated in the following experimental analysis section.

\section{Experimental Analysis}
\label{sec:empirical-results}

To study the impact of the causal discovery task on fairness, we apply the different causal discovery algorithms on two synthetic datasets and six real-world fairness benchmark datasets. A summary of all datasets is provided in Table~\ref{tab:datasetsInfo}. We use Tetrad~\cite{tetrad} implementation of PC, FCI and GES algorithms with a significance threshold ($\alpha$) set to $0.01$ for conditional independence testing. Depending on the type of the input data and the search algorithm, different CI tests are used. For instance, \textit{conditional gaussian likelihood ratio test} and \textit{conditional gaussian score} are used for mixed data. For continuous data,\textit{ K-CI test} and \textit{BIC score} are  applied.
For LiNGAM we use difference in Mutual Information for independence testing. Since LiNGAM algorithm aims to establish causal order, it is determined by collecting ordered ascending list of independence scores, the smallest corresponding to most exogenous variables. In the second phase of causal discovery, where the graph is refined by estimating connection weights, we set a threshold  ($\alpha$) to $0.05$ to exclude the connections with insignificant weights. For SBCN discovery, we use the same implementation as the original paper~\cite{bonchi2017exposing}. 

The only background knowledge we use in this study is temporal order using tiers. Variables are split into a set of ordered tiers (tier$1$, tier$2$, $\ldots$ tier$n$) which imply the following constraints. A variable in tier$i$ can be the cause of variables in the same tier or in subsequent tiers ($i+1 \ldots n$) but not of variables in previous tiers ($1 \ldots i-1$).

With the presence of the causal graph, several causal-based fairness notions can be used to assess fairness~\cite{makhlouf2020survey}. Some qualitative notions can be simply applied by checking the structure of the graph. For instance, to tell if there is (or not) a discrimination according to the ``no unresolved discrimination'' notion~\cite{kilbertus2017avoiding}, one needs to check if there is a directed path from the sensitive attribute $A$ to the outcome $Y$ which does not go through a resolving (explaining) variable. If such path exists in the graph, discrimination is concluded without further computation. A similar graph structure checking is needed for the ``no proxy discrimination''~\cite{kilbertus2017avoiding}. For other quantitative fairness notions, the structure of the graph is needed to distinguish between confounder, mediator, and collider variables. Quantitative fairness notions are typically computed by adjusting on variables. Adjusting on confounders allows to block spurious/backdoor paths. Adjusting on mediators is needed for mediation analysis (direct vs indirect vs path-specific discrimination). Identifying colliders, however, allows to avoid adjusting on them as this will introduce dependence that doesn't exist between variables.

We use five different causal-based fairness notions, namely, $ATE\_{IPW}$, $TE$, $DE$, $ID$, and $ED$ which correspond, respectively, to average total effect using inverse propensity weighting~\cite{rubin2015book}, total effect, direct effect, indirect discrimination, and explainable discrimination. 
Indirect discrimination and the explainable discrimination compute both the indirect causal effect between the sensitive variable and the outcome. However, indirect discrimination measures the path-specific effect with a proxy/redlining variable while the explainable discrimination considers the path-specific effect with an explaining variable. Thus, while the first is discriminatory, the second is legitimate and hence should be removed from the causal effect estimation. These and other causal-based fairness notions are described by Makhlouf et al.~\cite{makhlouf2020survey}. The $\textit{paths}$ package implementation~\cite{zhou2020tracing} is used to estimate  $TE$, $DE$, $ID$, and $ED$.

Computing (or estimating) discrimination using causal-based fairness notions consists in subtracting the probability of positive (desirable) output (e.g. hiring, granting a loan, etc.)  for the protected group (e.g. female) from the probability of positive output of the privileged group (e.g. male) as expressed in Equation~\ref{eq:TE}). This leads to values in the range $[-1,+1]$. A value of $0$ means the outcome is fair (no discrimination), a positive value indicates a discrimination \textit{against} the protected group, and a negative value indicates a discrimination \textit{in favor} of the protected group.

Estimating discrimination using all the above measures requires the knowledge of the confounder and mediator variables. However, PC, FCI, and GES algorithms can output partially directed graphs (PDAG) which do not guarantee to tell if a certain a variable is a confounder or mediator since some edges are left undirected. In such cases, we consider all possible ways of directing the (typically few) undirected edges\footnote{As long as they don't introduce a v-structure.}. For instance, if there are two undirected edges $X-W$ and $Z-Y$, there are 4 ways of directing them: $X\rightarrow W$ and $Z\rightarrow Y$, $X\leftarrow W$ and $Z\rightarrow Y$, $X\rightarrow W$ and $Z\leftarrow Y$, and $X\leftarrow W$ and $Z\leftarrow Y$. For each combination, we compute the discrimination and finally we report the range of values. This can be seen as bounding the discrimination value.

\footnotesize
\begin{table}[tbp]
\caption{Characteristics of the datasets used for the structural learning.}
\label{tab:datasetsInfo}
\renewcommand\arraystretch{1.2}
\begin{tabular}{@{} *5l @{}}    \toprule
\emph{Dataset} & \emph{Sample} & \emph{Data type}&\emph{Sensitive} &  \emph{Outcome} \\\midrule
Synthetic data ($1$)& $10000$ & continuous  & -  & -\\
Synthetic data ($2$)& $10000$ & continuous  & -  & - \\
Compas & $5915$ & mixed  & race  & recidivism \\
Adult & $32561$ & mixed  & race  & income \\
German credit & $1000$ & mixed  & sex & default \\
Dutch census & $60420$ & mixed & sex  & occupation \\
Boston housing & $506$ & continuous & race  & median price \\
Comm. \& crime & $1994$ & continuous & race  & violent crime rate \\\bottomrule
\hline
\end{tabular}
\end{table}
\normalsize

\subsection{Synthetic linear dataset}
\label{synthetic}

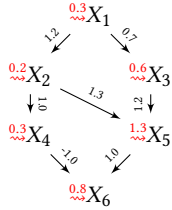
\begin{figure}[ht!]
  \centering
  \begin{tikzpicture}[>=latex',scale=0.8]

    \node (X1) at (1,0) [] {{\tiny\color{red} $\stackrel{0.3}{\rightsquigarrow}$}{$X_1$}};
    \node (X2) at (0,-1) [] {{\tiny\color{red} $\stackrel{0.2}{\rightsquigarrow}$}{$X_2$}};
    \node (X3) at (2,-1) [] {{\tiny\color{red} $\stackrel{0.6}{\rightsquigarrow}$}{$X_3$}};
    \node (X4) at (0,-2) [] {{\tiny\color{red} $\stackrel{0.3}{\rightsquigarrow}$}{$X_4$}};
    \node (X5) at (2,-2) [] {{\tiny\color{red} $\stackrel{1.3}{\rightsquigarrow}$}{$X_5$}};
    \node (X6) at (1,-3) [] {{\tiny\color{red} $\stackrel{0.8}{\rightsquigarrow}$}{$X_6$}};

    \path[every node/.style={sloped,anchor=south,auto=false}]
        (X1) edge [->] node {\tiny 1.2} (X2)    
    (X1) edge [->] node {\tiny 0.7} (X3)    
    (X2) edge [->] node {\tiny 1.0} (X4)    
    (X2) edge [->] node {\tiny 1.3} (X5)    
    (X3) edge [->] node {\tiny 1.2} (X5)    
    (X4) edge [->] node {\tiny -1.0} (X6)    
    (X5) edge [->] node {\tiny 1.0} (X6);
\end{tikzpicture}
  \caption{
    Scheme description of the synthetic linear datasets used.
    Each edge has a weight, and the noise standard deviations are indicated in red.
    The value of a node is the weighted sum of the values of the parents plus the noise.
  }
  \label{fig:synthetic-truth}
\end{figure}

In general, synthetic datasets are crucial for testing causal discovery algorithms systematically because, unlike real-world datasets, the ground truth graph is known and indisputable.
Here, we use synthetic datasets to illustrate the main differences and characteristics of the main causal discovery algorithms.

We generated two continuous linear datasets that have a very simple causal structure, but are rich enough for analyzing and discussing the algorithms.
Figure~\ref{fig:synthetic-truth} shows the six variables and the causal relationships between them.
The first dataset uses Gaussian noise and the second uniform noise, both centered at zero and scaled to achieve the desired standard deviation (shown in red). For instance, values of variable $X_5$ are generated in the first dataset as $X_5 = 1.3 X_2 + 1.2 X_3 + \mathcal{N}(0,1.3)$ while in the second dataset as $X_5 = 1.3 X_2 + 1.2 X_3 + \mathcal{U}(0,1.3)$. Note that the weights were chosen randomly.


\begin{figure}[H]
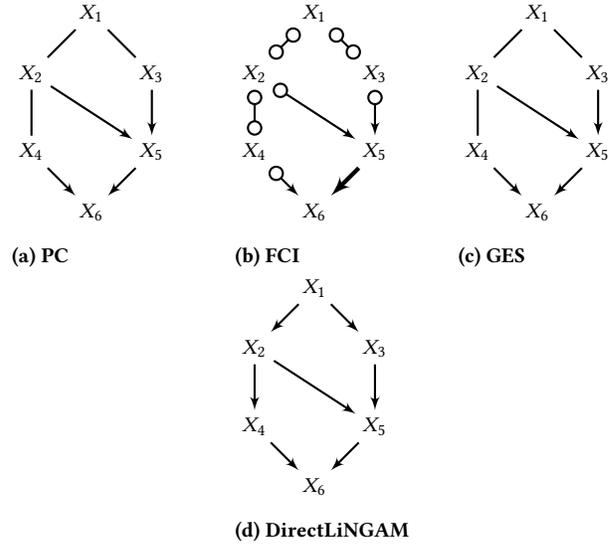

  \centering
  \begin{subfigure}[b]{0.3\linewidth}
    \begin{tikzpicture}[>=latex', scale=0.8]
      \small
      \input{TikZ/auto/Uniform/layout}
\draw[-,thick] (X1) -- (X2);
\draw[-,thick] (X1) -- (X3);
\draw[-,thick] (X2) -- (X4);
\draw[->,thick] (X2) -- (X5);
\draw[->,thick] (X3) -- (X5);
\draw[->,thick] (X4) -- (X6);
\draw[->,thick] (X5) -- (X6);
    \end{tikzpicture}
    \caption{PC} \label{subfig:Uniform-pc}
  \end{subfigure}
  \hfill
  \begin{subfigure}[b]{0.3\linewidth}
    \begin{tikzpicture}[>=latex', scale=0.8]
      \small
      \input{TikZ/auto/Uniform/layout}
\draw[->,ultra thick] (X5) -- (X6);
\draw[o-o,thick] (X1) -- (X2);
\draw[o-o,thick] (X1) -- (X3);
\draw[o-o,thick] (X2) -- (X4);
\draw[o->,thick] (X2) -- (X5);
\draw[o->,thick] (X3) -- (X5);
\draw[o->,thick] (X4) -- (X6);
    \end{tikzpicture}
    \caption{FCI} \label{subfig:Uniform-fci}
  \end{subfigure}
  \hfill
  \begin{subfigure}[b]{0.3\linewidth}
    \begin{tikzpicture}[>=latex', scale=0.8]
      \small
      \input{TikZ/auto/Uniform/layout}
\draw[-,thick] (X2) -- (X1);
\draw[-,thick] (X2) -- (X4);
\draw[->,thick] (X2) -- (X5);
\draw[-,thick] (X3) -- (X1);
\draw[->,thick] (X3) -- (X5);
\draw[->,thick] (X4) -- (X6);
\draw[->,thick] (X5) -- (X6);
    \end{tikzpicture}
    \caption{GES} \label{subfig:Uniform-ges}
  \end{subfigure}
  \\

  \begin{subfigure}[b]{0.3\linewidth}
    \begin{tikzpicture}[>=latex', scale=0.8]
      \small
      \input{TikZ/auto/Uniform/layout}
        \draw[->,thick] (X1) -- (X2);
  \draw[->,thick] (X1) -- (X3);
  \draw[->,thick] (X3) -- (X5);
  \draw[->,thick] (X2) -- (X4);
  \draw[->,thick] (X2) -- (X5);
  \draw[->,thick] (X4) -- (X6);
  \draw[->,thick] (X5) -- (X6);
    \end{tikzpicture}
    \caption{DirectLiNGAM} \label{subfig:Uniform-direct-lingam}
  \end{subfigure}
  \caption{Generated causal graphs for the synthetic dataset with Uniform noise.}
  \label{fig:UniformCG}
\end{figure}

Figure~\ref{fig:UniformCG} shows the graphs generated based on the first dataset. PC, FCI, and GES generate the correct causal graph skeleton, but fail to tell the direction of all edges.
The structure corresponds to a Markov equivalence class (CPDAG) where $4$ edges are (correctly) directed while the remaining $3$ are left undirected.
As expected, the constraint and score based algorithms could identify the directions of all edges involved in v-structures.
For the remaining edges ($X_1 - X_2, X_1 - X_3,$ and $X_2 - X_4$), they couldn't identify the direction because all possible combinations of directions will lead to the same conditional independence relations between variables\footnote{As long as the direction of edges do not introduce or remove a v-structure.}.
DirectLiNGAM, however, could generate the correct skeleton as well as the correct directions of the edges successfully.
This is possible because the first dataset satisfies exactly the assumptions for the applicability of LiNGAM.
That is, functional relations between variables are linear, values are continuous, and the noise distribution is non-Gaussian (uniform).
It is important to mention that, for DirectLiNGAM, finding the correct causal structure depends also on setting the right threshold ($\alpha$) for linear regression step (Line~\ref{lingam_step4} in Algorithm~\ref{alg:Direct_LiNGAM}).
For instance, the graph in Figure~\ref{subfig:Uniform-direct-lingam} is obtained with a threshold $\alpha = 0.05$.
Using a smaller value (e.g. $\alpha = 0.03$) leads to an extra false edge from $X_3$ to $X_6$.


\begin{figure}[H]
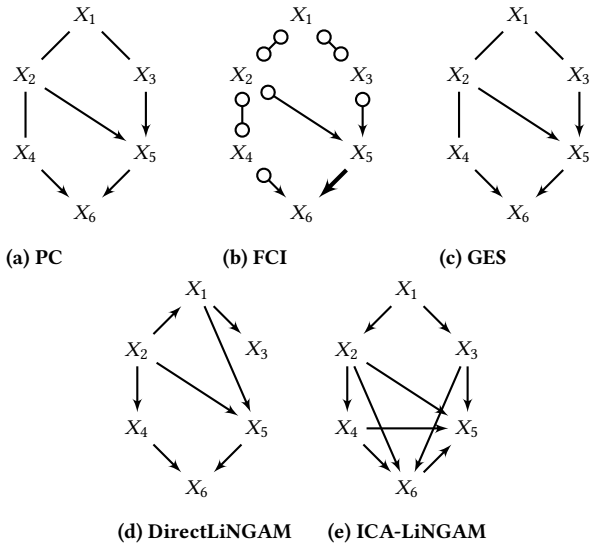

\centering
  \begin{subfigure}[b]{0.32\linewidth}
    \begin{tikzpicture}[>=latex',scale=0.8]
      \small
      \input{TikZ/auto/Gaussian/layout}
\draw[-,thick] (X1) -- (X2);
\draw[-,thick] (X1) -- (X3);
\draw[-,thick] (X2) -- (X4);
\draw[->,thick] (X2) -- (X5);
\draw[->,thick] (X3) -- (X5);
\draw[->,thick] (X4) -- (X6);
\draw[->,thick] (X5) -- (X6);
    \end{tikzpicture}
    \caption{PC} \label{subfig:Gaussian-pc}
  \end{subfigure}
  \hfill
  \begin{subfigure}[b]{0.32\linewidth}
    \begin{tikzpicture}[>=latex',scale=0.8]
      \small
      \input{TikZ/auto/Gaussian/layout}
\draw[->,ultra thick] (X5) -- (X6);
\draw[o-o,thick] (X1) -- (X2);
\draw[o-o,thick] (X1) -- (X3);
\draw[o-o,thick] (X2) -- (X4);
\draw[o->,thick] (X2) -- (X5);
\draw[o->,thick] (X3) -- (X5);
\draw[o->,thick] (X4) -- (X6);
    \end{tikzpicture}
    \caption{FCI} \label{subfig:Gaussian-fci}
  \end{subfigure}
  \hfill
  \begin{subfigure}[b]{0.32\linewidth}
    \begin{tikzpicture}[>=latex',scale=0.8]
      \small
      \input{TikZ/auto/Gaussian/layout}
\draw[-,thick] (X2) -- (X1);
\draw[-,thick] (X2) -- (X4);
\draw[->,thick] (X2) -- (X5);
\draw[-,thick] (X3) -- (X1);
\draw[->,thick] (X3) -- (X5);
\draw[->,thick] (X4) -- (X6);
\draw[->,thick] (X5) -- (X6);
    \end{tikzpicture}
    \caption{GES} \label{subfig:Gaussian-ges}
  \end{subfigure}
  \\
  \begin{subfigure}[b]{0.32\linewidth}
    \begin{tikzpicture}[>=latex',scale=0.8]
      \small
      \input{TikZ/auto/Gaussian/layout}
        \draw[->,thick] (X1) -- (X3);
  \draw[->,thick] (X1) -- (X5);
  \draw[->,thick] (X2) -- (X1);
  \draw[->,thick] (X2) -- (X5);
  \draw[->,thick] (X2) -- (X4);
  \draw[->,thick] (X4) -- (X6);
  \draw[->,thick] (X5) -- (X6);
    \end{tikzpicture}
    \caption{DirectLiNGAM} \label{subfig:Gaussian-direct_lingam}
  \end{subfigure}
  \begin{subfigure}[b]{0.32\linewidth}
      \begin{tikzpicture}[>=latex',scale=0.8]
        \small
        \input{TikZ/auto/Gaussian/layout}
\draw[->,thick] (X1) -- (X2);
\draw[->,thick] (X1) -- (X3);
\draw[->,thick] (X2) -- (X4);
\draw[->,thick] (X2) -- (X5);
\draw[->,thick] (X2) -- (X6);
\draw[->,thick] (X3) -- (X5);
\draw[->,thick] (X3) -- (X6);
\draw[->,thick] (X4) -- (X5);
\draw[->,thick] (X4) -- (X6);
\draw[->,thick] (X6) -- (X5);
      \end{tikzpicture}
    \caption{ICA-LiNGAM} \label{subfig:Gaussian-lingam}
  \end{subfigure}
  \caption{Generated causal graphs for the synthetic dataset with Gaussian noise.}
  \label{fig:GaussianCG}
\end{figure}

Figure~\ref{fig:GaussianCG} shows the graphs generated from the second dataset following the same causal structure (Figure~\ref{fig:synthetic-truth}) but with Gaussian noise. PC, FCI, and GES generate the same graph structure as the first dataset. The only additional detail is that FCI is very confident about the $X_5 \rightarrow X_6$ edge (highlighted with a thicker edge). DirectLiNGAM, however, generates a graph with several discrepancies compared with the correct graph. The same holds for ICA-LiNGAM (Figure~\ref{subfig:Gaussian-lingam}). Both graphs generated by LiNGAM fails even to correctly identify v-structures. This shows that LiNGAM is not reliable when the non-Gaussianity assumption of the noise does not hold.

\subsection{Compas}
\label{compas}
\begin{figure}[H]
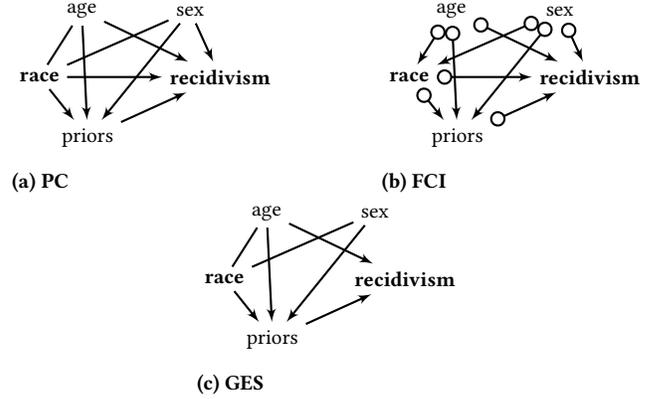

\centering
  \begin{subfigure}[b]{0.42\linewidth}
    \begin{tikzpicture}[>=latex', scale=0.8]
      \small
      \input{TikZ/auto/Compas/layout}
\draw[->,thick] (age) -- (priors_count);
\draw[->,thick] (age) -- (two_year_recid);
\draw[->,thick] (priors_count) -- (two_year_recid);
\draw[-,thick] (race) -- (age);
\draw[->,thick] (race) -- (priors_count);
\draw[->,thick] (race) -- (two_year_recid);
\draw[->,thick] (sex) -- (priors_count);
\draw[-,thick] (sex) -- (race);
\draw[->,thick] (sex) -- (two_year_recid);
    \end{tikzpicture}
    \caption{PC} \label{subfig:Compas-pc}
  \end{subfigure}
  \hfill
  \begin{subfigure}[b]{0.42\linewidth}
    \begin{tikzpicture}[>=latex', scale=0.8]
      \small
      \input{TikZ/auto/Compas/layout}
\draw[o->,thick] (age) -- (priors_count);
\draw[o->,thick] (age) -- (race);
\draw[o->,thick] (age) -- (two_year_recid);
\draw[o->,thick] (priors_count) -- (two_year_recid);
\draw[o->,thick] (race) -- (priors_count);
\draw[o->,thick] (race) -- (two_year_recid);
\draw[o->,thick] (sex) -- (priors_count);
\draw[o->,thick] (sex) -- (race);
\draw[o->,thick] (sex) -- (two_year_recid);
    \end{tikzpicture}
    \caption{FCI} \label{subfig:Compas-fci}
  \end{subfigure}
  \hfill
  \begin{subfigure}[b]{0.42\linewidth}
    \begin{tikzpicture}[>=latex', scale=0.8]
      \small
      \input{TikZ/auto/Compas/layout}
\draw[->,thick] (age) -- (priors_count);
\draw[->,thick] (age) -- (two_year_recid);
\draw[->,thick] (priors_count) -- (two_year_recid);
\draw[-,thick] (race) -- (age);
\draw[->,thick] (race) -- (priors_count);
\draw[->,thick] (sex) -- (priors_count);
\draw[-,thick] (sex) -- (race);
    \end{tikzpicture}
    \caption{GES} \label{subfig:Compas-ges}
  \end{subfigure}
  \caption{Generated causal graph for the Compas dataset.}
  \label{fig:CompasCG}
\end{figure}

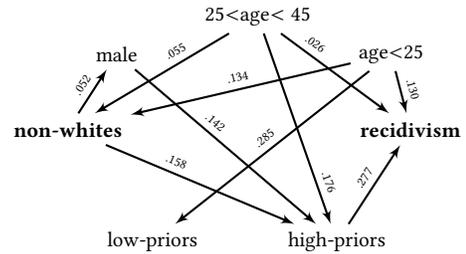
\begin{figure}[htp!]  
  \centering
  \begin{tikzpicture}[>=latex', scale=1.0]
    \small
    \draw[-,thick] (0,0) node[left] (non-whites) {\textbf{non-whites}};
\draw[-,thick](4.5,0) node[left] (recidivism) {\textbf{recidivism}};
\draw[-,thick] (.2,1) node[left] (male) {male};
\draw[-,thick](2.5,1.5) node[left] (25<age<45) {25$<$age$<4$5};
\draw[-,thick](4,1) node[left] (age<25) {age$<$25};
\draw[-,thick] (3.5,-1.5) node[left] (high-priors) {high-priors};
\draw[-,thick] (1,-1.5) node[left] (low-priors) {low-priors};
\draw[->,thick] (age<25) -- (non-whites) node [midway, above, sloped] (TextNode) {\tiny .134};
\draw[->,thick] (age<25) -- (low-priors) node [midway, above, sloped] (TextNode) {\tiny .285};
\draw[->,thick] (25<age<45) -- (non-whites) node [pos=.35, above, sloped] (TextNode) {\tiny .055};
\draw[->,thick] (25<age<45) -- (high-priors) node [pos=.8, above, sloped] (TextNode) {\tiny .176};
\draw[->,thick] (non-whites) -- (high-priors) node [pos=.35, above, sloped] (TextNode) {\tiny .158};
\draw[->,thick] (non-whites) -- (male) node [midway, above, sloped] (TextNode) {\tiny .052};
\draw[->,thick] (high-priors) -- (recidivism) node [midway, above, sloped] (TextNode) {\tiny .277};
\draw[->,thick] (25<age<45) -- (recidivism) node [pos=.25, above, sloped] (TextNode) {\tiny .026};
\draw[->,thick] (age<25) -- (recidivism) node [midway, above, sloped] (TextNode) {\tiny .130};
\draw[->,thick] (male) -- (high-priors) node [pos=.4, above, sloped] (TextNode) {\tiny .142};
  \end{tikzpicture}
  \caption{SBCN of non-whites} \label{subfig:compas_whites}  
  \label{fig:compasSBCN}
  \end{figure}

The \textit{Compas} dataset includes data from Broward County, Florida, initially compiled by ProPublica~\cite{angwin2016machine} and the goal is to predict the two-year violent recidivism. That is, whether a convicted individual would commit a violent crime in the following two years ($1$) or not ($0$). Only black and white defendants who were assigned \textit{Compas} risk scores within $30$ days of their arrest are kept for analysis~\cite{angwin2016machine} leading to $5915$ individuals in total. We consider race as sensitive feature. Five variables are used for the structural learning, namely: race, sex, age, priors and recidivism. Age and priors are continuous while the remaining variables are discrete. Three tiers in the partial order for temporal priority are used: race, sex and age are defined in the first tier, priors is in the second tier and recidivism is defined in the third tier. When found to be mediator, variables age and sex are considered as redlining, whereas priors as explaining. Since this dataset incudes mixed data, conditional Gaussian likelihood ratio test is used for PC and FCI, while conditional Gaussian test is used for GES. For the same reason (mixed dataset), LiNGAM is not applied. Figure~\ref{fig:CompasCG} shows the generated causal graphs for PC, FCI and GES. Note that, for clarity of illustration, in all subsequent causal graphs, the sensitive feature (on the left side of the graph) and the outcome (on the right side of the graph) are distinguished from the rest of the variables by highlighting them in bold. Figure~\ref{fig:compasSBCN} shows the SBCN for the protected group (non-white defendants). As mentioned previously (Section~\ref{subsec:SBCN}), SBCN graph provides also the edges weights. 


  Interestingly, GES and FCI could identify the direction of all edges in the graph skeleton. PC, however, still outputs undirected edges.

Applying SBCN on Compas dataset led to two disconnected subgraphs. 
  One for the protected group (non-white) and one for the privileged group (white). Similar to what Bonchi et al.~\cite{bonchi2017exposing} reported, the negative outcome (recidivism) node appears only in the protected group subgraph whereas the positive outcome (non-recidivism) appears only in the privileged group subgraph. Therefore, using random-walk formulas (Equations~\ref{eq:ds-} and~\ref{eq:rw}) will lead to trivial extreme values (total discrimination against the protected group). As an alternative, we interpret the SBCN as a typical causal graph to identify the confounder and mediator variables. Then, discrimination can be measured in the same way as the remaining graphs.
  
It is important to mention that the obtained graphs for Compas dataset do not agree on the direct edge from the sensitive attribute (race) to the outcome variable (recidivism). There is such an edge according to PC and FCI, but not according to GES and SBCN. This is of crucial importance to fairness as the direct effect is always discriminatory.

\begin{figure}[!ht]
    \includegraphics[scale=0.45, bb=0 0 520 300]{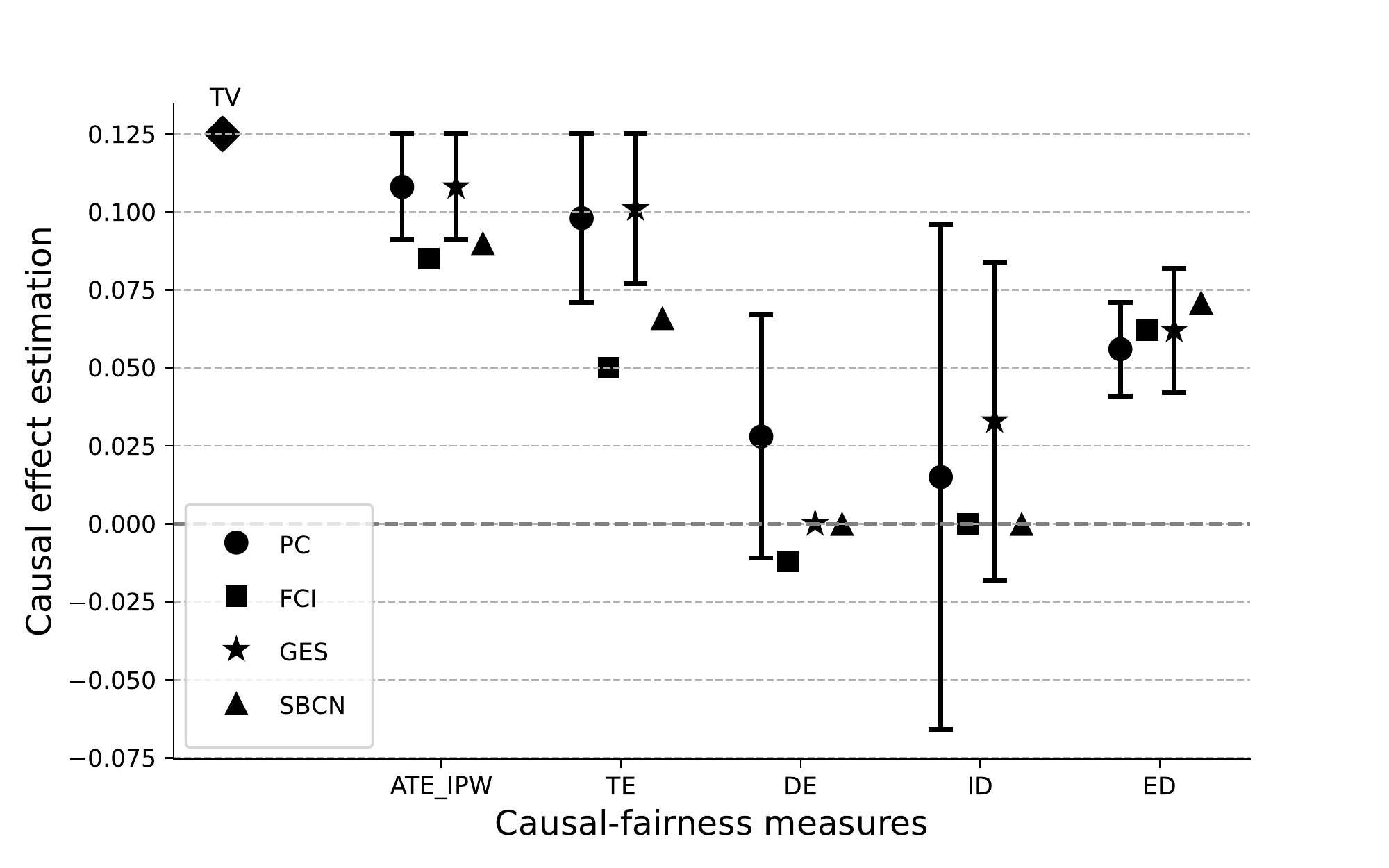} 
    \caption{Estimation of causal effects of the Compas dataset based on PC, FCI, GES and SBCN.}
    \label{fig:compasResults}
\end{figure}

Figure~\ref{fig:compasResults} shows the different discrimination measures using the different graphs.
Both $TE$ and $ATE\_IPW$ produce positive values which indicate a discrimination against non-white defendants. 

Considering the PC CPDAG (Figure~\ref{subfig:Compas-pc}), the highest value of $TE$ is obtained when there are no confounders (the two undirected edges are directed as $race \rightarrow age$ and $race \rightarrow sex$). In such graph, $TE$ coincides with $TV$ which is equal to $0.125$. The same high value of $TE$ is obtained with GES CPDAG (Figure~\ref{subfig:Compas-ges}) when the undirected edge is directed as $race \rightarrow age$. In such no confounding case, the presence or absence of the direct edge $race \rightarrow recidivism$ does not matter for $TE$. The smallest value for $TE$ ($0.050$) is only obtained in FCI PAG (Figure~\ref{subfig:Compas-fci}) where both age and sex variables are confounders. This implies that the total effect is going through only two paths $race \rightarrow recidivism$ and $race \rightarrow priors \rightarrow recidivism$. Such low $TE$ value cannot be obtained in PC nor in GES CPDAGs because the edges $age \rightarrow race$ and $sex \rightarrow race$ will create a new v-structure, and hence, lead to a causal graph outside the Markov equivalence class. In SBCN graph (Figure~\ref{fig:compasSBCN}), age is confounder but sex is a mediator which leads to a relatively low $TE$ value, but not as low as FCI.

For $DE$, the highest value ($0.067$) is obtained in PC CPDAG, when age is a mediator but sex is a confounder ($race \rightarrow age$ and $sex \rightarrow race$). The smallest value ($-0.012$) is obtained when both variables are mediators. $DE$ is naturally zero for GES and SBCN. 

$ID$ is highest ($0.096$) with PC when both age and sex are mediators ($race \rightarrow age$ and $race \rightarrow sex$). This is inline with GES as $ID$ is highest ($0.084$) with the same directions of the edges ($race \rightarrow age$ and $race \rightarrow sex$). 
Surprisingly, when age is a confounder while sex remains a redlining, the indirect discrimination \textit{against} blacks ($0.096$) becomes indirect discrimination \textit{in favor} of blacks ($-0.064$). This is an example of Simpson's paradox~\cite{simpson1951interpretation,berkeley75} when conditioning on a variable changes significantly the statistical conclusions. 

In the case where the edges as directed as $race \rightarrow age$ and $sex \rightarrow race$, both PC and GES graphs produce the same $ID$ value ($-0.018$). The case that leads the highest discrepancy in $ID$ values between PC and GES is $age \rightarrow race$ and $race \rightarrow sex$ (age is confounder and sex is a mediator). In such setup, according to PC, $ID$ is lowest ($-0.064$) while according to GES, $ID$ is zero as there is redlining path between race and recidivism.
It is important to mention here that if a causal path is going through redlining and explaining variables (e.g. $race \rightarrow sex \rightarrow priors \rightarrow recidivism$), it is considered as part of explained discrimination. The rule of thumb is that any path containing at least one explaining variable is considered as part of explained discrimination\footnote{This interpretation can be justified by considering the simple path $race \rightarrow priors \rightarrow recidivism$. Such path is clearly part of explained discrimination as priors is explaining variable. However, it contains also a  ``redlining'' variable which is the sensitive attribute race!}. $ID$ is zero for FCI and SBCN for the same reason (absence of redlining paths).

$ED$ values according to all graphs are comparable as all explained discrimination is going through the single explaining variable (priors).

Overall, Compas dataset shows that small variations in the graph structures can lead to significant differences in fairness conclusions. 
In particular, estimating $TE$ using graphs generated by different causal discovery algorithms can lead to a significant inconsistency ($0.125 - 0.050 = 0.075$) in assessing the amplitude of the discrimination against non-white defendants. Moreover, graphs generated by the same discovery algorithms (belong to the same Markov equivalence class), can lead to very different discrimination values ($ID$ goes from a positive discrimination of $0.096$ to a negative one ($-0.064$) due to reversing the direction of a single edge) which can be seen as a form of Simpson's paradox. 
Finally, the value of the threshold to decide about causal relations can have important consequences on fairness conclusion as well (missing $race \rightarrow recidivism$ edge in GES and SBCN).

\subsection{Adult}
\label{adult}
\begin{figure}[H]
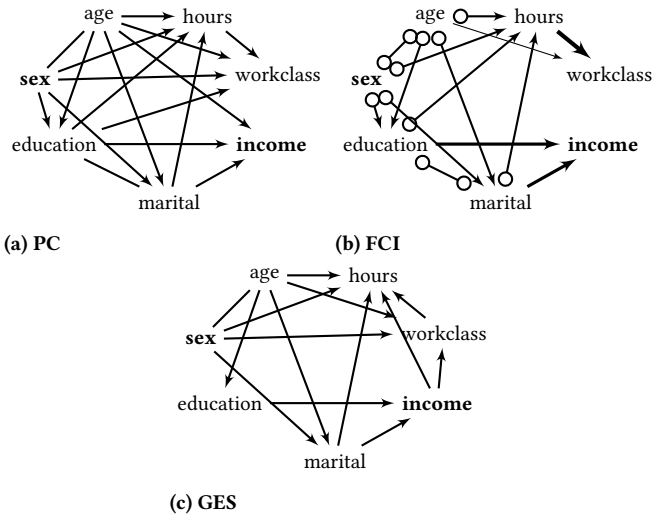

\centering
  \begin{subfigure}[b]{0.48\linewidth}
    \begin{tikzpicture}[>=latex', scale=0.85]
      \small
      \input{TikZ/auto/Adult/layout}
\draw[->,thick] (Sex) -- (education);
\draw[->,thick] (Sex) -- (hours_per_week);
\draw[->,thick] (Sex) -- (marital_status);
\draw[->,thick] (Sex) -- (workclass);
\draw[-,thick] (age) -- (Sex);
\draw[->,thick] (age) -- (education);
\draw[->,thick] (age) -- (hours_per_week);
\draw[->,thick] (age) -- (income);
\draw[->,thick] (age) -- (marital_status);
\draw[->,thick] (age) -- (workclass);
\draw[->,thick] (education) -- (hours_per_week);
\draw[->,thick] (education) -- (income);
\draw[-,thick] (education) -- (marital_status);
\draw[->,thick] (education) -- (workclass);
\draw[->,thick] (hours_per_week) -- (workclass);
\draw[->,thick] (marital_status) -- (hours_per_week);
\draw[->,thick] (marital_status) -- (income);
    \end{tikzpicture}
    \caption{PC} \label{subfig:Adult-pc}
  \end{subfigure}
  \hfill
  \begin{subfigure}[b]{0.48\linewidth}
    \begin{tikzpicture}[>=latex', scale=0.85]
      \small
      \input{TikZ/auto/Adult/layout}
\draw[->,ultra thick] (hours_per_week) -- (workclass);
\draw[->,very thick] (education) -- (income);
\draw[->,very thick] (marital_status) -- (income);
\draw[->,] (age) -- (workclass);
\draw[o->,thick] (Sex) -- (education);
\draw[o->,thick] (Sex) -- (hours_per_week);
\draw[o->,thick] (Sex) -- (marital_status);
\draw[o-o,thick] (age) -- (Sex);
\draw[o->,thick] (age) -- (education);
\draw[o->,thick] (age) -- (hours_per_week);
\draw[o->,thick] (age) -- (marital_status);
\draw[o->,thick] (education) -- (hours_per_week);
\draw[o-o,thick] (education) -- (marital_status);
\draw[o->,thick] (marital_status) -- (hours_per_week);
    \end{tikzpicture}
    \caption{FCI} \label{subfig:Adult-fci}
  \end{subfigure}
  \hfill
  \begin{subfigure}[b]{0.48\linewidth}
    \begin{tikzpicture}[>=latex', scale=0.85]
      \small
      \input{TikZ/auto/Adult/layout}
\draw[->,thick] (Sex) -- (hours_per_week);
\draw[->,thick] (Sex) -- (marital_status);
\draw[->,thick] (Sex) -- (workclass);
\draw[-,thick] (age) -- (Sex);
\draw[->,thick] (age) -- (education);
\draw[->,thick] (age) -- (hours_per_week);
\draw[->,thick] (age) -- (marital_status);
\draw[->,thick] (age) -- (workclass);
\draw[->,thick] (education) -- (income);
\draw[->,thick] (income) -- (hours_per_week);
\draw[->,thick] (income) -- (workclass);
\draw[->,thick] (marital_status) -- (hours_per_week);
\draw[->,thick] (marital_status) -- (income);
\draw[->,thick] (workclass) -- (hours_per_week);
    \end{tikzpicture}
    \caption{GES} \label{subfig:Adult-ges}
  \end{subfigure}
  \caption{Generated causal graph for the Adult dataset.}
  \label{fig:AdultCG}
\end{figure}

\begin{figure} [H] 
  \centering
  \captionsetup[sub]{justification=centering}
  \begin{tikzpicture}[>=latex', scale=0.7]
    \small
    \draw[-,thick] (-.5,-.5) node[left] (female) {\textbf{female}};
\draw[-,thick](9,-.5) node[left] (low-income) {\textbf{low-income}};
\draw[-,thick] (8,1.5) node[left] (low-hours) {low-hours};
\draw[-,thick] (7,-3) node[left] (work-other) {work-other};
\draw[-,thick] (5.5,-2) node[left] (work-self) {work-self};
\draw[-,thick] (1.5,1.5) node[left] (young-age) {young-age};
\draw[-,thick] (.5,-3) node[left] (no-college) {no-college};
\draw[-,thick] (2.1,-1.2) node[left] (not-married) {not-married};
\draw[-,thick] (4.5,1.5) node[left] (work-gov) {work-gov};
\draw[->,thick] (low-income) -- (work-other) node [midway, above, sloped] (TextNode) {\tiny .043};
\draw[->,thick] (low-income) -- (low-hours) node [midway, above, sloped] (TextNode) {\tiny .260};
\draw[->,thick] (female) -- (low-income) node [pos=.3, above, sloped] (TextNode) {\tiny .196};
\draw[->,thick] (female) -- (work-gov) node [pos=.15, above, sloped] (TextNode) {\tiny .027};
\draw[->,thick] (female) -- (not-married) node [pos=.45, above, sloped] (TextNode) {\tiny .447};
\draw[->,thick] (female) -- (young-age) node [pos=.5, above, sloped] (TextNode) {\tiny .076};
\draw[->,thick] (female) -- (no-college) node [pos=.5, above, sloped] (TextNode) {\tiny .046};
\draw[->,thick] (female) -- (low-hours) node [pos=.65, above, sloped] (TextNode) {\tiny .198};
\draw[->,thick] (not-married) -- (low-income) node [pos=.45, above, sloped] (TextNode) {\tiny .372};
\draw[->,thick] (not-married) -- (no-college) node [pos=.75, above, sloped] (TextNode) {\tiny .092};
\draw[->,thick] (not-married) -- (low-hours) node [pos=.65, above, sloped] (TextNode) {\tiny .164};
\draw[->,thick] (young-age) -- (low-income) node [pos=.75, above, sloped] (TextNode) {\tiny .215};
\draw[->,thick] (young-age) -- (work-self) node [pos=.85, above, sloped] (TextNode) {\tiny .057};
\draw[->,thick] (young-age) -- (not-married) node [pos=.25, above, sloped] (TextNode) {\tiny .281};
\draw[->,thick] (young-age) -- (no-college) node [pos=.65, above, sloped] (TextNode) {\tiny .084};
\draw[->,thick] (no-college) -- (low-income) node [pos=.75, above, sloped] (TextNode) {\tiny .324};
\draw[->,thick] (no-college) -- (work-other) node [pos=.75, above, sloped] (TextNode) {\tiny .033};
\draw[->,thick] (no-college) -- (work-self) node [pos=.75, above, sloped] (TextNode) {\tiny .078};
\draw[->,thick] (no-college) -- (low-hours) node [pos=.65, above, sloped] (TextNode) {\tiny .184};
\draw[->,thick] (low-hours) -- (work-other) node [pos=.75, above, sloped] (TextNode) {\tiny .049};
\draw[->,thick] (low-hours) -- (work-gov) node [pos=.65, above, sloped] (TextNode) {\tiny .036};
  \end{tikzpicture}

      \caption{SBCN of Females in the Adult dataset.} \label{subfig:adult_females}  
  \label{fig:adultSBCN}
  \end{figure}
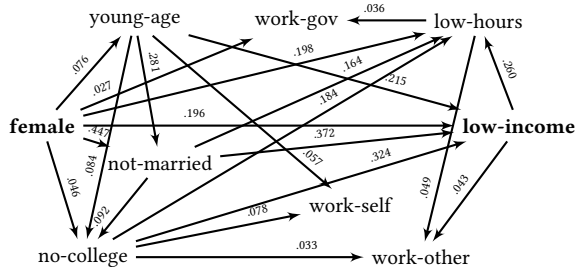

The \textit{Adult} dataset\footnote{https://archive.ics.uci.edu/ml/datasets/adult.} consists of $32,561$ samples and the goal is to predict the income of individuals based on several personal attributes such as age, race, sex, marital status, education, and employment. In this work, only $7$ variables are used for structural learning namely: age, sex, education level, marital status, work-class and number of working hours per week. The income of an individual can take two values namely: $\leq 50K$ (negative decision) or $>50K$ (positive decision). Age and number of working hours per week are continuous while the remaining variables are discrete. Three tiers in the partial order for temporal priority are used: age and sex are defined in the first tier, education and marital status in the second tier, and work-class, number of working hours per week and the income are defined in the last tier. When found to be mediators, variables age and marital status are considered as redlining, whereas education as explaining.The causal graphs generated by PC, FCI and GES are shown in Figure~\ref{fig:AdultCG}. Figure~\ref{fig:adultSBCN} shows the SBCN for females. As in the Compas dataset, LiNGAM cannot be used as data is mixed as well. 


\begin{figure}[!ht]
    \includegraphics[scale=0.45, bb=0 0 520 300]{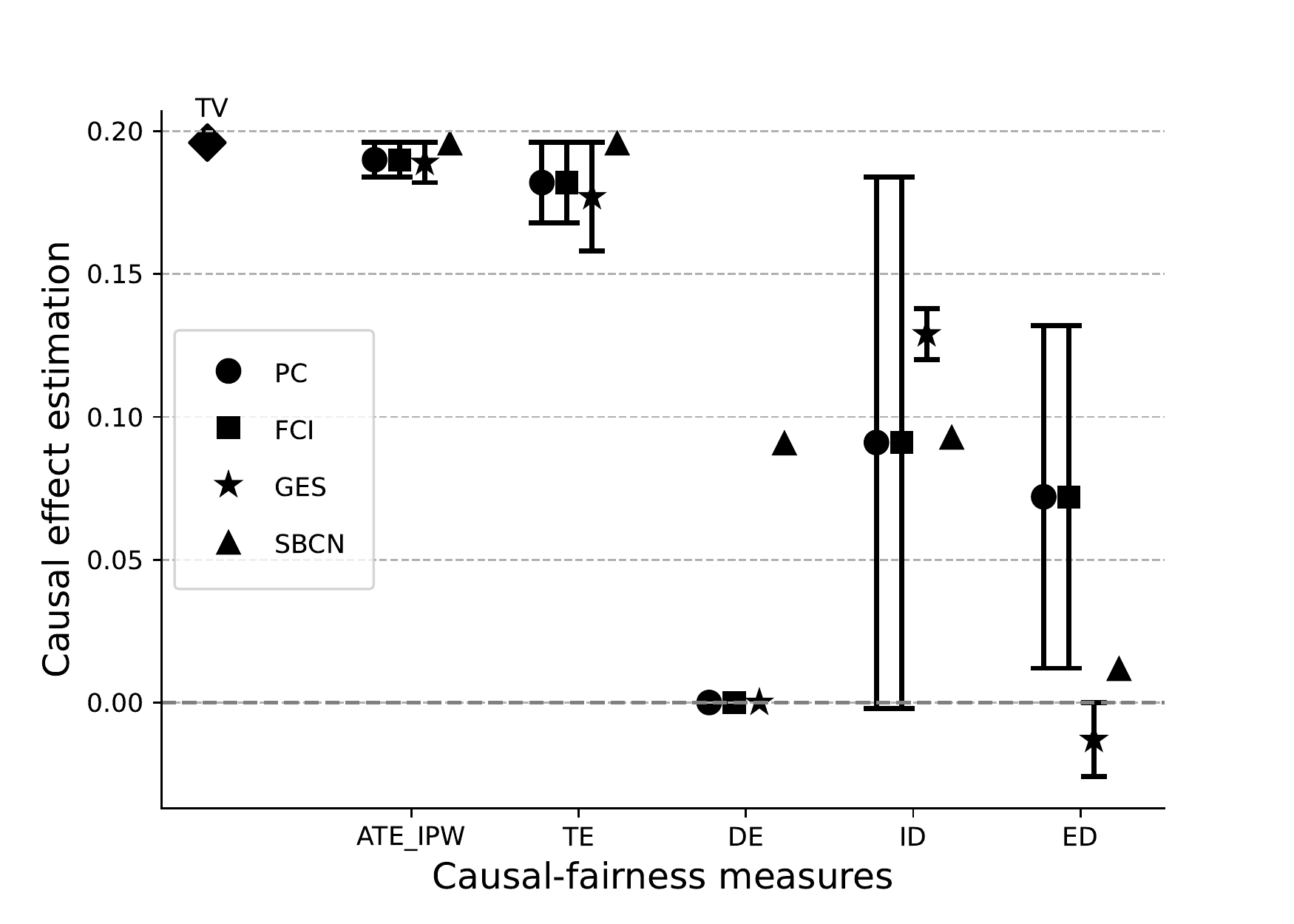} 
    \caption{Estimation of causal effects of the Adult dataset based on PC, FCI, GES and SBCN.}
    \label{fig:adultResults}
\end{figure}

There are two important notes about the generated graphs. First, only SBCN exhibits a direct edge between sex and income. Second, all remaining graphs have undirected edges (in particular, between sex and age). This leads to variability in the fairness measures as shown in Figure~\ref{fig:adultResults}. For instance, although all $TE$ and $ATE_{IPW}$ values are positive which indicates a discrimination against females, there is some variability in the extent of this discrimination. The highest discrimination can be seen in the GES CPDAG (Figure~\ref{subfig:Adult-ges}) where $sex \rightarrow age$ (age is a mediator) yields to $TE = 0.196$ whereas $age \rightarrow sex$ (age is a confounder) yields to $TE = 0.157$. $DE$ is zero according to all graphs except for SBCN since it is the only one with a direct edge between sex and income. 
For PC and FCI graphs (having the same structure with two undecided edges), $ID$ value ranges between $-0.003$ and $0.184$ where the former is obtained with $age \rightarrow sex$ and $education \rightarrow marital$ and the latter is obtained with $sex \rightarrow age$ and $education \rightarrow marital$. This is expected as $sex \rightarrow age$ opens an additional redlining path $sex \rightarrow age \rightarrow income$. In other words, having only one redlining path $sex \rightarrow marital \rightarrow income$ shows a very small indirect discrimination \textit{in favor} of females. Opening the other redlining path (through age) turns that into a clear indirect discrimination \textit{against} females. A possible explanation is that young married women tend to have low income due to motherhood responsibilities, while older married women passed that part of their life and are more available for their professional careers. 
Notice that, the lowest value of $ID$ in GES ($0.119$ obtained with $age \rightarrow sex$) is significantly higher than the lowest $ID$ value in PC and FCI ($-0.003$). The reason is that in GES, there is only one indirect (redlining and explained) path $sex \rightarrow marital \rightarrow recidivism$ while in PC and FCI, there are three different paths ($sex \rightarrow marital \rightarrow income$, $sex \rightarrow education \rightarrow income$, and $sex \rightarrow education \rightarrow marital \rightarrow income$). Hence, the causal effect between sex and income in GES is only conveyed through the redlining path. Whereas in PC and FCI, it is split between the redlining path and also the two other explained discrimination paths. 

For $ED$, the highest value ($0.132$) is obtained in PC and FCI when age is confounder ($age \rightarrow sex$) and marital status is a mediator between education and income ($education \rightarrow marital$). The smallest value ($-0.027$) is obtained in GES when age is a mediator ($sex \rightarrow age$) which indicates a small explained discrimination \textit{in favor} of females through the path $sex \rightarrow age \rightarrow education \rightarrow income$. This path is only possible as a single explaining path in GES CPDAG. In all the graphs obtained by the other algorithms, such path is possible but along other explaining paths, in particular, $sex \rightarrow education \rightarrow income$. This explains why the discrimination in favor of females is only observable with GES. It is interesting to notice that in PC and FCI graphs, the explained discrimination through $sex \rightarrow education \rightarrow income$ is slightly positive ($0.016$) whereas in GES graph, adding another mediator $sex \rightarrow age \rightarrow education \rightarrow income$ yields a slightly negative explained discrimination. As there is no overlap between the ranges of $ED$ values in PC and FCI graphs on one hand and GES on another, and that values (although small) have different signs (positive vs negative), the explained discrimination conclusions depend on which algorithm is used to discover causal relations.

Compared to Compas dataset, the mediation analysis on adult dataset reveals two additional fairness relevant observations. First, a specific causal path can be discovered by several causal discovery algorithms. However, the causal effect through that path may significantly differ depending the presence of other causal paths not necessarily with the same interpretation (redlining or explaining paths). 
Second, even with the same causal path (e.g. $sex \rightarrow education \rightarrow income$), considering a mediator (e.g. age) can reverse the type of the discrimination (e.g. $sex \rightarrow age \rightarrow education \rightarrow income$.

\subsection{Dutch census}
\label{dutch}
The \textit{Dutch Census} dataset consists of $60,420$ tuples where the sensitive attribute is the sex of an individual and the outcome is her occupation (job). Six attributes are used for the structural learning namely: age, sex, economic status, education, marital status and occupation. Age is continuous while the remaining variables are discrete. Three tiers in the partial order for temporal priority are used: age and sex are defined in the first tier, education in the second tier, and marital status, economic status and occupation are defined in the third tier. When found to be mediators, only education considered explaining variable. The rest (age, employment status, and marital status) are considered redlining variables.


\begin{figure}[H]
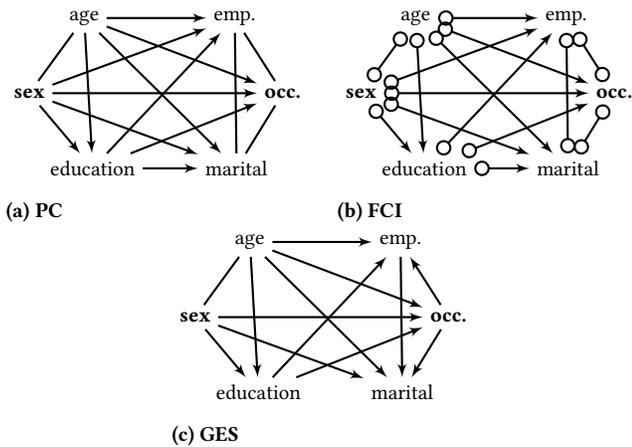

\centering
  \begin{subfigure}[b]{0.48\linewidth}
    \begin{tikzpicture}[>=latex', scale=0.8]
      \small
      \input{TikZ/auto/Dutch/layout}
\draw[-,thick] (Marital_status) -- (economic_status);
\draw[->,thick] (age) -- (Marital_status);
\draw[->,thick] (age) -- (economic_status);
\draw[->,thick] (age) -- (edu_level);
\draw[->,thick] (age) -- (occupation);
\draw[->,thick] (edu_level) -- (Marital_status);
\draw[->,thick] (edu_level) -- (economic_status);
\draw[->,thick] (edu_level) -- (occupation);
\draw[-,thick] (occupation) -- (Marital_status);
\draw[-,thick] (occupation) -- (economic_status);
\draw[->,thick] (sex) -- (Marital_status);
\draw[-,thick] (sex) -- (age);
\draw[->,thick] (sex) -- (economic_status);
\draw[->,thick] (sex) -- (edu_level);
\draw[->,thick] (sex) -- (occupation);
    \end{tikzpicture}
    \caption{PC} \label{subfig:Dutch-pc}
  \end{subfigure}
  \hfill
  \begin{subfigure}[b]{0.48\linewidth}
    \begin{tikzpicture}[>=latex', scale=0.8]
      \small
      \input{TikZ/auto/Dutch/layout}
\draw[o-o,thick] (Marital_status) -- (economic_status);
\draw[o->,thick] (age) -- (Marital_status);
\draw[o->,thick] (age) -- (economic_status);
\draw[o->,thick] (age) -- (edu_level);
\draw[o->,thick] (age) -- (occupation);
\draw[o->,thick] (edu_level) -- (Marital_status);
\draw[o->,thick] (edu_level) -- (economic_status);
\draw[o->,thick] (edu_level) -- (occupation);
\draw[o-o,thick] (occupation) -- (Marital_status);
\draw[o-o,thick] (occupation) -- (economic_status);
\draw[o->,thick] (sex) -- (Marital_status);
\draw[o-o,thick] (sex) -- (age);
\draw[o->,thick] (sex) -- (economic_status);
\draw[o->,thick] (sex) -- (edu_level);
\draw[o->,thick] (sex) -- (occupation);
    \end{tikzpicture}
    \caption{FCI} \label{subfig:Dutch-fci}
  \end{subfigure}
  \hfill
  \begin{subfigure}[b]{0.48\linewidth}
    \begin{tikzpicture}[>=latex', scale=0.8]
      \small
      \input{TikZ/auto/Dutch/layout}
\draw[->,thick] (age) -- (Marital_status);
\draw[->,thick] (age) -- (economic_status);
\draw[->,thick] (age) -- (edu_level);
\draw[->,thick] (age) -- (occupation);
\draw[->,thick] (economic_status) -- (Marital_status);
\draw[->,thick] (edu_level) -- (economic_status);
\draw[->,thick] (edu_level) -- (occupation);
\draw[->,thick] (occupation) -- (Marital_status);
\draw[->,thick] (occupation) -- (economic_status);
\draw[->,thick] (sex) -- (Marital_status);
\draw[-,thick] (sex) -- (age);
\draw[->,thick] (sex) -- (edu_level);
\draw[->,thick] (sex) -- (occupation);
    \end{tikzpicture}
    \caption{GES} \label{subfig:Dutch-ges}
  \end{subfigure}
  \caption{Generated causal graph for the Dutch census dataset. Occ. stands for occupation and emp. stands for whether an individual is employed or not.}
  \label{fig:DutchCG}
\end{figure}

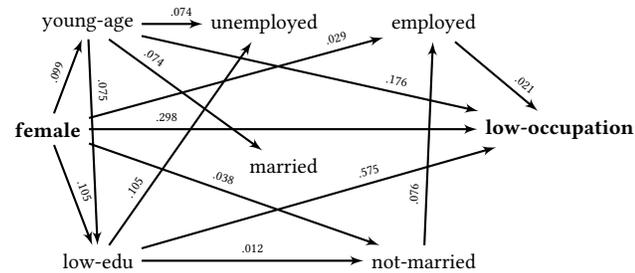
\begin{figure}[H]  
  \captionsetup[sub]{justification=centering}
    \begin{tikzpicture}[>=latex', scale=0.7]
      \small
      \draw[-,thick] (-.5,-.5) node[left] (female) {\textbf{female}};
\draw[-,thick](10,-.5) node[left] (low-occupation) {\textbf{low-occupation}};
\draw[-,thick] (7,1.5) node[left] (employed) {employed};
\draw[-,thick] (7,-3) node[left] (not-married) {not-married};
\draw[-,thick] (4,-1.2) node[left] (married) {married};
\draw[-,thick] (.5,1.5) node[left] (young-age) {young-age};
\draw[-,thick] (.5,-3) node[left] (low-edu) {low-edu};
\draw[-,thick] (4,1.5) node[left] (unemployed) {unemployed};
\draw[->,thick] (female) -- (low-occupation) node [pos=.2, above, sloped] (TextNode) {\tiny .298};
\draw[->,thick] (female) -- (not-married) node [pos=.45, above, sloped] (TextNode) {\tiny .038};
\draw[->,thick] (female) -- (young-age) node [pos=.5, above, sloped] (TextNode) {\tiny .099};
\draw[->,thick] (female) -- (low-edu) node [pos=.5, above, sloped] (TextNode) {\tiny .105};
\draw[->,thick] (female) -- (employed) node [pos=.85, above, sloped] (TextNode) {\tiny .029};
\draw[->,thick] (low-edu) -- (not-married) node [midway, above, sloped] (TextNode) {\tiny .012};
\draw[->,thick] (not-married) -- (employed) node [pos=.25, above, sloped] (TextNode) {\tiny .076};
\draw[->,thick] (young-age) -- (low-occupation) node [pos=.75, above, sloped] (TextNode) {\tiny .176};
\draw[->,thick] (young-age) -- (married) node [pos=.25, above, sloped] (TextNode) {\tiny .074};
\draw[->,thick] (young-age) -- (low-edu) node [pos=.25, above, sloped] (TextNode) {\tiny .075};
\draw[->,thick] (young-age) -- (unemployed) node [pos=.65, above, sloped] (TextNode) {\tiny .074};
\draw[->,thick] (low-edu) -- (low-occupation) node [pos=.65, above, sloped] (TextNode) {\tiny .575};
\draw[->,thick] (low-edu) -- (unemployed) node [pos=.25, above, sloped] (TextNode) {\tiny .105};
\draw[->,thick] (employed) -- (low-occupation) node [pos=.75, above, sloped] (TextNode) {\tiny .021};
    \end{tikzpicture}
      \caption{SBCN of Females in the Dutch census dataset.} \label{subfig:dutch_females}  
    
  \label{fig:dutchSBCN}
  \end{figure}  
  
The obtained graphs in Figures~\ref{fig:DutchCG} and~\ref{fig:dutchSBCN} show that PC and FCI produce very similar structures which are significantly different than GES and SBCN graphs. In particular, the $status - occupation$, $marital - occupation$, and $status - marital$ edges are undirected in PC and FCI, but directed in GES. This has a significant consequence on the set of possible causal paths between the sensitive attribute and the outcome. 

\begin{figure}[!ht]
    \includegraphics[scale=0.45, bb=0 0 520 300]{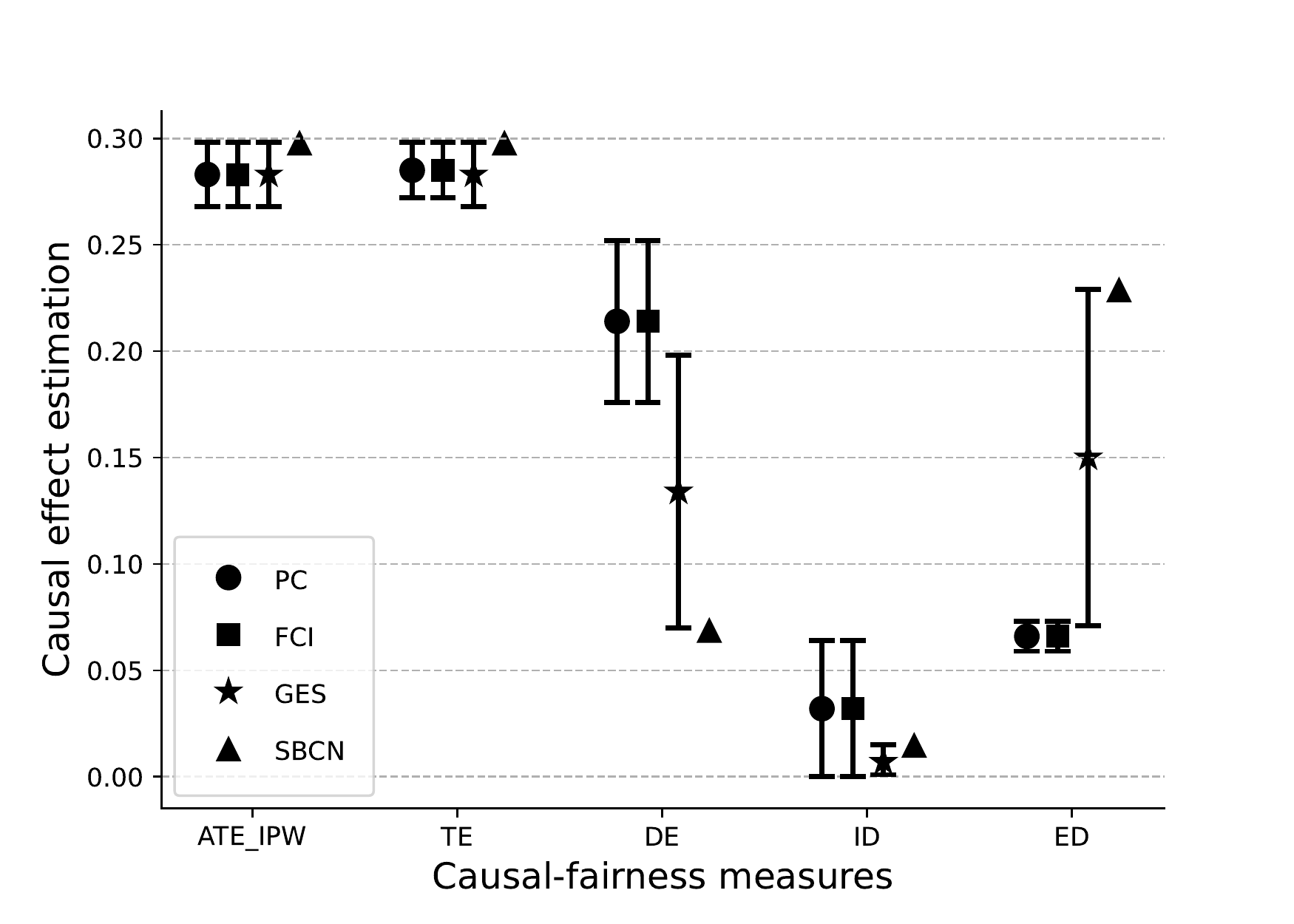} 
    \caption{Estimation of causal effects of the Dutch census dataset based on PC, FCI, GES and SBCN.}
    \label{fig:dutchResults}
\end{figure}

As shown in Figure~\ref{fig:dutchResults}, total effect measures (as high as $0.3$ obtained when age is a confounder $age \rightarrow sex$)) indicate a significant discrimination against females. The highest variability is observed for $DE$ values. When age is a confounder and employment status and marital status variables are mediators, PC and FCI graphs exhibit $6$ causal paths

For PC and FCI, there are in total $17$ causal paths between $sex$ and $occupation$, whereas in $GES$, there are only $4$. 

Note that the edge $emp. - marital$ should be left unoriented as orienting it in either way will create a collider. The edges $emp. - \textbf{occ.}$ and $marital. - \textbf{occ.}$ should be oriented as $emp. \rightarrow \textbf{occ.}$ and $marital. \rightarrow \textbf{occ.}$ for the same reason (i.e.; not creating colliders).

Figure~\ref{fig:dutchResults} show the values of fairness measures
As shown in Figure~\ref{fig:dutchResults}, all measures (except $ID$) are negative and hence indicate a discrimination \textit{against} female individuals. In particular, total effect values are around $-0.3$ according to all graphs. There is a high variability in the value of $ED$ according to PC and FCI graphs depending on whether $age$ variable is a confounder or a mediator (redlining) and whether $status$ and $marital$ variables are mediators or colliders. For instance, if $age$ is a confounder and $status$ and $marital$ are mediators, apart from the direct causal link ($sex \rightarrow occupation$), there will be only one causal path, namely, $sex \rightarrow education \rightarrow occupation$, and hence the $DE$ will be at its lowest ($-0.25$). The only measure that might return positive values is $ID$ for PC and FCI where there are $7$ possible indirect causal paths going through redlining variables. The total indirect discrimination can be slightly positive, indicating a discrimination \textit{in favor} of females. If taken separately, such values are misleading because they should be considered along the direct discrimination ($DE$). For GES, there is only one possible indirect discrimination path ($sex \rightarrow age \rightarrow occupation$) and two possible explaining discrimination paths ($sex \rightarrow education \rightarrow occuptation$ and $sex \rightarrow age \rightarrow education \rightarrow occupation$). 

\subsection{German credit}
\label{german}
The \textit{German credit} dataset\footnote{https://archive-beta.ics.uci.edu/ml/datasets/statlog+german+credit+data} contains data of $1000$ individuals
applying for loans. The variables used for causal graph generation are: sex, age, credit amount, employment length and default. Age and credit amount are continuous while the remaining variables are discrete. This dataset is designed for binary classification to predict whether an individual will default on the loan $(1)$ or not $(0)$. We consider sex as sensitive feature where female applicants are compared to male applicants. Three tiers in the partial order for temporal priority are used: age and sex are defined in the first tier, credit amount and employment length in the second tier, and default is defined in the third tier. If found as mediator, age variable is considered as redlining.


\begin{figure}[H]
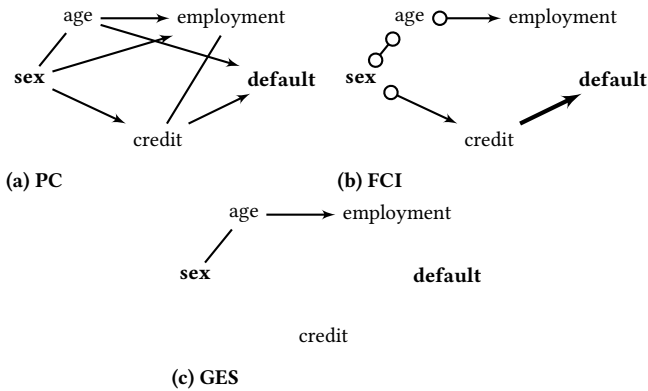

\centering
  \begin{subfigure}[b]{0.48\linewidth}
    \begin{tikzpicture}[>=latex', scale=0.8]
      \small
      \input{TikZ/auto/German/layout}
\draw[->,thick] (age) -- (default);
\draw[->,thick] (age) -- (present_emp_since);
\draw[->,thick] (credit_amount) -- (default);
\draw[-,thick] (personal_status_sex) -- (age);
\draw[->,thick] (personal_status_sex) -- (credit_amount);
\draw[->,thick] (personal_status_sex) -- (present_emp_since);

\draw[-,thick] (present_emp_since) -- (credit_amount);
    \end{tikzpicture}
    \caption{PC} \label{subfig:German-pc}
  \end{subfigure}
  \hfill
  \begin{subfigure}[b]{0.48\linewidth}
    \begin{tikzpicture}[>=latex', scale=0.8]
      \small
      \input{TikZ/auto/German/layout}
\draw[->,ultra thick] (credit_amount) -- (default);
\draw[o->,thick] (age) -- (present_emp_since);
\draw[o-o,thick] (personal_status_sex) -- (age);
\draw[o->,thick] (personal_status_sex) -- (credit_amount);

    \end{tikzpicture}
    \caption{FCI} \label{subfig:German-fci}
  \end{subfigure}
  \hfill
  \begin{subfigure}[b]{0.48\linewidth}
    \begin{tikzpicture}[>=latex', scale=0.8]
      \small
      \input{TikZ/auto/German/layout}
\draw[-,thick] (age) -- (personal_status_sex);
\draw[->,thick] (age) -- (present_emp_since);
    \end{tikzpicture}
    \caption{GES} \label{subfig:German-ges}
  \end{subfigure}
  \caption{Generated causal graph for the German credit dataset.}
  \label{fig:GermanCG}
\end{figure}

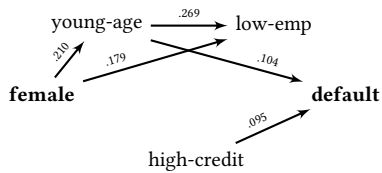
\begin{figure}[ht]  
  \centering
  \captionsetup[sub]{justification=centering}
    \begin{tikzpicture}[>=latex', scale=0.9]
      \small
      \draw[-,thick] (0,0) node[left] (female) {\textbf{female}};
\draw[-,thick](4.5,0) node[left] (default) {\textbf{default}};
\draw[-,thick] (1,1) node[left] (young-age) {young-age};
\draw[-,thick](3.5,1) node[left] (low-emp) {low-emp};
\draw[-,thick] (2.5,-1) node[left] (high-credit) {high-credit};
\draw[->,thick] (young-age) -- (default) node [pos=.75, above, sloped] (TextNode) {\tiny .104};
\draw[->,thick] (young-age) -- (low-emp) node [midway, above, sloped] (TextNode) {\tiny .269};
\draw[->,thick] (high-credit) -- (default) node [pos=.35, above, sloped] (TextNode) {\tiny .095};
\draw[->,thick] (female) -- (low-emp) node [pos=.25, above, sloped] (TextNode) {\tiny .179};
\draw[->,thick] (female) -- (young-age) node [midway, above, sloped] (TextNode) {\tiny .210};
    \end{tikzpicture}
    \caption{SBCN of Females in the German credit dataset.} \label{subfig:german_females}  
\end{figure}  

\begin{figure}[!ht]
    \includegraphics[scale=0.45, bb=0 0 520 300]{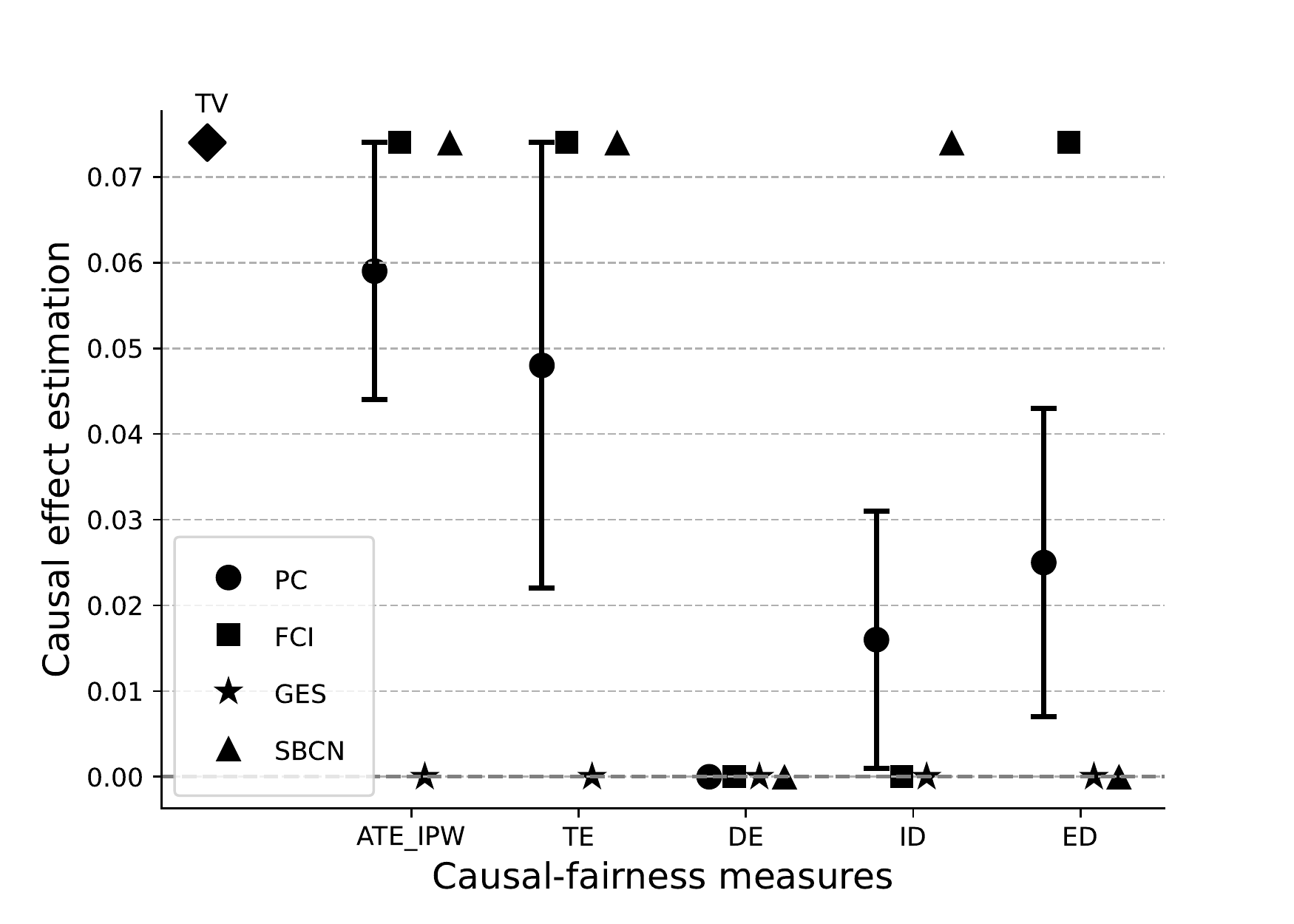} 
    \caption{Estimation of causal effects of the German credit dataset based on PC, FCI, GES and SBCN.}
    \label{fig:germanResults}
\end{figure}

Compared to previous datasets, German credit leads to sparser causal graphs (Figures~\ref{fig:GermanCG} and~\ref{subfig:german_females}). The most extreme case is GES which could not identify any dependence between sex and default variables. Besides, no algorithm could detect a direct dependence between sex and default variables.  
Interestingly, all graphs (except GES) show a causal relation from credit amount to default, with FCI very confident about it. Discrimination values in Figure~\ref{fig:germanResults} show that, all discrimination measures are either zero or slightly positive indicating a small discrimination against females. For GES, all causal effects are equal zero due to the absence of any causal path from sex to default. The range of $TE$ and $ATE_{IPW}$ values for PC is relatively wide. The lowest value ($0.021$) is obtained when age is a confounder ($age \rightarrow sex$). The highest value ($0.074$) is obtained when age is a mediator (redlining). In total, there are $4$ possible causal paths from sex to default according to PC. For FCI and SBCN, there is only one causal path: $sex \rightarrow credit \rightarrow default$ in FCI, and $sex \rightarrow age \rightarrow default$ in SBCN.
Therefore, $ID$ is different than zero in SBCN (and PC since the same path is the only possible indirect discrimination) and $ED$ is different than zero in FCI.

\subsection{Boston Housing}
\label{Boston}
The Boston housing dataset holds the statistics on $506$ cases of Boston areas including diverse variables used to predict median real estate value in the district. The data has a sensitive predictor variable - the proportion of black people living in the area. The data is collected by U.S Census Service and can be found in  StatLib archive \footnote{http://lib.stat.cmu.edu/datasets/boston} and was originally published by Harrison et al.~\cite{harrison1978hedonic}. The dataset has been used extensively to benchmark machine learning algorithms, however its use for fairness in machine learning is very limited\footnote{To the best of our knowledge, it has been only used to illustrate fairness preprocessing tools in SciKitLearn https://scikit-fairness.readthedocs.io/en/latest/fairness\_boston\_housing.html}. The dataset originally contains $14$ variables, but only $7$ are used for empirical experiments. We removed two variables because of missing values and another $5$ to avoid multicollinearity and simplify the graphs. All the variables in the data are continuous, mostly follow non Gaussian distribution (as found by the quantiles tests (QQ)). 
By contrast to the above datasets, LiNGAM is applied along with all the other search algorithms since the data is totally continuous.

\begin{figure}[H]
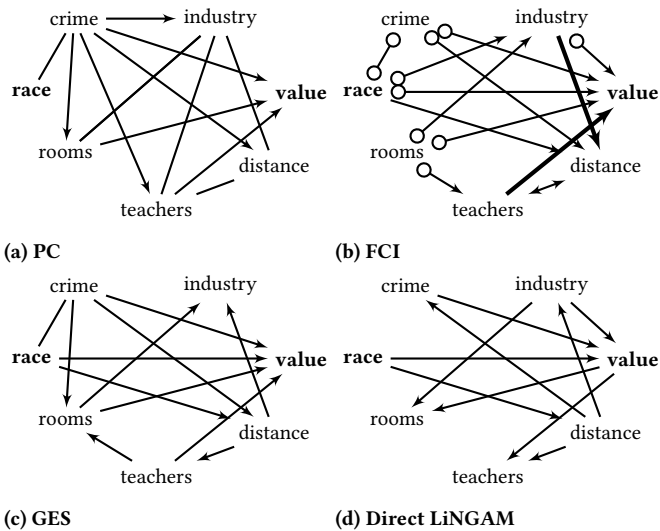

\centering
  \begin{subfigure}[b]{0.48\linewidth}
    \begin{tikzpicture}[>=latex', scale=0.78]
      \small
      \input{TikZ/auto/Boston/layout}
\draw[-,thick] (CRIM) -- (B);
\draw[->,thick] (CRIM) -- (MEDV);
\draw[->,thick] (CRIM) -- (RAD);
\draw[->,thick] (CRIM) -- (PTRATIO);
\draw[->,thick] (CRIM) -- (RM);
\draw[->,thick] (CRIM) -- (INDUS);
\draw[-,thick] (INDUS) -- (RAD);
\draw[-,thick] (INDUS) -- (PTRATIO);
\draw[-,thick] (INDUS) -- (RM);
\draw[->,thick] (PTRATIO) -- (MEDV);
\draw[-,thick] (RAD) -- (PTRATIO);
\draw[->,thick] (RM) -- (MEDV);
\draw[-,thick] (RM) -- (INDUS);
    \end{tikzpicture}
    \caption{PC} \label{subfig:Boston-pc}
  \end{subfigure}
  \hfill
  \begin{subfigure}[b]{0.48\linewidth}
    \begin{tikzpicture}[>=latex', scale=0.78]
      \small
      \input{TikZ/auto/Boston/layout}
\makeatletter
\def\race@{B}
\def\distance@{RAD}
\def\value@{MEDV}
\def\crime@{CRIM}
\def\teachers@{PTRATIO}
\def\industry@{INDUS}
\def\rooms@{RM}

\draw[->,thick] (\race@) -- (\distance@);
\draw[o->,thick] (\race@) -- (\value@);
\draw[o-o,thick] (\race@) -- (\crime@);
\draw[o->,thick] (\race@) -- (\industry@);
\draw[->,ultra thick] (\industry@) -- (\distance@);
\draw[<->,thick] (\distance@) -- (\teachers@);
\draw[o->,thick] (\rooms@) -- (\industry@);
\draw[o->,thick] (\industry@) -- (\value@);
\draw[o->,thick] (\crime@) -- (\value@);
\draw[o->,thick] (\crime@) -- (\distance@);
\draw[o->,thick] (\rooms@) -- (\teachers@);
\draw[o->,thick] (\rooms@) -- (\value@);
\draw[->,ultra thick] (\teachers@) -- (\value@);
\makeatother
    \end{tikzpicture}
    \caption{FCI} \label{subfig:Boston-fci}
  \end{subfigure}
  \hfill
  \begin{subfigure}[b]{0.48\linewidth}
    \begin{tikzpicture}[>=latex', scale=0.78]
      \small
      \input{TikZ/auto/Boston/layout}
\draw[-,thick] (B) -- (CRIM);
\draw[->,thick] (B) -- (MEDV);
\draw[->,thick] (B) -- (RAD);
\draw[->,thick] (CRIM) -- (MEDV);
\draw[->,thick] (CRIM) -- (RAD);
\draw[->,thick] (CRIM) -- (RM);
\draw[->,thick] (PTRATIO) -- (MEDV);
\draw[->,thick] (PTRATIO) -- (RM);
\draw[->,thick] (RAD) -- (INDUS);
\draw[->,thick] (RAD) -- (PTRATIO);
\draw[->,thick] (RM) -- (INDUS);
\draw[->,thick] (RM) -- (MEDV);
    \end{tikzpicture}
    \caption{GES} \label{subfig:Boston-ges}
  \end{subfigure}
  \hfill
  \begin{subfigure}[b]{0.48\linewidth}
    \begin{tikzpicture}[>=latex', scale=0.78]
      \small
      \input{TikZ/auto/Boston/layout}
\makeatletter
\def\race@{B}
\def\distance@{RAD}
\def\value@{MEDV}
\def\crime@{CRIM}
\def\teachers@{PTRATIO}
\def\industry@{INDUS}
\def\rooms@{RM}

\draw[->,thick] (\race@) -- (\distance@);
\draw[->,thick] (\race@) -- (\value@);
\draw[->,thick] (\distance@) -- (\crime@);
\draw[->,thick] (\distance@) -- (\industry@);
\draw[->,thick] (\distance@) -- (\teachers@);
\draw[->,thick] (\industry@) -- (\rooms@);
\draw[->,thick] (\industry@) -- (\value@);
\draw[->,thick] (\crime@) -- (\value@);
\draw[->,thick] (\value@) -- (\rooms@);
\draw[->,thick] (\value@) -- (\teachers@);
\makeatother
    \end{tikzpicture}
    \caption{Direct LiNGAM} \label{subfig:Boston-ling}
  \end{subfigure}
  \caption{Generated causal graph for the Boston Housing dataset.}
  \label{fig:BostonCG}
\end{figure}

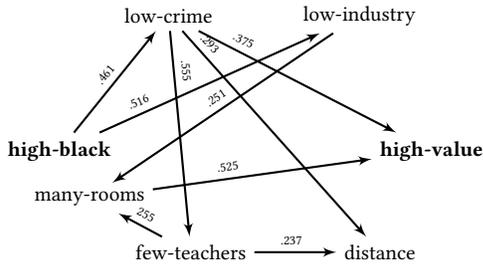
\begin{figure}[H]  
  \centering
  \begin{tikzpicture}[>=latex', scale=0.9]
    \small
    \draw[-,thick] (-.5,-.5) node[left] (high-black) {\textbf{high-black}};
\draw[-,thick](5,-.5) node[left] (high-value) {\textbf{high-value}};
\draw[-,thick] (4,1.5) node[left] (low-industry) {low-industry};
\draw[-,thick] (4,-2) node[left] (distance) {distance};
\draw[-,thick] (1,1.5) node[left] (low-crime) {low-crime};
\draw[-,thick] (1.5,-2) node[left] (few-teachers) {few-teachers};
\draw[-,thick] (0,-1.2) node[left] (many-rooms) {many-rooms};
\draw[->,thick] (high-black) -- (low-crime) node [midway, above, sloped] (TextNode) {\tiny .461};
\draw[->,thick] (high-black) -- (low-industry) node [pos=.2, above, sloped] (TextNode) {\tiny .516};
\draw[->,thick] (low-crime) -- (high-value) node [pos=.2, above, sloped] (TextNode) {\tiny .375};
\draw[->,thick] (low-crime) -- (distance) node [pos=.1, above, sloped] (TextNode) {\tiny .293};
\draw[->,thick] (low-crime) -- (few-teachers) node [pos=.2, above, sloped] (TextNode) {\tiny .555};
\draw[->,thick] (low-industry) -- (many-rooms) node [pos=.5, above, sloped] (TextNode) {\tiny .251};
\draw[->,thick] (many-rooms) -- (high-value) node [pos=.35, above, sloped] (TextNode) {\tiny .525};
\draw[->,thick] (few-teachers) -- (many-rooms) node [pos=.5, above, sloped] (TextNode) {\tiny 255};
\draw[->,thick] (few-teachers) -- (distance) node [pos=.45, above, sloped] (TextNode) {\tiny .237};
  \end{tikzpicture}
\caption{SBCN for the communities with high rate of blacks in the Boston Housing dataset.}
\label{fig:bostonSBCN}
\end{figure}  

\begin{figure}[!ht]
    \includegraphics[scale=0.45, bb=0 0 520 300]{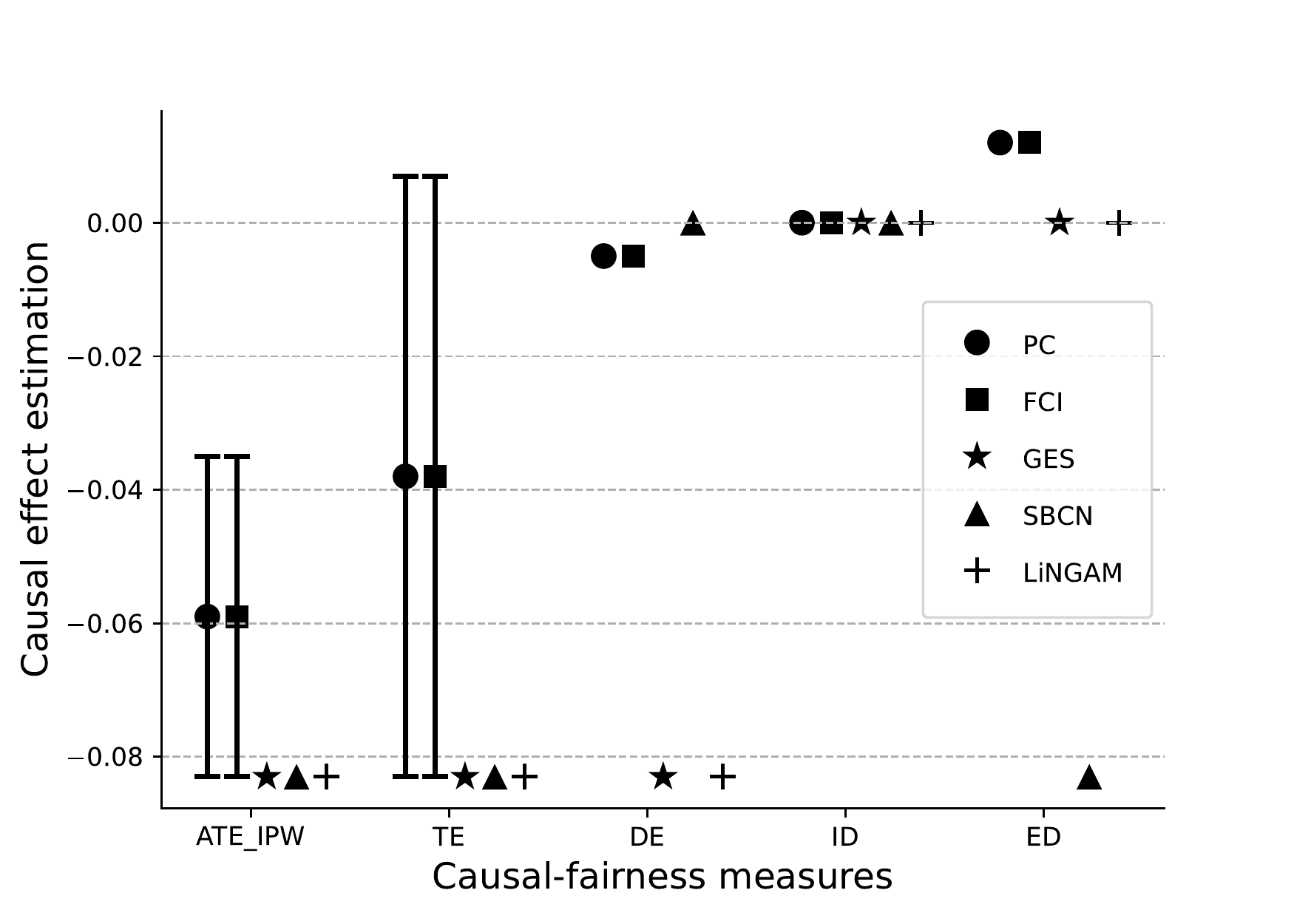} 
    \caption{Estimation of causal effects of the Boston housing dataset based on PC, FCI, GES and SBCN.}
    \label{fig:bostonResults}
\end{figure}

The generated graphs are shown in Figures~\ref{fig:BostonCG} and~\ref{fig:bostonSBCN}. The direct effect appears in all graphs except the one of SBCN. The number of causal paths between race and value vary greatly between graphs. These are $13$, $6$, $2$, $3$, and $3$ according to PC, FCI, GES, LiNGAM, and SBCN respectively. For instance, the two possible paths according to GES are $race \rightarrow value$ and $race \rightarrow distance \rightarrow industry \rightarrow value$.  The most notable feature of the discrimination values in Figure~\ref{fig:bostonResults} is that $ID$ is zero according to all graphs. This is due to the fact that all mediator variables are explaining; crime rate, distance to employment centers, number of rooms in houses, etc. These can be clearly used to legitimately justify discrimination. All measures return either zero or some slightly positive values which, surprisingly, indicate a slight ($\leq 0.08$) discrimination \textit{in favor} of areas with higher proportions of black individuals. 

\subsection{Communities and crime}
\label{communities}
The communities and crime dataset\footnote{https://archive.ics.uci.edu/ml/datasets/communities+and+crime} contains data relevant to per capita violent crime rates in several communities in the United States and the outcome is this crime rate. The variables used for causal graph generation are continuous, namely: race, age, poverty rate, unemployment rate, divorce rate and violent crime rate. Race is considered as a sensitive variable. Three tiers are used: race, age and poverty rate are defined in the first tier. Divorce and unemployment rates are defined in the second tier and  violent crime rate in the last tier. No variable can be considered as explaining and hence we treat them as redlining if found as mediators.


\begin{figure}[H]
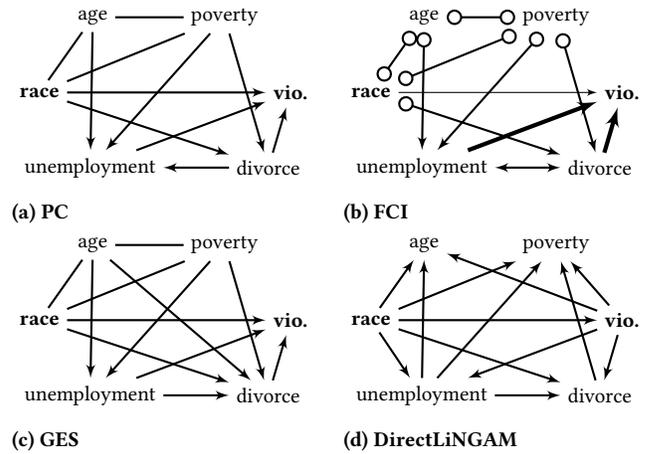

\centering
  \begin{subfigure}[b]{0.48\linewidth}
    \begin{tikzpicture}[>=latex', scale=0.8]
      \small
      \input{TikZ/auto/Communities/layout}
\draw[->,thick] (PctPopUnderPov) -- (PctUnemployed);
\draw[->,thick] (PctPopUnderPov) -- (TotalPctDiv);
\draw[->,thick] (PctUnemployed) -- (ViolentCrimesPerPop);
\draw[->,thick] (TotalPctDiv) -- (PctUnemployed);
\draw[->,thick] (TotalPctDiv) -- (ViolentCrimesPerPop);
\draw[-,thick] (agePct12t29) -- (PctPopUnderPov);
\draw[->,thick] (agePct12t29) -- (PctUnemployed);
\draw[-,thick] (blackrate) -- (PctPopUnderPov);
\draw[->,thick] (blackrate) -- (TotalPctDiv);
\draw[->,thick] (blackrate) -- (ViolentCrimesPerPop);
\draw[-,thick] (blackrate) -- (agePct12t29);
    \end{tikzpicture}
    \caption{PC} \label{subfig:Communities-pc}
  \end{subfigure}
  \hfill
  \begin{subfigure}[b]{0.48\linewidth}
    \begin{tikzpicture}[>=latex', scale=0.8]
      \small
      \input{TikZ/auto/Communities/layout}
\draw[->,ultra thick] (PctUnemployed) -- (ViolentCrimesPerPop);
\draw[->,ultra thick] (TotalPctDiv) -- (ViolentCrimesPerPop);
\draw[->,] (blackrate) -- (ViolentCrimesPerPop);
\draw[o->,thick] (PctPopUnderPov) -- (PctUnemployed);
\draw[o->,thick] (PctPopUnderPov) -- (TotalPctDiv);
\draw[<->,thick] (TotalPctDiv) -- (PctUnemployed);
\draw[o-o,thick] (agePct12t29) -- (PctPopUnderPov);
\draw[o->,thick] (agePct12t29) -- (PctUnemployed);
\draw[o-o,thick] (blackrate) -- (PctPopUnderPov);
\draw[o->,thick] (blackrate) -- (TotalPctDiv);
\draw[o-o,thick] (blackrate) -- (agePct12t29);
    \end{tikzpicture}
    \caption{FCI} \label{subfig:Communities-fci}
  \end{subfigure}
  \hfill
  \begin{subfigure}[b]{0.48\linewidth}
    \begin{tikzpicture}[>=latex', scale=0.8]
      \small
      \input{TikZ/auto/Communities/layout}
\draw[->,thick] (PctPopUnderPov) -- (PctUnemployed);
\draw[->,thick] (PctPopUnderPov) -- (TotalPctDiv);
\draw[-,thick] (PctPopUnderPov) -- (blackrate);
\draw[->,thick] (PctUnemployed) -- (TotalPctDiv);
\draw[->,thick] (PctUnemployed) -- (ViolentCrimesPerPop);
\draw[->,thick] (TotalPctDiv) -- (ViolentCrimesPerPop);
\draw[-,thick] (agePct12t29) -- (PctPopUnderPov);
\draw[->,thick] (agePct12t29) -- (PctUnemployed);
\draw[->,thick] (agePct12t29) -- (TotalPctDiv);
\draw[-,thick] (agePct12t29) -- (blackrate);
\draw[->,thick] (blackrate) -- (TotalPctDiv);
\draw[->,thick] (blackrate) -- (ViolentCrimesPerPop);
    \end{tikzpicture}
    \caption{GES} \label{subfig:Communities-ges}
  \end{subfigure}
  \hfill
  \begin{subfigure}[b]{0.48\linewidth}
    \begin{tikzpicture}[>=latex', scale=0.8]
      \small
      \input{TikZ/auto/Communities/layout}
      \makeatletter
\def\race@{blackrate}
\def\violence@{ViolentCrimesPerPop}
\def\age@{agePct12t29}
\def\poverty@{PctPopUnderPov}
\def\unemployment@{PctUnemployed}
\def\divorce@{TotalPctDiv}

\draw[->,thick] (\race@) -- (\violence@);
\draw[->,thick] (\race@) -- (\poverty@);
\draw[->,thick] (\race@) -- (\unemployment@);
\draw[->,thick] (\race@) -- (\divorce@);
\draw[->,thick] (\race@) -- (\age@);
\draw[->,thick] (\violence@) -- (\unemployment@);
\draw[->,thick] (\violence@) -- (\age@);
\draw[->,thick] (\violence@) -- (\divorce@);
\draw[->,thick] (\violence@) -- (\poverty@);
\draw[->,thick] (\unemployment@) -- (\age@);
\draw[->,thick] (\unemployment@) -- (\divorce@);
\draw[->,thick] (\unemployment@) -- (\poverty@);
\draw[->,thick] (\divorce@) -- (\poverty@);
\makeatother
    \end{tikzpicture}
    \caption{DirectLiNGAM} \label{subfig:Communities-direct_lingam}
  \end{subfigure}
  \caption{Generated causal graph for the communities and crime census dataset. Vio. stands for violence.}
  \label{fig:CommunitiesCG}
\end{figure}

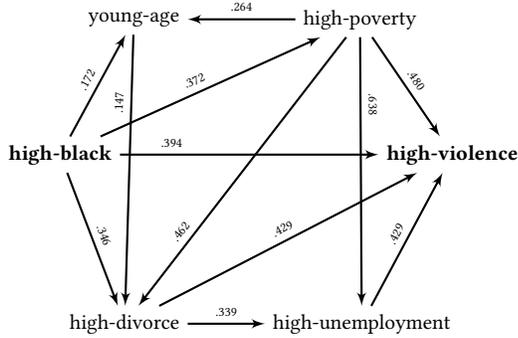
\begin{figure}[H]  
\captionsetup[sub]{justification=centering}
  \begin{subfigure}[b]{0.9\linewidth}
    \begin{tikzpicture}[>=latex', scale=0.9]
      \small
      \draw[-,thick] (-.5,-.5) node[left] (high-black) {\textbf{high-black}};
\draw[-,thick](5.5,-.5) node[left] (high-violence) {\textbf{high-violence}};
\draw[-,thick] (4,1.5) node[left] (high-poverty) {high-poverty};
\draw[-,thick] (4.5,-3) node[left] (high-unemployment) {high-unemployment};
\draw[-,thick] (.5,1.5) node[left] (young-age) {young-age};
\draw[-,thick] (.5,-3) node[left] (high-divorce) {high-divorce};
\draw[->,thick] (high-black) -- (high-violence) node [pos=.2, above, sloped] (TextNode) {\tiny .394};
\draw[->,thick] (high-black) -- (high-poverty) node [pos=.45, above, sloped] (TextNode) {\tiny .372};
\draw[->,thick] (high-black) -- (young-age) node [pos=.5, above, sloped] (TextNode) {\tiny .172};
\draw[->,thick] (high-black) -- (high-divorce) node [pos=.5, above, sloped] (TextNode) {\tiny .346};
\draw[->,thick] (high-poverty) -- (high-violence) node [midway, above, sloped] (TextNode) {\tiny .480};
\draw[->,thick] (high-poverty) -- (high-unemployment) node [pos=.25, above, sloped] (TextNode) {\tiny .638};
\draw[->,thick] (high-poverty) -- (high-divorce) node [pos=.75, above, sloped] (TextNode) {\tiny .462};
\draw[->,thick] (high-poverty) -- (young-age) node [midway, above, sloped] (TextNode) {\tiny .264};
\draw[->,thick] (high-unemployment) -- (high-violence) node [midway, above, sloped] (TextNode) {\tiny .429};
\draw[->,thick] (high-divorce) -- (high-violence) node [midway, above, sloped] (TextNode) {\tiny .429};
\draw[->,thick] (high-divorce) -- (high-unemployment) node [midway, above, sloped] (TextNode) {\tiny .339};
\draw[->,thick] (young-age) -- (high-divorce) node [pos=.25, above, sloped] (TextNode) {\tiny .147};
    \end{tikzpicture}
    \end{subfigure} 
  \caption{SBCN for the communities and crime dataset.}
  \label{fig:communitiesSBCN}
\end{figure} 

\begin{figure}[!ht]
    \includegraphics[scale=0.45, bb=0 0 520 400]{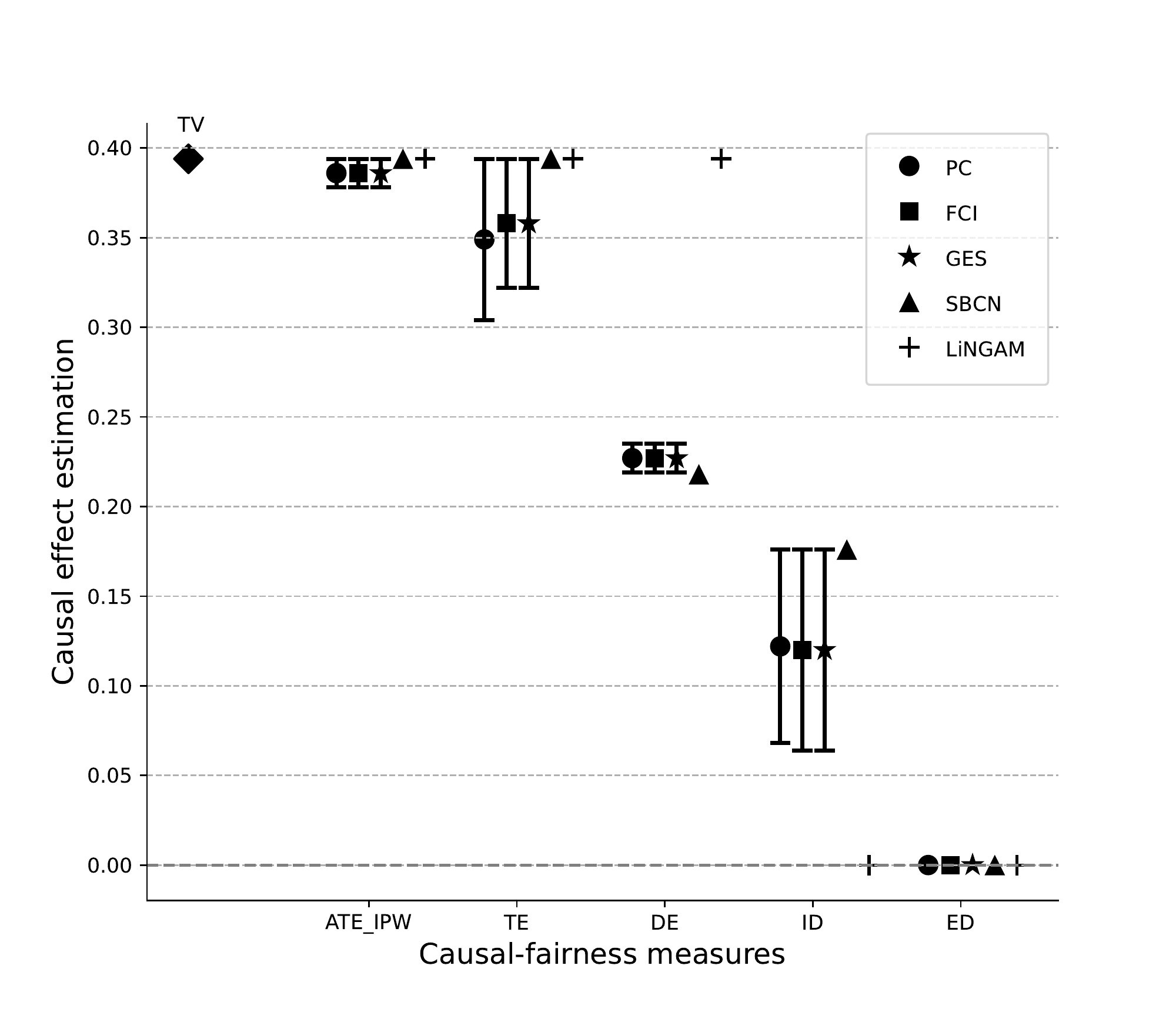} 
    \caption{Estimation of causal effects of the communities and crime dataset based on PC, FCI, GES and SBCN.}
    \label{fig:communitiesResults}
\end{figure}

Figures~\ref{fig:CommunitiesCG} and~\ref{fig:communitiesSBCN} show the generated graphs. The direct edge between race and violence is identified by all algorithms. PC and FCI generated graphs differ only in the direction of the unemployment-divorce edge. The numbers of causal paths possible in all graphs are similar with the striking exception of LiNGAM. These numbers are $7$, $9$, $9$, and $9$ for PC, FCI, GES, and SBCN respectively, but only $1$ (the direct path $race \rightarrow violence$) for LiNGAM. This can be a strong indicator that the dataset does not satisfy the LiNGAM assumption. In particular, the non-Gaussianity of the noise terms. Figure~\ref{fig:communitiesResults} shows the values of the discrimination measures. Both measures of total effect indicate a significant discrimination (almost $0.4$) against blacks. $TE$ ranges from $0.303$ when age is a confounder and $0.394$ when age is a mediator. There are no confounders according to LiNGAM and SBCN graphs, hence, $TE$ coincides with total variation (Equation~\ref{eq:tv}). $DE$ values are comparable with the exception of LiNGAM. In the latter, since the direct edge is the only causal path between race and violence, $DE$ coincides also with $TE$ and all indirect effects ($ID$ and $ED$) are equal to zero. The high variability in $ID$ values is directly linked to the role of age variable (whether it is a confounder or a mediator). $ED$ is zero according to all graphs since there are no explaining variables.

It is important to mention that despite the flawed graph returned by LiNGAM, the total effect is similar to the values computed based on other graphs because all discrimination is considered direct and hence the indirect discrimination is zero. This does not reflect the correct mediation analysis returned by the other more reliable graphs. In other words, the total effect value according to LiNGAM is correct, but the direct and indirect discrimination values are flawed. 
More generally, in case of absence of confounding between the sensitive attribute and the outcome variable, a flawed causal model does not impact the reliability of total effect as the latter coincides with total variation which can be computed independently of the causal graph. However, the splitting of the causal effect between direct, indirect and explained types of discrimination depends heavily on the mediation structure of the graph.

\section{Conclusion}
\label{sec:conc}
In this paper, we provided a detailed and intuitive explanation of the major causal discovery algorithms in the literature. Causal relations between variables are typically identified from observable data using causal discovery algorithms as experiments and interventions (randomized controlled trials (RCT) and A-B testing) are difficult to carry out in discrimination scenarios (requires changing inherent attributes of individuals such as gender or race).
Constraint and score based approaches to causal discovery rely mainly on conditional independence tests and hence generate typically PDAGs with undirected edges. The third category relies rather on the independence between the cause variable and the residual of the regression to decide about the direction of the edges.

The main contributions of the paper are two-fold. First, we show how the subtle differences between the causal discovery algorithms can explain why they generate different causal graphs. Second, and foremost, we demonstrate how slight differences  between causal graphs may have significant impact on fairness/discrimination conclusions.   

Most of the causal approaches to fairness in the literature do not tackle the causal graph generation task. With this study, we hope to raise the awareness about the importance of this step in the fairness assessment and enforcement pipeline as any difference in the structure of the graph may lead to very different fairness conclusions. 
A natural follow-up work after this study is to design a new causal discovery algorithm specifically tuned for fairness. This algorithm can be an adaptation of an existing algorithm but geared towards discovering accurately the causal effect of the sensitive attribute on the outcome variable along the various directed paths. Another future direction would be to study the impact of pre-processing transformations on the structure of the generated graph and consequently on the fairness conclusions.

\section{Acknowledgements}
This work was supported by the European Research Council (ERC) project HYPATIA under the European Union’s Horizon 2020 research and innovation programme. Grant agreement n. 835294.

\bibliography{Bibliography}

\begin{thebibliography}{77}
\providecommand{\natexlab}[1]{#1}
\providecommand{\url}[1]{\texttt{#1}}
\expandafter\ifx\csname urlstyle\endcsname\relax
  \providecommand{\doi}[1]{doi: #1}\else
  \providecommand{\doi}{doi: \begingroup \urlstyle{rm}\Url}\fi

\bibitem[Act(1964)]{act1964civil}
Civil~Rights Act.
\newblock Civil rights act of 1964.
\newblock \emph{Title VII, Equal Employment Opportunities}, 1964.

\bibitem[Andrews et~al.(2018)Andrews, Ramsey, and Cooper]{andrews2018scoring}
Bryan Andrews, Joseph Ramsey, and Gregory~F Cooper.
\newblock Scoring bayesian networks of mixed variables.
\newblock \emph{International journal of data science and analytics},
  6\penalty0 (1):\penalty0 3--18, 2018.

\bibitem[Angwin et~al.(2016)Angwin, Larson, Mattu, and
  Kirchner]{angwin2016machine}
Julia Angwin, Jeff Larson, Surya Mattu, and Lauren Kirchner.
\newblock Machine bias. propublica.
\newblock \emph{See https://www. propublica.
  org/article/machine-bias-risk-assessments-in-criminal-sentencing}, 2016.

\bibitem[Bach and Jordan(2002)]{bach2002kernel}
Francis~R Bach and Michael~I Jordan.
\newblock Kernel independent component analysis.
\newblock \emph{Journal of machine learning research}, 3\penalty0
  (Jul):\penalty0 1--48, 2002.

\bibitem[Barocas and Selbst(2016)]{barocas2016big}
Solon Barocas and Andrew~D Selbst.
\newblock Big data's disparate impact.
\newblock \emph{Calif. L. Rev.}, 104:\penalty0 671, 2016.

\bibitem[Bickel et~al.(1975)Bickel, Hammel, and O'Connell]{berkeley75}
Peter~J Bickel, Eugene~A Hammel, and J~William O'Connell.
\newblock Sex bias in graduate admissions: Data from berkeley.
\newblock \emph{Science}, 187\penalty0 (4175):\penalty0 398--404, 1975.

\bibitem[Bonchi et~al.(2017)Bonchi, Hajian, Mishra, and
  Ramazzotti]{bonchi2017exposing}
Francesco Bonchi, Sara Hajian, Bud Mishra, and Daniele Ramazzotti.
\newblock Exposing the probabilistic causal structure of discrimination.
\newblock \emph{International Journal of Data Science and Analytics},
  3\penalty0 (1):\penalty0 1--21, 2017.

\bibitem[Cameron(2018)]{ECHR}
Iain Cameron.
\newblock \emph{An introduction to the European convention on human rights}.
\newblock Iustus f{\"o}rlag, 2018.

\bibitem[Cheng et~al.(2022)Cheng, Guo, Moraffah, Sheth, Candan, and
  Liu]{cheng2022evaluation}
Lu~Cheng, Ruocheng Guo, Raha Moraffah, Paras Sheth, Kasim~Selcuk Candan, and
  Huan Liu.
\newblock Evaluation methods and measures for causal learning algorithms.
\newblock \emph{IEEE Transactions on Artificial Intelligence}, 2022.

\bibitem[Chiappa(2019)]{chiappa2019path}
Silvia Chiappa.
\newblock Path-specific counterfactual fairness.
\newblock In \emph{Proceedings of the AAAI Conference on Artificial
  Intelligence}, volume~33, pages 7801--7808, 2019.

\bibitem[Chickering(2002)]{chickering2002optimal}
David~Maxwell Chickering.
\newblock Optimal structure identification with greedy search.
\newblock \emph{Journal of machine learning research}, 3\penalty0
  (Nov):\penalty0 507--554, 2002.

\bibitem[Chouldechova(2017)]{chouldechova2017fair}
Alexandra Chouldechova.
\newblock Fair prediction with disparate impact: A study of bias in recidivism
  prediction instruments.
\newblock \emph{Big data}, 5\penalty0 (2):\penalty0 153--163, 2017.

\bibitem[Darlington(1971)]{darlington1971}
Richard~B Darlington.
\newblock Another look at “cultural fairness”.
\newblock \emph{Journal of educational measurement}, 8\penalty0 (2):\penalty0
  71--82, 1971.

\bibitem[De~Schutter(2019)]{InternationalHR}
Olivier De~Schutter.
\newblock \emph{International human rights law}.
\newblock Cambridge University Press, 2019.

\bibitem[Dor and Tarsi(1992)]{dor1992simple}
Dorit Dor and Michael Tarsi.
\newblock A simple algorithm to construct a consistent extension of a partially
  oriented graph.
\newblock \emph{Technicial Report R-185, Cognitive Systems Laboratory, UCLA},
  1992.

\bibitem[Eberhardt(2009)]{eberhardt2009introduction}
Frederick Eberhardt.
\newblock Introduction to the epistemology of causation.
\newblock \emph{Philosophy Compass}, 4\penalty0 (6):\penalty0 913--925, 2009.

\bibitem[Fisher(1936)]{fisher1936design}
Ronald~Aylmer Fisher.
\newblock Design of experiments.
\newblock \emph{British Medical Journal}, 1\penalty0 (3923):\penalty0 554,
  1936.

\bibitem[Gamella(2021)]{gamella2021python}
Juan Gamella.
\newblock Greedy equivalence search (ges) algorithm for causal discovery.
\newblock \url{https://github.com/juangamella/ges}, 2021.
\newblock Accessed: 2022-03-16.

\bibitem[Glymour et~al.(2019)Glymour, Zhang, and Spirtes]{glymour2019review}
Clark Glymour, Kun Zhang, and Peter Spirtes.
\newblock Review of causal discovery methods based on graphical models.
\newblock \emph{Frontiers in genetics}, 10:\penalty0 524, 2019.

\bibitem[Guo et~al.(2020)Guo, Cheng, Li, Hahn, and Liu]{guo2020survey}
Ruocheng Guo, Lu~Cheng, Jundong Li, P~Richard Hahn, and Huan Liu.
\newblock A survey of learning causality with data: Problems and methods.
\newblock \emph{ACM Computing Surveys (CSUR)}, 53\penalty0 (4):\penalty0 1--37,
  2020.

\bibitem[Hardt et~al.(2016)Hardt, Price, and Srebro]{hardt2016equality}
Moritz Hardt, Eric Price, and Nati Srebro.
\newblock Equality of opportunity in supervised learning.
\newblock In \emph{Advances in neural information processing systems}, pages
  3315--3323, Spain, 2016.

\bibitem[Harrison~Jr and Rubinfeld(1978)]{harrison1978hedonic}
David Harrison~Jr and Daniel~L Rubinfeld.
\newblock Hedonic housing prices and the demand for clean air.
\newblock \emph{Journal of environmental economics and management}, 5\penalty0
  (1):\penalty0 81--102, 1978.

\bibitem[Haslam and McGarty(2004)]{haslam2004experimental}
S~Alexander Haslam and Craig McGarty.
\newblock Experimental design and causality in social psychological research.
\newblock \emph{C., Sansone, CC, Morf, AT Panter,(Eds.), The Sage handbook of
  methods in social psychology}, pages 237--264, 2004.

\bibitem[Hauser and B\"uhlmann(2012)]{hauser2012gies}
Alain Hauser and Peter B\"uhlmann.
\newblock Characterization and greedy learning of interventional {M}arkov
  equivalence classes of directed acyclic graphs.
\newblock \emph{Journal of Machine Learning Research}, 13:\penalty0 2409--2464,
  2012.
\newblock URL \url{https://jmlr.org/papers/v13/hauser12a.html}.

\bibitem[Hitchcock(2002)]{hitchcock2002probabilistic}
Christopher Hitchcock.
\newblock Probabilistic causation.
\newblock \emph{Stanford Encylopedia of Philosophy (archive)}, 2002.

\bibitem[Hoyer et~al.(2008{\natexlab{a}})Hoyer, Janzing, Mooij, Peters, and
  Sch{\"o}lkopf]{hoyer2008nonlinear}
Patrik Hoyer, Dominik Janzing, Joris~M Mooij, Jonas Peters, and Bernhard
  Sch{\"o}lkopf.
\newblock Nonlinear causal discovery with additive noise models.
\newblock \emph{Advances in neural information processing systems}, 21,
  2008{\natexlab{a}}.

\bibitem[Hoyer et~al.(2008{\natexlab{b}})Hoyer, Shimizu, Kerminen, and
  Palviainen]{hoyer2008estimation}
Patrik~O Hoyer, Shohei Shimizu, Antti~J Kerminen, and Markus Palviainen.
\newblock Estimation of causal effects using linear non-gaussian causal models
  with hidden variables.
\newblock \emph{International Journal of Approximate Reasoning}, 49\penalty0
  (2):\penalty0 362--378, 2008{\natexlab{b}}.

\bibitem[Hoyer et~al.(2012)Hoyer, Hyvarinen, Scheines, Spirtes, Ramsey,
  Lacerda, and Shimizu]{hoyer2012causal}
Patrik~O Hoyer, Aapo Hyvarinen, Richard Scheines, Peter~L Spirtes, Joseph
  Ramsey, Gustavo Lacerda, and Shohei Shimizu.
\newblock Causal discovery of linear acyclic models with arbitrary
  distributions.
\newblock \emph{arXiv preprint arXiv:1206.3260}, 2012.

\bibitem[Huan et~al.(2020)Huan, Wu, Zhang, and Wu]{wu2020effort}
Wen Huan, Yongkai Wu, Lu~Zhang, and Xintao Wu.
\newblock Fairness through equality of effort.
\newblock In \emph{Companion Proceedings of the Web Conference 2020}, pages
  743--751, 2020.

\bibitem[Hyv{\"a}rinen and Smith(2013)]{hyvarinen2013pairwise}
Aapo Hyv{\"a}rinen and Stephen~M Smith.
\newblock Pairwise likelihood ratios for estimation of non-gaussian structural
  equation models.
\newblock \emph{Journal of Machine Learning Research}, 14\penalty0
  (Jan):\penalty0 111--152, 2013.

\bibitem[Hyv{\"a}rinen et~al.(2010)Hyv{\"a}rinen, Zhang, Shimizu, and
  Hoyer]{hyvarinen2010estimation}
Aapo Hyv{\"a}rinen, Kun Zhang, Shohei Shimizu, and Patrik~O Hoyer.
\newblock Estimation of a structural vector autoregression model using
  non-gaussianity.
\newblock \emph{Journal of Machine Learning Research}, 11\penalty0 (5), 2010.

\bibitem[Imbens and Rubin(2015)]{rubin2015book}
Guido~W Imbens and Donald~B Rubin.
\newblock \emph{Causal inference in statistics, social, and biomedical
  sciences}.
\newblock Cambridge University Press, 2015.

\bibitem[Kalainathan and Goudet(2019)]{kalainathan2019causal}
Diviyan Kalainathan and Olivier Goudet.
\newblock Causal discovery toolbox: Uncover causal relationships in python.
\newblock \emph{arXiv preprint arXiv:1903.02278}, 2019.

\bibitem[Kalisch et~al.(2012)Kalisch, M\"achler, Colombo, Maathuis, and
  B\"uhlmann]{pcalg}
Markus Kalisch, Martin M\"achler, Diego Colombo, Marloes~H. Maathuis, and Peter
  B\"uhlmann.
\newblock Causal inference using graphical models with the {R} package {pcalg}.
\newblock \emph{Journal of Statistical Software}, 47\penalty0 (11):\penalty0
  1--26, 2012.
\newblock \doi{10.18637/jss.v047.i11}.

\bibitem[Kilbertus et~al.(2017)Kilbertus, Carulla, Parascandolo, Hardt,
  Janzing, and Sch{\"o}lkopf]{kilbertus2017avoiding}
Niki Kilbertus, Mateo~Rojas Carulla, Giambattista Parascandolo, Moritz Hardt,
  Dominik Janzing, and Bernhard Sch{\"o}lkopf.
\newblock Avoiding discrimination through causal reasoning.
\newblock In \emph{Advances in Neural Information Processing Systems}, pages
  656--666, 2017.

\bibitem[Kohavi et~al.(2020)Kohavi, Tang, and Xu]{abtesting20}
Ron Kohavi, Diane Tang, and Ya~Xu.
\newblock \emph{Trustworthy online controlled experiments: A practical guide to
  a/b testing}.
\newblock Cambridge University Press, 2020.

\bibitem[Kusner et~al.(2017)Kusner, Loftus, Russell, and
  Silva]{kusner2017counterfactual}
Matt~J Kusner, Joshua Loftus, Chris Russell, and Ricardo Silva.
\newblock Counterfactual fairness.
\newblock \emph{Advances in neural information processing systems}, 30, 2017.

\bibitem[Lacerda et~al.(2012)Lacerda, Spirtes, Ramsey, and
  Hoyer]{lacerda2012discovering}
Gustavo Lacerda, Peter~L Spirtes, Joseph Ramsey, and Patrik~O Hoyer.
\newblock Discovering cyclic causal models by independent components analysis.
\newblock \emph{arXiv preprint arXiv:1206.3273}, 2012.

\bibitem[Le et~al.(2016)Le, Hoang, Li, Liu, Liu, and Hu]{le2016fast}
Thuc~Duy Le, Tao Hoang, Jiuyong Li, Lin Liu, Huawen Liu, and Shu Hu.
\newblock A fast pc algorithm for high dimensional causal discovery with
  multi-core pcs.
\newblock \emph{IEEE/ACM transactions on computational biology and
  bioinformatics}, 16\penalty0 (5):\penalty0 1483--1495, 2016.

\bibitem[Makhlouf et~al.(2020{\natexlab{a}})Makhlouf, Zhioua, and
  Palamidessi]{makhlouf2020causal}
Karima Makhlouf, Sami Zhioua, and Catuscia Palamidessi.
\newblock Survey on causal-based machine learning fairness notions.
\newblock \emph{arXiv preprint arXiv:2010.09553}, 2020{\natexlab{a}}.

\bibitem[Makhlouf et~al.(2020{\natexlab{b}})Makhlouf, Zhioua, and
  Palamidessi]{makhlouf2020survey}
Karima Makhlouf, Sami Zhioua, and Catuscia Palamidessi.
\newblock Survey on causal-based machine learning fairness notions.
\newblock \emph{arXiv preprint arXiv:2010.09553}, 2020{\natexlab{b}}.

\bibitem[Makhlouf et~al.(2021)Makhlouf, Zhioua, and
  Palamidessi]{makhlouf2021ipm}
Karima Makhlouf, Sami Zhioua, and Catuscia Palamidessi.
\newblock Machine learning fairness notions: Bridging the gap with real-world
  applications.
\newblock \emph{Information Processing \& Management}, 58\penalty0
  (5):\penalty0 102642, 2021.

\bibitem[Makhlouf et~al.(2022)Makhlouf, Zhioua, and
  Palamidessi]{makhlouf2022identifiability}
Karima Makhlouf, Sami Zhioua, and Catuscia Palamidessi.
\newblock Identifiability of causal-based fairness notions: A state of the art.
\newblock \emph{arXiv preprint arXiv:2203.05900}, 2022.

\bibitem[Malinsky and Danks(2018)]{malinsky2018causal}
Daniel Malinsky and David Danks.
\newblock Causal discovery algorithms: A practical guide.
\newblock \emph{Philosophy Compass}, 13\penalty0 (1):\penalty0 e12470, 2018.

\bibitem[Mani et~al.(2012)Mani, Spirtes, and Cooper]{mani2012theoretical}
Subramani Mani, Peter~L Spirtes, and Gregory~F Cooper.
\newblock A theoretical study of y structures for causal discovery.
\newblock \emph{arXiv preprint arXiv:1206.6853}, 2012.

\bibitem[Nabi and Shpitser(2018)]{nabi2018fair}
Razieh Nabi and Ilya Shpitser.
\newblock Fair inference on outcomes.
\newblock In \emph{Proceedings of the... AAAI Conference on Artificial
  Intelligence. AAAI Conference on Artificial Intelligence}, volume 2018, page
  1931. NIH Public Access, 2018.

\bibitem[Nogueira et~al.(2021)Nogueira, Gama, and Ferreira]{nogueira2021causal}
Ana~Rita Nogueira, Jo{\~a}o Gama, and Carlos~Abreu Ferreira.
\newblock Causal discovery in machine learning: Theories and applications.
\newblock \emph{Journal of Dynamics \& Games}, 8\penalty0 (3):\penalty0 203,
  2021.

\bibitem[Nogueira et~al.(2022)Nogueira, Pugnana, Ruggieri, Pedreschi, and
  Gama]{nogueira2022methods}
Ana~Rita Nogueira, Andrea Pugnana, Salvatore Ruggieri, Dino Pedreschi, and
  Jo{\~a}o Gama.
\newblock Methods and tools for causal discovery and causal inference.
\newblock \emph{Wiley Interdisciplinary Reviews: Data Mining and Knowledge
  Discovery}, page e1449, 2022.

\bibitem[Oja and Hyvarinen(2000)]{oja2000independent}
Erkki Oja and A~Hyvarinen.
\newblock Independent component analysis: algorithms and applications.
\newblock \emph{Neural networks}, 13\penalty0 (4-5):\penalty0 411--430, 2000.

\bibitem[Pearl(1988)]{pearl1988probabilistic}
Judea Pearl.
\newblock \emph{Probabilistic reasoning in intelligent systems: networks of
  plausible inference}.
\newblock Morgan kaufmann, 1988.

\bibitem[Pearl(2001)]{pearl01direct}
Judea Pearl.
\newblock Direct and indirect effects.
\newblock In \emph{Proceedings of the Seventeenth conference on Uncertainty in
  artificial intelligence}, pages 411--420, 2001.

\bibitem[Pearl(2009)]{pearl2009causality}
Judea Pearl.
\newblock \emph{Causality}.
\newblock Cambridge university press, 2009.

\bibitem[Ramsey et~al.(2018{\natexlab{a}})Ramsey, Zhang, Glymour, Romero,
  Huang, Ebert-Uphoff, Samarasinghe, Barnes, and Glymour]{ramsey2018tetrad}
Joseph~D Ramsey, Kun Zhang, Madelyn Glymour, Ruben~Sanchez Romero, Biwei Huang,
  Imme Ebert-Uphoff, Savini Samarasinghe, Elizabeth~A Barnes, and Clark
  Glymour.
\newblock Tetrad—a toolbox for causal discovery.
\newblock In \emph{8th International Workshop on Climate Informatics},
  2018{\natexlab{a}}.

\bibitem[Ramsey et~al.(2018{\natexlab{b}})Ramsey, Zhang, Glymour, Romero,
  Huang, Ebert-Uphoff, Samarasinghe, Barnes, and Glymour]{tetrad}
Joseph~D Ramsey, Kun Zhang, Madelyn Glymour, Ruben~Sanchez Romero, Biwei Huang,
  Imme Ebert-Uphoff, Savini Samarasinghe, Elizabeth~A Barnes, and Clark
  Glymour.
\newblock Tetrad—a toolbox for causal discovery.
\newblock In \emph{8th International Workshop on Climate Informatics},
  2018{\natexlab{b}}.

\bibitem[Schwarz(1978)]{schwarz1978estimating}
Gideon Schwarz.
\newblock Estimating the dimension of a model.
\newblock \emph{The annals of statistics}, pages 461--464, 1978.

\bibitem[Scutari(2010)]{scutari2010learning}
M~Scutari.
\newblock Learning bayesian networks with the bnlearn r package.
\newblock \emph{Journal of Statistical Software}, 35\penalty0 (3), 2010.

\bibitem[Shimizu(2014)]{shimizu2014lingam}
Shohei Shimizu.
\newblock Lingam: Non-gaussian methods for estimating causal structures.
\newblock \emph{Behaviormetrika}, 41\penalty0 (1):\penalty0 65--98, 2014.

\bibitem[Shimizu et~al.(2006)Shimizu, Hoyer, Hyv{\"a}rinen, Kerminen, and
  Jordan]{shimizu2006linear}
Shohei Shimizu, Patrik~O Hoyer, Aapo Hyv{\"a}rinen, Antti Kerminen, and Michael
  Jordan.
\newblock A linear non-gaussian acyclic model for causal discovery.
\newblock \emph{Journal of Machine Learning Research}, 7\penalty0 (10), 2006.

\bibitem[Shimizu et~al.(2011)Shimizu, Inazumi, Sogawa, Hyv{\"a}rinen, Kawahara,
  Washio, Hoyer, and Bollen]{shimizu2011directlingam}
Shohei Shimizu, Takanori Inazumi, Yasuhiro Sogawa, Aapo Hyv{\"a}rinen,
  Yoshinobu Kawahara, Takashi Washio, Patrik~O Hoyer, and Kenneth Bollen.
\newblock Directlingam: A direct method for learning a linear non-gaussian
  structural equation model.
\newblock \emph{The Journal of Machine Learning Research}, 12:\penalty0
  1225--1248, 2011.

\bibitem[Shpitser and Pearl(2008)]{shpitser08}
Ilya Shpitser and Judea Pearl.
\newblock Complete identification methods for the causal hierarchy.
\newblock \emph{Journal of Machine Learning Research}, 9\penalty0
  (Sep):\penalty0 1941--1979, 2008.

\bibitem[Simpson(1951)]{simpson1951interpretation}
Edward~H Simpson.
\newblock The interpretation of interaction in contingency tables.
\newblock \emph{Journal of the Royal Statistical Society: Series B
  (Methodological)}, 13\penalty0 (2):\penalty0 238--241, 1951.

\bibitem[Sondhi and Shojaie(2019)]{sondhi2019reduced}
Arjun Sondhi and Ali Shojaie.
\newblock The reduced pc-algorithm: Improved causal structure learning in large
  random networks.
\newblock \emph{J. Mach. Learn. Res.}, 20\penalty0 (164):\penalty0 1--31, 2019.

\bibitem[Spirtes and Glymour(1991)]{spirtes1991algorithm}
Peter Spirtes and Clark Glymour.
\newblock An algorithm for fast recovery of sparse causal graphs.
\newblock \emph{Social science computer review}, 9\penalty0 (1):\penalty0
  62--72, 1991.

\bibitem[Spirtes and Zhang(2016)]{spirtes2016survey}
Peter Spirtes and Kun Zhang.
\newblock Causal discovery and inference: concepts and recent methodological
  advances.
\newblock In \emph{Applied informatics}, volume~3, pages 1--28. SpringerOpen,
  2016.

\bibitem[Spirtes et~al.(1999)Spirtes, Meek, and
  Richardson]{spirtes1999algorithm}
Peter Spirtes, Christopher Meek, and Thomas Richardson.
\newblock An algorithm for causal inference in the presence of latent variables
  and selection bias.
\newblock \emph{Computation, causation, and discovery}, 21:\penalty0 211--252,
  1999.

\bibitem[Suppes(1973)]{suppes1973probabilistic}
Patrick Suppes.
\newblock A probabilistic theory of causality.
\newblock \emph{British Journal for the Philosophy of Science}, 24\penalty0
  (4), 1973.

\bibitem[Tsagris(2019)]{tsagris2019bayesian}
Michail Tsagris.
\newblock Bayesian network learning with the pc algorithm: an improved and
  correct variation.
\newblock \emph{Applied Artificial Intelligence}, 33\penalty0 (2):\penalty0
  101--123, 2019.

\bibitem[Wu et~al.(2018)Wu, Zhang, and Wu]{wu18Ranking}
Yongkai Wu, Lu~Zhang, and Xintao Wu.
\newblock On discrimination discovery and removal in ranked data using causal
  graph.
\newblock In \emph{Proceedings of the 24th ACM SIGKDD International Conference
  on Knowledge Discovery \& Data Mining}, pages 2536--2544, 2018.

\bibitem[Wu et~al.(2019)Wu, Zhang, and Wu]{wu2019counterfactual}
Yongkai Wu, Lu~Zhang, and Xintao Wu.
\newblock Counterfactual fairness: Unidentification, bound and algorithm.
\newblock In \emph{Proceedings of the Twenty-Eighth International Joint
  Conference on Artificial Intelligence}, 2019.

\bibitem[Yan et~al.(2020)Yan, Gu, Lin, and Rzeszotarski]{silva2020}
Jing~Nathan Yan, Ziwei Gu, Hubert Lin, and Jeffrey~M Rzeszotarski.
\newblock Silva: Interactively assessing machine learning fairness using
  causality.
\newblock In \emph{Proceedings of the 2020 CHI Conference on Human Factors in
  Computing Systems}, pages 1--13, 2020.

\bibitem[Yu et~al.(2016)Yu, Li, and Liu]{yu2016review}
Kui Yu, Jiuyong Li, and Lin Liu.
\newblock A review on algorithms for constraint-based causal discovery.
\newblock \emph{arXiv preprint arXiv:1611.03977}, 2016.

\bibitem[Zhang(2008)]{zhang2008completeness}
Jiji Zhang.
\newblock On the completeness of orientation rules for causal discovery in the
  presence of latent confounders and selection bias.
\newblock \emph{Artificial Intelligence}, 172\penalty0 (16-17):\penalty0
  1873--1896, 2008.

\bibitem[Zhang and Hyvarinen(2012)]{zhang2012identifiability}
Kun Zhang and Aapo Hyvarinen.
\newblock On the identifiability of the post-nonlinear causal model.
\newblock \emph{arXiv preprint arXiv:1205.2599}, 2012.

\bibitem[Zhang et~al.(2012)Zhang, Peters, Janzing, and
  Sch{\"o}lkopf]{zhang2012kernel}
Kun Zhang, Jonas Peters, Dominik Janzing, and Bernhard Sch{\"o}lkopf.
\newblock Kernel-based conditional independence test and application in causal
  discovery.
\newblock \emph{arXiv preprint arXiv:1202.3775}, 2012.

\bibitem[Zhang et~al.(2016)Zhang, Wu, and Wu]{zhang2016causal}
Lu~Zhang, Yongkai Wu, and Xintao Wu.
\newblock A causal framework for discovering and removing direct and indirect
  discrimination.
\newblock \emph{arXiv preprint arXiv:1611.07509}, 2016.

\bibitem[Zhang et~al.(2017)Zhang, Wu, and Wu]{zhang2017achieving}
Lu~Zhang, Yongkai Wu, and Xintao Wu.
\newblock Achieving non-discrimination in data release.
\newblock In \emph{Proceedings of the 23rd ACM SIGKDD International Conference
  on Knowledge Discovery and Data Mining}, pages 1335--1344, 2017.

\bibitem[Zhou and Yamamoto(2020)]{zhou2020tracing}
Xiang Zhou and Teppei Yamamoto.
\newblock Tracing causal paths from experimental and observational data.
\newblock \emph{SocArXiv. January}, 11, 2020.

\end{thebibliography}
\bibliographystyle{plainnat}

\end{document}